\documentclass{article}

\usepackage[main, final]{neurips_2026}
\makeatletter
\renewcommand{\@notice}{}
\makeatother

\usepackage[utf8]{inputenc} 
\usepackage[T1]{fontenc}    
\usepackage{hyperref}       
\usepackage{url}            
\usepackage{booktabs}       
\usepackage{amsfonts}       
\usepackage{nicefrac}       
\usepackage{microtype}      
\usepackage{xcolor}         

\usepackage{titletoc}

\definecolor{cc1}{HTML}{184E77}
\definecolor{cc2}{HTML}{34a0a4}
\definecolor{cc3}{HTML}{52B69A}
\definecolor{cc4}{HTML}{c200fb}
\definecolor{cc5}{HTML}{9b72cf}

\usepackage{amsmath, amsthm, amssymb}
\usepackage{bm}
\usepackage{graphicx} 
\usepackage{tabularx}
\usepackage{booktabs}
\usepackage{subcaption}
\usepackage{enumitem}
\usepackage[disable]{todonotes}

\usepackage{tikz}
\usetikzlibrary{circuits.logic.US, positioning, arrows.meta, backgrounds, shapes.arrows, shapes.symbols, shapes.geometric}
\usetikzlibrary{decorations.pathreplacing,calligraphy, patterns, calc, decorations.pathmorphing}
\usepackage{tikz-cd}
\usetikzlibrary{calc}

\newcommand{\EC}[1]{%
	\todo[
	color=blue!15,
	size=\footnotesize
	]{\textbf{\color{blue!70!black}EC:}~#1}%
}

\newcommand{\AB}[1]{%
	\todo[
	color=purple!15,
	size=\footnotesize
	]{\textbf{\color{purple!70!black}AB:}~#1}%
}

\newcommand{\rev}[1]{%
	{#1}%
}

\newcommand{\arev}[1]{%
	{#1}%
}

\newcommand*\diff{\mathop{}\!\mathrm{d}}
\newcommand{\R}{\mathbb R}
\renewcommand{\P}{\mathbb P}
\newcommand{\C}{\mathbb C} 
\newcommand{\N}{\mathbb N}
\newcommand{\E}{\mathbb E}

\newcommand{\cN}{\mathcal N}

\newcommand{\cH}{\mathcal H}

\newcommand{\hSig}{\hat{\Sigma}}
\newcommand{\hSigl}{\hat{\Sigma}_{\lambda}}
\newcommand{\hW}{\hat{W}}

\newtheorem{prop}{Proposition}[section]
\newtheorem*{prop*}{Proposition}
\newtheorem{rmk}[prop]{Remark}
\newtheorem*{rmk*}{Remark}
\newtheorem{thm}[prop]{Theorem}
\newtheorem*{thm*}{Theorem}
\newtheorem{defn}[prop]{Definition}
\newtheorem*{defn*}{Definition}
\newtheorem{lem}[prop]{Lemma}
\newtheorem*{lem*}{Lemma}
\newtheorem{cor}[prop]{Corollary}
\newtheorem*{cor*}{Corollary}
\newtheorem{example}[prop]{Example}
\newtheorem*{example*}{Example}
\newtheorem{fact}[prop]{Fact}
\newtheorem{assumption*}{Assumption}

\DeclareMathOperator{\Dom}{Dom}

\DeclareMathOperator{\Int}{Int}
\DeclareMathOperator{\Var}{Var}

\DeclareMathOperator{\supp}{supp}
\DeclareMathOperator{\trace}{tr}
\DeclareMathOperator{\softmax}{Softmax}

\DeclareMathOperator{\SNR}{\sf SNR}
\DeclareMathOperator{\lSNR}{\sf dSNR}
\DeclareMathOperator{\mSNR}{\sf mSNR}
\DeclareMathOperator{\mlSNR}{\sf mdSNR}
\DeclareMathOperator{\dRNR}{\sf dRNR}
\DeclareMathOperator{\RNR}{\sf RNR}
\DeclareMathOperator{\n}{\mathfrak{n}}
\DeclareMathOperator{\m}{\mathfrak{m}}

\DeclareMathOperator*{\argmax}{argmax}

\title{Detection and Classification in Latent Spaces: High-Dimensional Analysis and Validation}

\author{%
  Enis Chenchene\footnotemark[1] \hspace{1cm} Anil S.~Baslamisli\footnotemark[1] \hspace{1cm} Paolo Braca%
    \thanks{Emails: \texttt{\{enis.chenchene,anil.baslamisli,paolo.braca\}@cmre.nato.int}}\\[0.2cm]
    \textsuperscript{*}NATO STO CMRE
}

\begin{document}

\maketitle
\setcounter{footnote}{0}

\begin{abstract}
Foundation models produce high-dimensional latent representations in which suitably designed statistical decision rules can be effective yet analytically tractable. We study such detectors in the high-dimensional regime, where latent dimension and training sample size grow proportionally.
For linear and quadratic statistics, we derive explicit large-deviation characterizations of false-alarm and miss-detection performance under the Neyman--Pearson paradigm, and show how the binary analysis extends to multi-class classification. The resulting error probabilities and corresponding decay rates exhibit non-monotonic dependence on the shape parameter, including double-descent and, in some regimes, multiple-descent.
Experiments on vision, text, and audio datasets using modern foundation models show close agreement between theoretical predictions and empirical performance. In particular, the proposed framework allows us to set a prescribed false-alarm level and accurately predict the corresponding miss-detection probability. Overall, our results provide a theoretically grounded and interpretable framework with competitive empirical performance across different modalities.
\end{abstract}

\begin{figure}[h]
	\centering
	\resizebox{\textwidth}{!}{
		\begin{tikzpicture}

 

  
		\node at (11,10) {\includegraphics[width=0.125\linewidth]{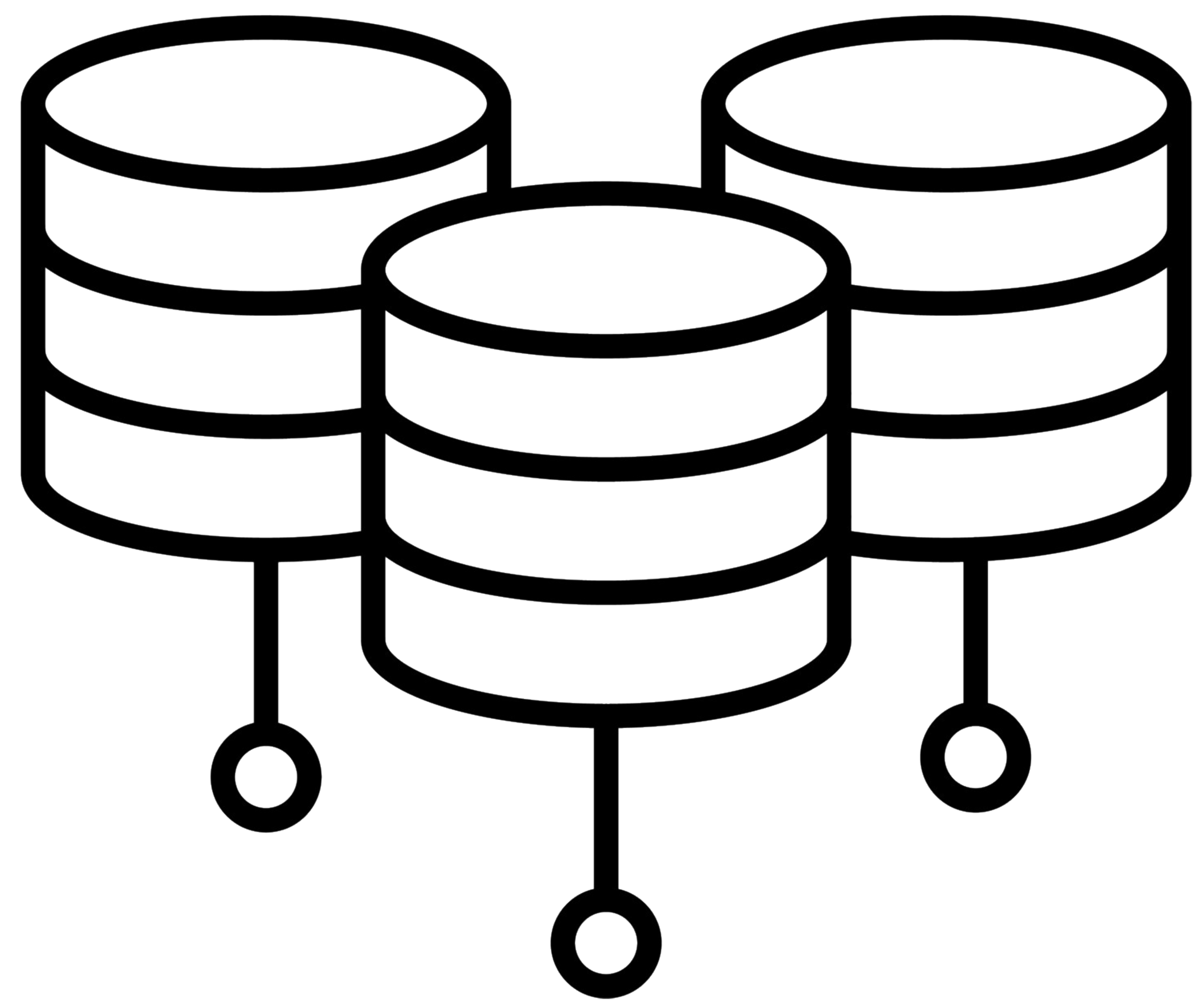}};
		
		\node at (11,8.4) {\small \shortstack{Dataset \\ $M$ classes}};
        \node at (11.2, 11.1) {\small  $\cH_0 \hspace{0.1cm} \raisebox{0.3cm}{$\boldsymbol{.}$}\, \raisebox{0.35cm}{$\boldsymbol{.}$}\,\raisebox{0.3cm}{$\boldsymbol{.}$}  \hspace{0.1cm} \cH_{M-1}$};
		
		
		\node[and gate US, draw=cc1, line width=1pt, fill=cc1!10, logic gate inputs=nn, scale=1.5, transform shape] (encoder) at (13.3,10) {\scriptsize Encoder};
		\node[draw=cc1, line width=1pt, circle, minimum size=0.9cm, fill=cc1!10] at (13.8, 9.2) {};
		\node at (13.8, 9.2) {\includegraphics[width=0.7cm]{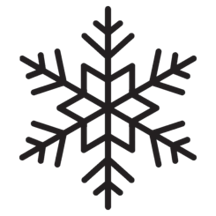}};

		\node at (13.3,8.4) {\small Foundation Model};
		
		
		\node[draw=cc2, line width=1pt, fill=cc2!10, rectangle, minimum width=1cm, minimum height=2.5cm, rounded corners] (latent) at (16, 10) {\rotatebox{90}{Latent Space}};
		
		
		\node[draw=cc3!90, rectangle, rounded corners, line width=1pt, minimum width=6cm, minimum height=4cm, align=center, fill=cc3!10] (stat) at (21, 10) {};
		\node[draw=cc3!40, rectangle, rounded corners, minimum width=2cm, minimum height=0.5cm, align=center, fill=cc3!40] at (21,12) {\small Decision Statistics};
		
		\node[draw=none, rectangle, rounded corners, minimum width=2cm, minimum height=0.5cm, align=center] at (21, 10) {\small $(\forall i) \ \hat \Omega_i := \hat s_i^T \hat \Sigma^{-1}_\lambda\,, \quad \hat \omega^T := \hat{s}^T \hat \Sigma^{-1}_\lambda$};
		
		\node[draw, rectangle, rounded corners, minimum width=5cm, minimum height=1cm, align=center, fill=white] (M) at (21, 11) {$\hat p(x) = $ Softmax$\big(\hat{\Omega} x + \hat{b}\big)$};
		
		\node[draw, rectangle, rounded corners, minimum width=5cm, minimum height=1cm, align=center, fill=white] (M2) at (21, 9) {$\hat{T}(x)=\hat \omega^T x \gtrless \tau$};

		
		\node[draw=cc5, line width=1pt, fill=cc5!10, rectangle, minimum width=1cm, minimum height=2.5cm, rounded corners] (decision) at (25.25, 10) {\rotatebox{90}{Decision}};
		
		\draw[->, line width=0.7pt, draw=black, >=Stealth] (encoder) -- (latent);
		
		\draw[->, line width=0.7pt, draw=black, >=Stealth] (16.5, 11) -- (M.west) node[midway, above, xshift=-0.2cm] {$M>2$};
		\draw[->, line width=0.7pt, draw=black, >=Stealth] (16.5, 9) -- (M2.west) node[midway, above, xshift=-0.2cm] {$M=2$};

		\draw[->, line width=0.7pt, draw=black, >=Stealth] (M.east) -- (24.75, 11);
		\draw[->, line width=0.7pt, draw=black, >=Stealth] (M2.east) -- (24.75, 9);
		
		
		\node[draw=cc2, line width=1pt, fill=cc2!10, rectangle, minimum width=7cm, minimum height=4cm, align=left, rounded corners, font=\large] (rect1) at (14, 5) {
			$\bullet$   $\cH_k : x \sim s_k + \Sigma^{1/2}z\,, \, \, s_k \in \R^p$\\[0.3cm]
			$\bullet$ \textit{High-dimensional} regime: \hspace{0.7cm} \phantom{.} \\
			\hspace{0.3cm} $n$: training data\\
			\hspace{0.3cm} $p$: dimension of the latent space\\
			\hspace{0.3cm} $p,\, n \to +\infty\,, \  p/n=\gamma$
		};
		
		\node at (16.6, 3.6) {\includegraphics[width=1.3cm]{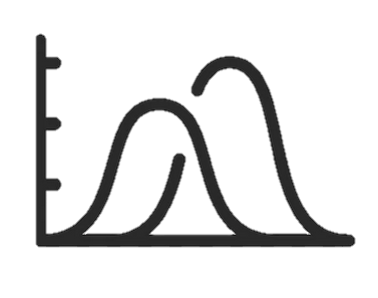}};

		\node[draw=cc5, line width=1pt, fill=cc5!10, rectangle, minimum width=7cm, minimum height=4cm, align=left, rounded corners, font=\large] (rect2) at (22.25, 5) {
			\textbf{Error decay:} \\[0.3cm]
			$\mathbb{P}(\hat k(x) \neq k \mid \mathcal{H}_k) \lesssim \exp\{ -p R^{(k)}_{\gamma, \lambda}\}$\\[0.3cm]
			$\displaystyle 
			\left\{
			\begin{aligned}
			&P_{MD} \lesssim  \exp \{ -p \textcolor{cc4}{\boldsymbol{R}_{\boldsymbol{\gamma}, \boldsymbol{\lambda}}}  \}\\
			&P_{FA} = \alpha 
			\end{aligned}\right.
			$\vspace{0.4cm}
		};
		
		\node at (24.8, 3.75) {\includegraphics[width=1.3cm]{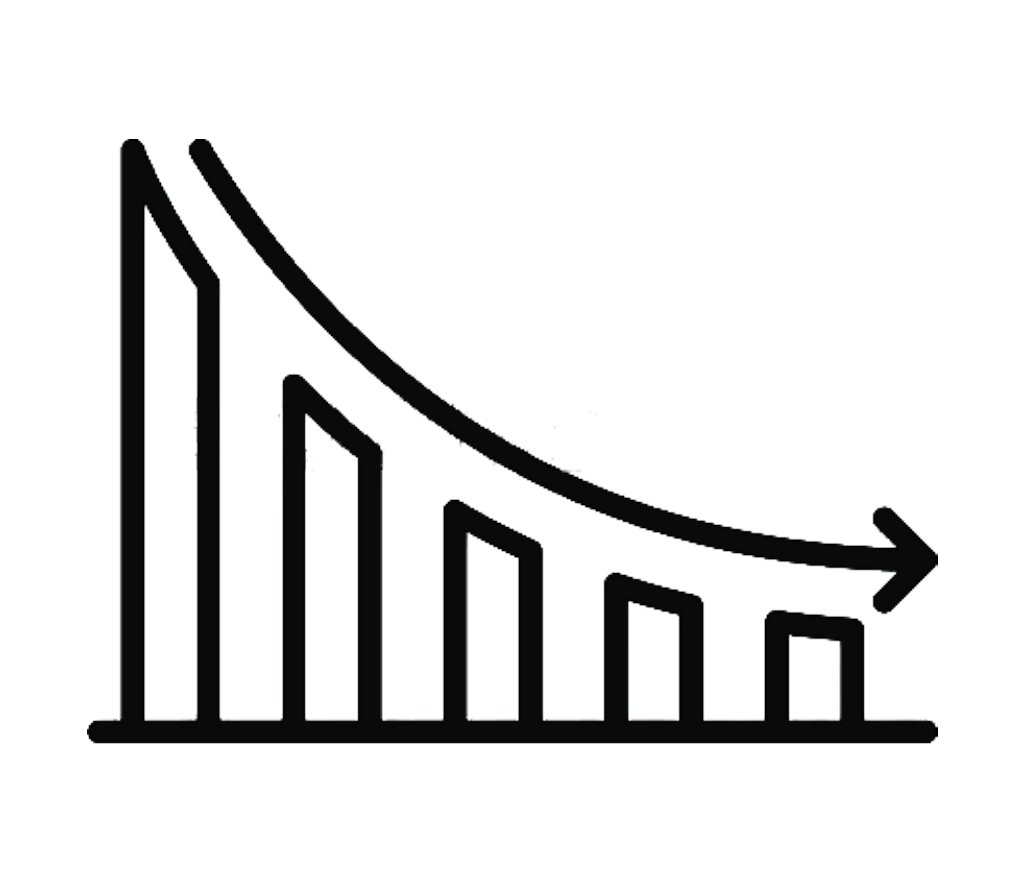}};

        \node at (24.5, 6.2) {\textcolor{cc5}{\footnotesize{$\rhd \ M > 2$}}}; 
        \node at (22.7, 4.1) {\textcolor{cc5}{\footnotesize{$\rhd \ M = 2$}}}; 
		
		\filldraw[draw=none, line width=0.5pt, fill=cc2!10] (latent.south west) .. controls (14.5, 7.5) .. (rect1.north west) -- (rect1.north east) .. controls (17.4, 7.1) .. (latent.south east) -- cycle;
		
		\filldraw[draw=none, line width=0.5pt, fill=cc5!10] (decision.south west) .. controls (24, 7.5) .. (rect2.north west) -- (rect2.north east) -- (decision.south east) -- cycle;

		
		\node[anchor=west] at (26.5, 7.5) {\includegraphics[width=9cm]{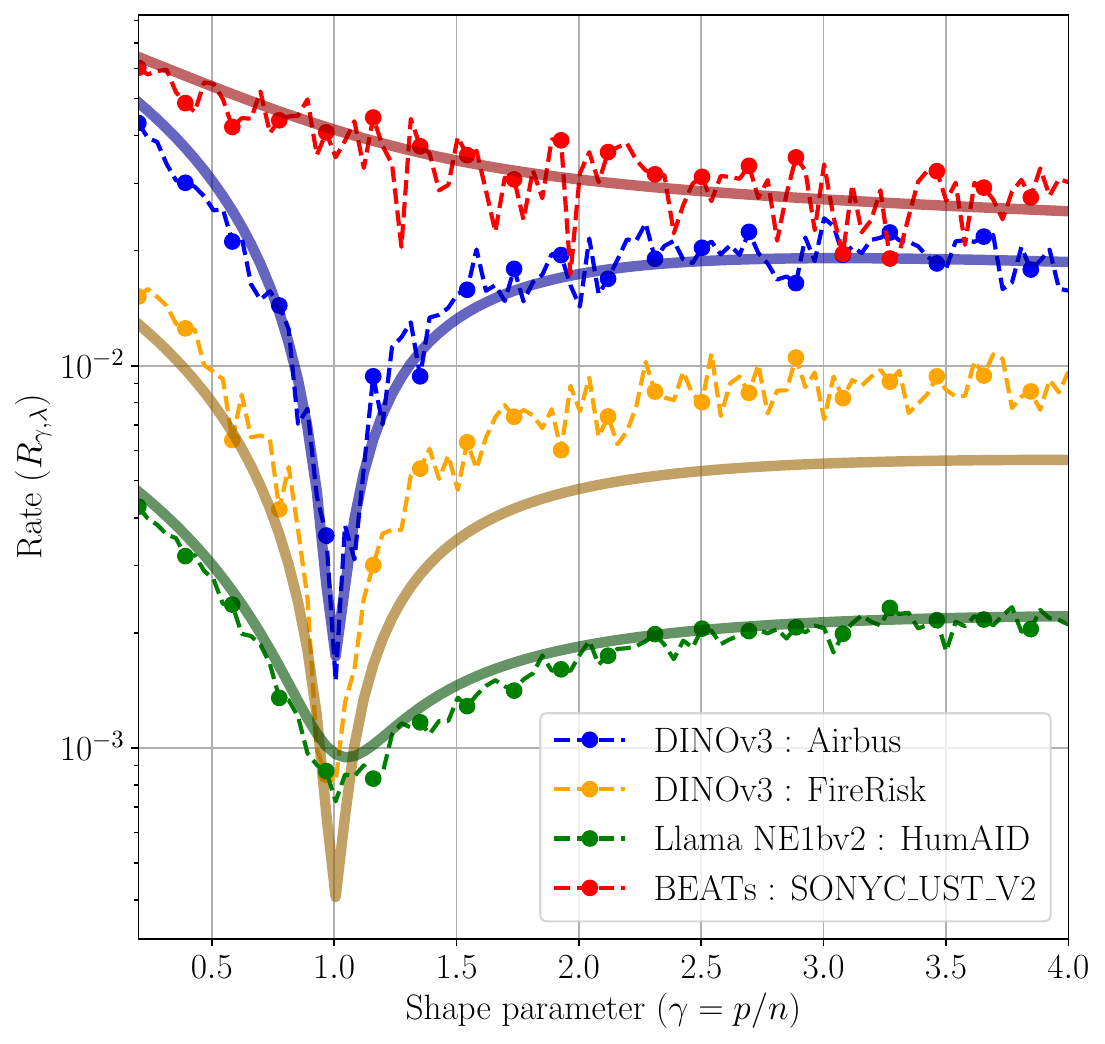}};
		
		\node[draw, rectangle, rounded corners, minimum width=9.5cm, minimum height=9cm, align=center, line width=0.8mm, draw=cc4] at (31, 7.5) {};
		
		\draw[-{Circle[fill=cc4]}, draw=cc4, line width=0.8mm] (26.25, 4.7) -- (24, 4.7);
		
		
		\end{tikzpicture}
	}
	\caption{Leveraging foundation models as abstract feature extractors to design theoretically grounded data-driven decision statistics. \textbf{Left:} Processing pipeline for the multi-class ($M > 2$) and binary ($M=2$) classification problems. The latent features $x$ are processed via a linear combination layer, whose weights are obtained from the regularized sample covariance matrix $\hat{\Sigma}_\lambda:= \hat \Sigma + \lambda I$ and the sample mean of $\cH_i$ training data $\hat{s}_i$. The decision is taken as maximum a posteriori for $M>2$ or with a threshold comparison for $M=2$.  
    \textbf{Right:} \arev{Error probability decay rate} as a function of the shape parameter for $M=2$. Comparison between empirical (dashed) and theoretical curves (solid lines) across different foundation models and datasets. We are able to predict and control the performance in terms of error probabilities when both $p$ and $n$ are large enough.}
	\label{fig:proposed_model}
\end{figure}

\section{Introduction}

Modern Machine Learning (ML) systems are increasingly driven by scale, with model size, data volume, and compute growing rapidly to improve performance, raising fundamental questions about their statistical, economic, and energy efficiency~\cite{kmhbccgrwa20,brown2020language}. \rev{Counterintuitively, large models are more sample-efficient than smaller ones, reaching the same level of performance with fewer optimization steps}~\cite{kmhbccgrwa20}.\EC{Taken from [25]. We can prove this for regression :)} This points to the central role of the model-size/sample-size ratio, a key parameter governing the emergence of nontrivial performance transitions~\cite{bhmm19,brt19,bhx20,hmrt22high,nakkiran2020deep}.

Already in the late 1980s \cite{vcr89, okkn90} (see \cite{lvmkt20} for a historical perspective) it was pointed out, both theoretically and empirically, that in several settings including random forests, neural networks and kernel methods \cite{brt19, bhx20, mm22, hmrt22high, bach24high}, the generalization error as a function of the \emph{shape parameter} $\gamma := p/n$, where $p$ is the underlying model size and $n$ is the number of data, exhibits a fascinating trend. The classical U-shaped risk curve, which initially decreases and then increases as the interpolation threshold $\gamma = 1$ is approached from below, decreases once again beyond the interpolation regime, i.e., for $\gamma > 1$---a phenomenon that is nowadays referred to as \emph{double-descent} \cite{bhmm19}.

Over the past five years several comprehensive mathematical models of double-descent have been developed, with notable contributions by Tibshirani, Hastie, Montanari, Bach, Candès, and their collaborators~\cite{sur2019modern, hmrt22high, bach24high, lpjbt24}, providing i) a detailed characterization of this intriguing phenomenon for linear regression, an important setting due to its connections to neural networks, cf.~\cite[Section 1.2]{hmrt22high}, and ii) the demonstration of the fundamental role played by \emph{random matrix theory} (RMT) and the tools therein \cite{bs10spectral, av04}. 
While double-descent is now well understood for regression problems, the impact of high dimensionality on detection and classification error probabilities remains largely unexplored. This motivates our first research question:
\begin{center}
\textit{Do detection methods exhibit a double-descent phenomenon?}
\end{center}
\rev{We answer this question in the affirmative by showing that, for two standard decision statistics, the rate governing the decay of error probabilities---a natural performance criterion in this context, dating back to the seminal work of Chernoff~\cite{chernoff1952} in the 1950s---exhibits double-descent and even multiple-descent behavior in the high-dimensional regime. For binary classification, our results complement an extensive body of research on high-dimensional logistic regression \cite{sur2019modern, salehi2019impact, deng2022model}.}\EC{Note that Deng has a work on double-descent for logistic regression (partial results)}

To\EC{I thought this rearrangement works for Reviewer 1} corroborate our theoretical findings, we conduct systematic comparisons on synthetic data and, more importantly, validate our approach on multiple real-world datasets. Since real data rarely satisfies the assumptions of \rev{our statistical setting}\AB{or we can make it more general and not attack
ourselves? real data never satisfies anything}, the proposed detection methods cannot be applied directly in the original data space. Instead, to extend our statistical framework to real data, we first map the observations into a latent space using encoders from \emph{foundation} models such as DINOv3 \cite{dinov3}, Llama-Nemotron-Embed-1B-v2 \cite{llama_nemotron}, and BEATs \cite{BEATs}. We then consider the resulting bottleneck representations, which are naturally high-dimensional, and apply the decision statistic in this abstract feature space, leading to the data-driven detection (or classification) pipeline depicted in Figure~\ref{fig:proposed_model}. This leads to our second research question:
\begin{center}
\textit{Can high-dimensional asymptotic analysis predict the performance of Neyman--Pearson}\\ \textit{decision statistics when applied in the latent space of foundation models?}
\end{center}
Our focus on foundation models is motivated by recent insights from Balestriero and LeCun \cite{bl25}. They show that, to minimize prediction risk uniformly across broad families of downstream tasks, the optimal embedding distribution produced by foundation models should be isotropic Gaussian, thereby offering an ideal testing ground for our statistical framework. We demonstrate that our approach yields results in close agreement with our theoretical predictions, both in terms of moment estimation and in overall detection performance, as shown in Figures \ref{fig:proposed_model} and \ref{fig:nebula_probability_of_detection}. To the best of our knowledge, this is the first time that high-dimensional analysis of data-driven methods is shown to closely predict performances with real-world data. Overall, our approach establishes an explainable and theoretically grounded methodology that achieves competitive empirical performance \rev{in terms of classification accuracy relative to standard baselines, such as logistic regression and linear SVM.}

\newpage

\subsection{Problem formulation}
\EC{Subsection title can be improved}
\paragraph{Binary case} %

We begin with a binary hypothesis testing problem. Given $x \in \R^p$, we test whether it is sampled from a distribution with zero mean (null hypothesis) or from a distribution with mean $s\in \R^p$. We assume that both distributions share the same non-singular covariance matrix $\Sigma \in \R^{p\times p}$ and are given by:
\begin{equation}\label{eq:intro_hypotheses_testing}
{\cal H}_0: \ x \sim \Sigma^{1/2}z \,, \quad \text{and} \quad {\cal H}_1:  \ x \sim s + \Sigma^{1/2}z\,,
\end{equation}
where $z$ is a vector of independent and identically distributed (i.i.d.) sub-Gaussian variables with zero mean and unit variance, which for simplicity we shall assume to be Gaussian\footnote{This assumption can be relaxed at several points in the analysis; we indicate explicitly whenever this is the case.}. If both $s$ and $\Sigma$ are known, the log-likelihood ratio (LLR) defines the optimal detector in the sense of Neyman--Pearson, which in this setting yields $T^\dagger(x) :=  s^T {\Sigma}^{-1} {x} = \omega^T x$~\cite{lehmann2022testing, Kay:B93_2}. As soon as $\Sigma$ and $s$ are unknown, we resort to a strategy to estimate them. There are several possible ways to address this task, one of them being to estimate $\omega$ from a training set of size $n$, yielding the \emph{data-driven decision statistic} $\hat T$, defined for a \emph{regularization parameter} $\lambda> 0$ and all $x\in \R^p$ by:
\begin{align}
	&\hat T(x) := \hat s^T  (\hat \Sigma + \lambda I)^{-1}x\,, \label{eq:def_That_linear_unknown}\\
	&\hat T(x) := x^T  (\hat \Sigma + \lambda I)^{-1}x \  \text{if no $\cH_1$ data is available in the training set}, \label{eq:def_T_quadratic}
\end{align}
where $\hat \Sigma:= \frac{1}{n}\hat X \hat X^T$ is the sample covariance matrix of the centered training set (stored in a matrix $\hat X\in \R^{p \times n}$), $\hat s$ is the sample mean of $\cH_1$ training data and $I$ is the identity matrix. The statistics reject the null hypothesis if $\hat T(x) > \tau$, for a predefined threshold $\tau\in \R$ and accept it otherwise.

\paragraph{Large deviations performance in high-dimensions} In contrast to popular approaches like logistic regression, in the Neyman--Pearson setup, we aim to minimize the probability of miss detection subject to an upper bound on the probability of false alarm, which is a crucial property in safety-critical fields like healthcare or defense. \rev{As a performance criterion, we consider} the rate in the sense of the \emph{large deviation principle} (LDP) \cite{dz10ld}. Formally, it characterizes the error probability decay when, e.g., the number of independent samples observed to take a decision diverges---a viewpoint increasingly used in recent analyses of data-driven methods \cite{bmasdw22,hbmw24}. In the proposed work, we go beyond these fixed-dimensional settings and study the \emph{high-dimensional regime}, i.e., we consider both the number of training samples $n$ and model size $p$ to grow to $+\infty$ but keeping a constant ratio:
\begin{equation}\label{eq:intro_high_dimensional_regime}
n, p \to +\infty\,, \quad \text{and} \quad \gamma = p/n \geq 0\,.
\end{equation}
For both the decision statistics \eqref{eq:def_That_linear_unknown} and \eqref{eq:def_T_quadratic}, we show that, in the high-dimensional regime 
\eqref{eq:intro_high_dimensional_regime}, 
the decision statistics $\hat T(x)$ satisfy an LDP: For $\tau_p \in \R$ such that  $P_{FA}:=\mathbb{P}\big( \hat T(x) > \tau_p \mid \hat X, \mathcal{H}_0 \big) = \alpha$ (typically small), we show that\footnote{To simplify notation, throughout the paper we omit the explicit dependence on $p$ and $n$, whenever this dependence is clear from the context. Furthermore, we write
$\displaystyle \lim_{p\to+\infty} a_{p}$ as shorthand for the high-dimensional limit
$\displaystyle \lim_{\substack{p,n\to+\infty \\ p/n=\gamma}} a_{p}$.}
\begin{equation}\label{eq:intro_ldp_hypotheses}
\lim_{p\to +\infty} \frac{1}{p} \log P_{MD} := \lim_{p\to +\infty} \frac{1}{p} \log \mathbb{P}\big( \hat T(x) \leq \tau_p \mid  \hat X, \mathcal{H}_1 \big) = -R_{\gamma, \lambda}\,,
\end{equation}
for $R_{\gamma, \lambda}>0$ named \emph{rate}, where $P_{FA}$ is the probability of false alarm and $P_{MD}$ is the probability of miss detection. We also show analytically that the rate exhibits a double-descent and even a \emph{multiple}-descent behavior in some cases, thereby giving a positive answer to our first research question. Characterizing $R_{\gamma, \lambda}$ constitutes our main technical challenge. For \eqref{eq:def_That_linear_unknown}, by linearity, the rate can be deduced from the moments of $\hat T$. In contrast, for \eqref{eq:def_T_quadratic}, we rely on more delicate arguments based on the G\"artner--Ellis Theorem \cite{gartner1977, ellis84}, and nontrivial RMT and convex analysis tools \cite{bc2017convex}.

\paragraph{Multi-hypotheses setting} The strength of our analysis for the binary case is that it immediately generalizes to a shift-in-mean \emph{multi-hypothesis} testing setting, where we aim to discriminate among $M\geq 3$ competing distributions: \begin{equation}\label{eq:intro_multi_hypotheses_testing}
\mathcal{H}_i:\quad x \sim s_i + \Sigma^{1/2} z\,, \qquad s_i \in \R^p\,, \quad i = 0, \dots, M-1\,.
\end{equation}
In this setting, the Maximum A Posteriori (MAP) rule selects the hypothesis with the largest posterior probability, which exhibits the standard linear--softmax form: $p(x) = \softmax\big(  \Omega x + b \big)$, where $\Omega_i  := s_i^T \Sigma^{-1}$, $b_i := \log \pi_i - \tfrac{1}{2}s_i^T \Sigma^{-1}s_i$, and $\Omega_i$ is the $i^{th}$ row of $\Omega$ and $\pi_i$ is the prior of $\cH_i$, see \cite[Chapter 4]{bishop2006pattern}. The corresponding data-driven rule picks $\hat k(x) :=  \argmax_{0 \le i \le M-1} \hat p_i(x)$, where
\begin{equation}
\hat p(x) : = \softmax\big(  \hat \Omega x + \hat b \big)\,, \quad \text{and} \quad \hat \Omega_i  := \hat s_i^T (\hat \Sigma + \lambda I)^{-1}\,, \ \hat b_i := \log \pi_i - \tfrac{1}{2}\hat s_i^T (\hat \Sigma + \lambda I)^{-1} \hat s_i\;,
\end{equation}\EC{Changed $W$ to $\Omega$ because we already use $W$ extensively.}%
where $\hat s_i$ is the sample mean of $\cH_i$ data, for all $i=0, \dots, M-1$. As we show in detail in Appendix \ref{sec:multi_hypotheses}, all error probabilities exhibit an exponential decay of the form:
\begin{equation}
	\limsup_{p \to +\infty} \frac{1}{p} \log \P(\hat k(x) \neq k \mid  \hat X, \cH_k ) \leq - R_{\gamma, \lambda}^{(k)}\,, \quad \text{for all} \ k = 0, \dots, M-1\,,
\end{equation}
where $R_{\gamma, \lambda}^{(k)}$ can be obtained by studying the rate of the binary classifiers defined in \eqref{eq:def_That_linear_unknown} between $\cH_k$ and all competing hypotheses. Motivated by this connection, we henceforth restrict our attention to the binary case, which nonetheless provides a rich and representative setting.

\subsection{Contributions}

We summarize our main contributions as follows:

\begin{enumerate}
	\item \textbf{High-dimensional LDP analysis:} We derive closed-form expressions of the rate in the sense of the LDP for $\hat T$ defined as in \eqref{eq:def_That_linear_unknown} and \eqref{eq:def_T_quadratic} in the high-dimensional regime \eqref{eq:intro_high_dimensional_regime}, see Theorem \ref{thm:rate_linear} and Theorem \ref{thm:quadratic_error_decay}. We show that these exhibit double-descent, and even multiple-descent behaviors in some cases. These formulas can be computed explicitly in isotropic regimes, and adapted swiftly to several mismatched scenarios, modeling a variety of realistic settings, where overparameterization might yield the best performance.
	\item \textbf{Detection in foundation models' latent space:} We demonstrate the practical utility of our decision statistics in real-world detection tasks using the pipeline shown in Figure~\ref{fig:proposed_model} ($M=2$). We show that these modern architectures naturally lie within our high-dimensional regime, making our theoretical analysis fundamental for predicting empirical performance. Finally, we validate our framework across multiple data modalities (vision, text, and audio) and foundation models, confirming the \emph{robustness} and \emph{universality} of our approach.
\end{enumerate}

\subsection{Assumptions and definitions}
The notation used throughout this paper follows standard conventions and is summarized in Appendix \ref{sec:supp_notation}. Throughout the paper we consider the following assumptions, which are in line with \cite{bach24high}.

\begin{assumption*}\label{ass:1}
	The following holds:
	\begin{enumerate}[label=A\arabic*), ref=A\arabic*]
		\item \label{ass:A1} As $p\to +\infty$, the spectral measure of $\Sigma$, denoted by $H_p$, converges to a probability distribution $H$ on $\R_+$, the so-called \emph{population spectral distribution}. Moreover, $H$ has compact support in $\R_{++}$ \arev{and $aI \preceq \Sigma \preceq bI$, for $0< a\leq b$}.
		\item \label{ass:A2} For \arev{every signal sequence} $\theta^* \in \R^p$,  $\|\theta^*\|^2=\mathcal{O}(p)$ as $p\to +\infty$, and there exists some $\nu_{\theta^*}\in \mathcal{M}(\R_+)$, such that $\frac{1}{p}\sum_{i=1}^p (v_i^T \theta^*)^2 \delta_{\xi_i} \rightharpoonup \nu_{\theta^*}$ as $p\to +\infty$, where $v_i$ is $i^{th}$ eigenvector of $\Sigma$ with unit norm associated to the eigenvalue $\xi_i$. If $\theta^* = s$ in \eqref{eq:intro_hypotheses_testing}, then $\nu_s \ll H$.
	\end{enumerate}
\end{assumption*}

The absolute continuity of $\nu_s$ with respect to $H$ in Assumption \ref{ass:A2} is not really needed in our analysis, but is imposed to ease interpretation, as it allows us to define:
\begin{defn}[Signal-to-noise ratio]\label{def:snr}
	Suppose Assumption \ref{ass:1} holds. We say that the \emph{asymptotic signal-to-noise ratio} and the \emph{asymptotic directional signal-to-noise ratio} are, respectively,
	\begin{equation}
	\SNR := \lim_{p\to +\infty} \frac{1}{p} s^T\Sigma^{-1}s\,, \quad \text{and} \quad \lSNR(\xi) := \frac{1}{\xi} \frac{\partial \nu_s}{\partial H}(\xi)\,, 
	\end{equation}
	where $\frac{\partial \nu_s}{\partial H}$ is the Radon--Nikodym derivative of $\nu_s$ over $H$.
\end{defn}
Formally, the asymptotic directional SNR, from now on referred to as $\lSNR$, encodes the limiting behavior of the function that to each eigenvalue $\xi_i$ of $\Sigma$ returns $(v_i^T s)^2 / \xi_i$, which represents the power of the signal $s$ in the direction of the eigenvector $v_i$ associated to the eigenvalue $\xi_i$, weighted by the latter, hence its name. Further, it allows us to define the following \emph{observables}.
\begin{defn}[Observables]\label{def:observables}
Under Assumption \ref{ass:1}, let for $q\in \N$ and all $\kappa\geq 0$
	\begin{equation}\label{eq:def_observables}
	\mathfrak{m}_q(\kappa):= \int_{\R_+} \left[ \frac{\xi}{\xi + \kappa}\right]^q \diff H(\xi) \quad \text{and} \quad \mathfrak{n}_q(\kappa) :=  \int_{\R_+} \left[ \frac{\xi}{\xi + \kappa}\right]^q \lSNR(\xi) \diff H(\xi)\,.
	\end{equation}
\end{defn}

These notions,  \rev{which arise naturally in our analysis as limits of trace functionals, cf.~Fact \ref{fact:supp_observables_as_trace_limits}}, admit clear formulations in the isotropic and ridgeless regimes, see Examples \ref{example:supp_isotropic_data} and \ref{example:supp_k_0}.

\paragraph{Random matrix theory} The classical result from the sixties by Marchenko and Pastur \cite{marchenko1967distribution}, and later refined by Bai and Silverstein \cite{bs10spectral} states that, under Assumption \ref{ass:1}---but not necessarily the Gaussianity assumption---in the high-dimensional regime of \eqref{eq:intro_high_dimensional_regime}, the companion Stieltjes transform of the spectral measure of $\hat \Sigma$, denoted from now on $\hat H$, almost surely converges point-wisely to a certain $\varphi$ defined as the unique solution to the \emph{Silverstein} equation:
\begin{equation}\label{eq:silverstein_fpe}
\frac{1}{ \varphi(z)} + z = \gamma \int_0^{+\infty} \frac{\xi}{1 + \xi  \varphi(z)} \diff H(\xi)\,.
\end{equation}
See Theorem \ref{thm:rmb}, due to \cite{rm11spectral, bach24high}. We will need to compute in particular, for $\lambda > 0$,
\begin{equation}\label{eq:supp_silverstein_on_k}
    \kappa(\lambda) := \varphi(-\lambda)^{-1}\,, \quad \text{satisfying} \quad \kappa(\lambda) = \lambda + \gamma \kappa(\lambda) \m_1(\kappa(\lambda))\,,
\end{equation}
which can be efficiently solved by fixed-point iterations. We refer the reader to Appendix \ref{sec:supp_rmt} for a more complete account on RMT.

\section{Linear decision statistics}\label{sec:linear}

Since $\hat T$ is linear, it inherits the Gaussianity of $x$, yielding $\hat T(x) \sim {\cal N}(0, \hat \sigma^2 )$ under $\cH_0$, and $\hat T(x) \sim {\cal N}(\hat \mu, \hat \sigma^2)$ under $\cH_1$, where $\hat \mu := \E [\hat T(x) \mid  \hat X, {\cal H}_1 ]$, and $\hat \sigma^2 := \Var [\hat T(x) \mid  \hat X, {\cal H}_1 ]$. Being $\hat T$ Gaussian, the probability of miss detection with fixed probability of false alarm vanishes exponentially with $p$, with a rate depending on the \emph{deflection} $\hat d := \hat 
\mu / \hat \sigma$, see Example \ref{example:supp_ldp_gaussian}:
\begin{equation}\label{eq:def_rate_linear}
\lim_{p\to +\infty} \frac{1}{p} \log P_{MD} = -R_{\gamma, \lambda}\,, \quad  R_{\gamma, \lambda}  :=  \lim_{p\to +\infty}  \hat d^2 / (2p)\,,
\end{equation}
whenever the above almost sure limits exist. We refer to Appendix \ref{app:supp_preliminaries_LDP} for a general account of LDPs for non-Gaussian statistics, \rev{and to Appendix \ref{app:supp_calibrate_errors_nongaussian} for the calibration of error probabilities in this setting}. Clearly, the same result holds switching the role of $P_{FA}$ and $P_{MD}$. Observe in particular that $R_{\gamma, \lambda}$ can be obtained from the asymptotic behaviors of $\hat \mu$ and $\hat \sigma^2$.

\begin{prop}\label{prop:mean_var_linear}
	Suppose that Assumption \ref{ass:1} holds, let $\kappa(\lambda)$ be defined as in \eqref{eq:supp_silverstein_on_k}, and $\mathfrak{n}_q$, $\mathfrak{m}_q$, $q = 1, 2$ be the observables defined in Definition \ref{def:observables}. Then, as $p, n\to+\infty$ with $p/ n = \gamma$, $\hat \mu / p$ and $\hat \sigma^2/p$ converge almost surely to $\mu_\infty$ and $\sigma^2_\infty$ defined by 
	\begin{equation}\label{eq:mean_var_linear}
 	\mu_{\infty} :=  \frac{\kappa(\lambda)}{\lambda} \n_1(\kappa(\lambda))\,, \quad \text{and}\quad 
	\sigma_{\infty}^2:= \frac{\kappa(\lambda)^2}{\lambda^2 } \frac{\n_2(\kappa(\lambda))}{\big(1 - \gamma \m_2(\kappa(\lambda))\big) }\,.
	\end{equation} 
\end{prop}

A similar result, but in a different statistical setting, has been shown by Dobriban and Wager in \cite{dobriban2018high}.

While the Gaussian assumption in \eqref{eq:intro_hypotheses_testing} is not required for Proposition \ref{prop:mean_var_linear}, \rev{cf.~Remark \ref{rmk:beyond_gaussianity_prop}}, we retain it to simplify the derivation of the LDP rate (see Appendix \ref{app:supp_preliminaries_LDP}):

\begin{thm}\label{thm:rate_linear}
	Suppose that Assumption \ref{ass:1} holds, let $\kappa(\lambda)$ be defined as in \eqref{eq:supp_silverstein_on_k}, and $\mathfrak{n}_q$, $\mathfrak{m}_q$, $q = 1, 2$ be the observables defined in Definition \ref{def:observables}. Let $\hat T$ be defined as in \eqref{eq:def_That_linear_unknown}. Then, \eqref{eq:def_rate_linear} holds with
	\begin{equation}\label{eq:rate_linear_general}
	R_{\gamma, \lambda} := \frac{1}{2}\frac{\n_1(\kappa(\lambda))^2}{\n_2(\kappa(\lambda))} \Big(1 - \gamma \m_2(\kappa(\lambda))\Big)\,.
	\end{equation}
\end{thm}

\begin{figure}[t]
	\centering
	\begin{subfigure}{0.43\linewidth}
		\includegraphics[width=\linewidth]{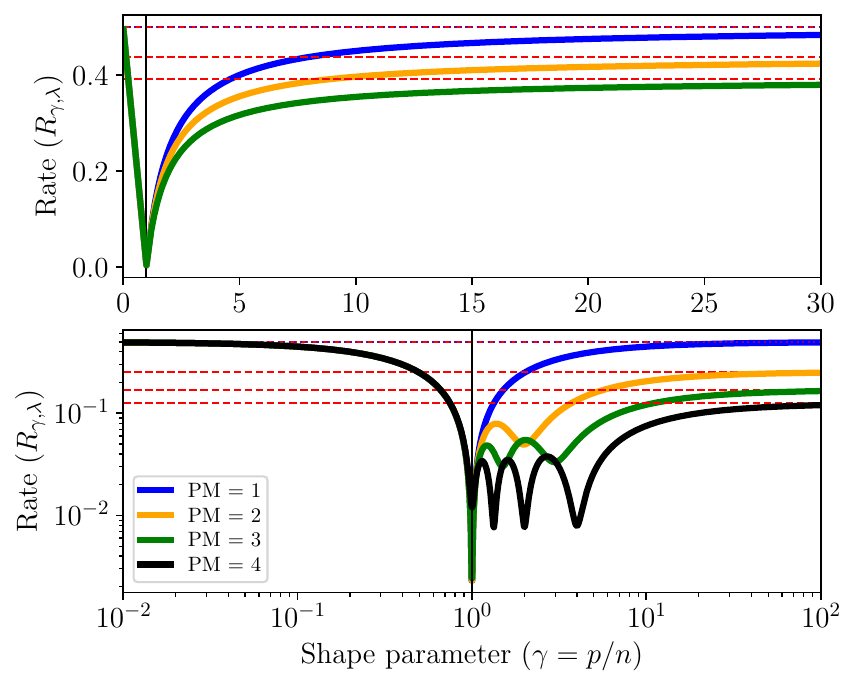}
	\caption{Performances as a function of $\gamma$.}
	\label{fig:rate_linear_examples}
	\end{subfigure}
	\begin{subfigure}{0.52\linewidth}
		\includegraphics[trim={0cm 0cm 7cm 0cm}, clip, width=1\textwidth]{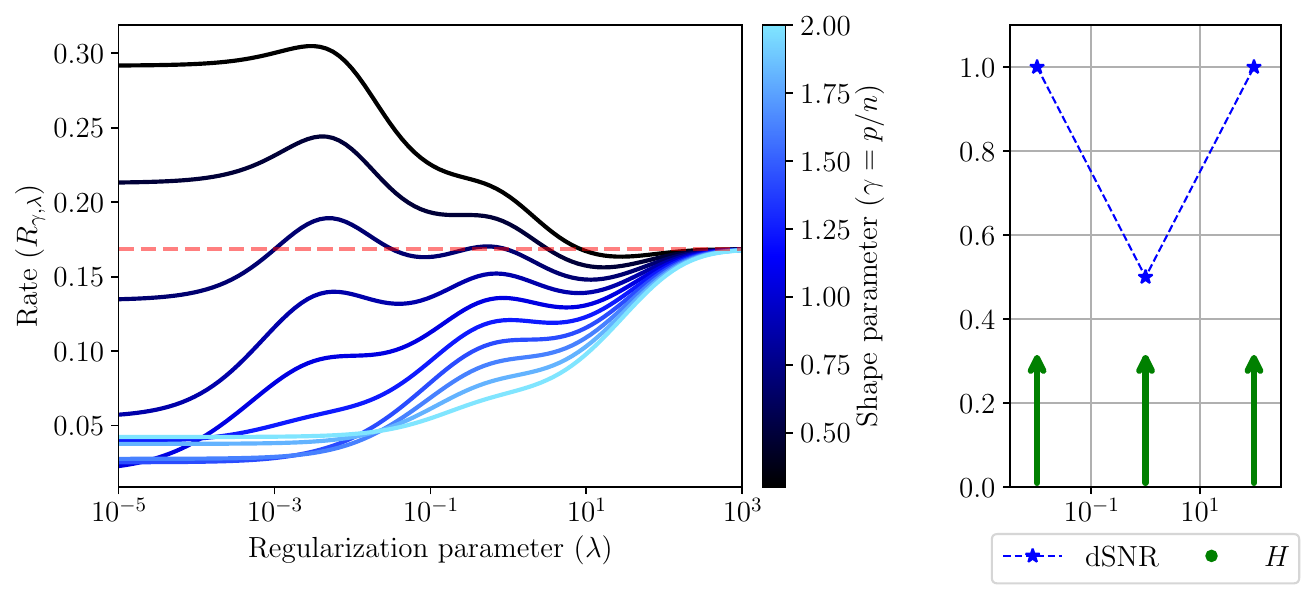}
		\caption{Performances as a function of $\lambda$.}
		\label{fig:rate_linear_vs_lam}
	\end{subfigure}
	\caption{Rate $R_{\gamma, \lambda}$ according to \eqref{eq:rate_linear_general} for the decision statistic defined in \eqref{eq:def_That_linear_unknown}. The asymptotic values for $\gamma\to0^+$ and $\gamma\to +\infty$ according to \eqref{eq:supp_def_R0_RInf} are represented with dotted blue and dashed red lines, respectively. \textbf{Left:} In the first row, we compare the rate in one isotropic setting with the rates in two non-isotropic ones with $\lSNR$ constant equal to $1$ and $\lambda=10^{-5}$. In the second row, we compare four settings with discrete $H$ with \rev{$\text{PM} \in \N$ point masses} spaced uniformly in logarithmic scale and again $\lSNR$ constant equal to $1$ and $\lambda=10^{-8}$, producing a \emph{multiple-descent} behavior. \textbf{Right:} Rate $R_{\gamma, \lambda}$ as a function of $\lambda$ for $10$ different values of $\gamma$. The precise definitions of the corresponding \arev{population spectral distributions} are detailed in Section \ref{sec:linear_missing_proofs}, and in Figure \ref{fig:supp_simulated_data_geometry}.}
	\label{fig:simulated_data}
\end{figure}

We show the dependence of $R_{\gamma, \lambda}$ as a function of $\gamma$ and $\lambda$ in Figure \ref{fig:simulated_data}. In particular, from Figure \ref{fig:rate_linear_examples} we observe that after the critical threshold $\gamma=1$, the performances start improving again, reflecting the \emph{double-descent}\footnote{Strictly speaking, we should refer to this as \emph{double-ascent} in this context (the higher the rate, the better for the performances). More precisely, in Figure \ref{fig:rate_linear_examples} we do not strictly observe a \emph{double}-ascent, but a single ascent after the critical threshold $\gamma =1$. This is expected and indeed occurs even for linear regression models \cite{hmrt22high}. We can recreate the popular $U$-shaped curves in mismatched cases, as we discuss in Appendix \ref{sec:supp_linear_miss}.} phenomenon, and giving a positive answer to our first research question. \rev{Additionally, the double-descent phenomenon persists in the multi-class setting, cf.~Figure \ref{fig:multi_class_results}.} More surprisingly, we also observe an interesting multiple-descent behavior, which is emphasized when the limiting spectral distribution $H$ has support with point masses spaced uniformly in logarithmic scale. In Figure \ref{fig:rate_linear_vs_lam}, we display the rate as a function of $\lambda$ for $10$ different values of $\gamma$ ranging from $0.5$ to $2$ and spaced uniformly. We observe that if $\gamma > 0$ and is small enough, there are local maxima suggesting an optimal tuning of the regularization parameter $\lambda$. \rev{See also Figures~\ref{fig:supp_linear_miss_general_gamma} and~\ref{fig:supp_linear_miss} for further illustrations of the sensitivity of the rate to $\lambda$.}

\paragraph{Insights and properties} In Appendix \ref{sec:supp_linear_isotropic} and \ref{sec:supp_linear_ridgeless} we investigate two notable settings in which computations can be performed in closed form---the case of \emph{isotropic} data, and the \emph{ridgeless} regime, see Proposition \ref{prop:supp_linear_rate_isotropic} and Proposition \ref{prop:supp_linear_rate_ridgeless}, respectively. Specifically, using the asymptotic behavior $\kappa(\lambda) \approx \lambda + \gamma \E_H(\xi)$ as $\gamma \to +\infty$, cf.~Lemma \ref{lem:limits_of_k_gamma}, we get in the isotropic regime:
\begin{equation}\label{eq:limits_rate_isotropic}
\tfrac{1}{2}\SNR = \lim_{\gamma \to 0^+} R_{\gamma, \lambda} = \lim_{\gamma \to + \infty} R_{\gamma, \lambda}\,,
\end{equation}
and $R_{\gamma, \lambda}$ has a minimum in $(1, \infty)$. While the first limit in \eqref{eq:limits_rate_isotropic} is well-known \cite{Kay:B93_2}, the fact that even with $\gamma\to +\infty$ the rate tends to the optimal value is a new interesting fact. In general, however, it turns out that \eqref{eq:limits_rate_isotropic} is usually not the case:

\begin{prop}\label{lem:dispersion_index}
	Suppose Assumption \ref{ass:1} holds, and let $R_{\gamma, \lambda}$ be given by Theorem \ref{thm:rate_linear}. Then
	\begin{equation}
	\lim_{\gamma \to 0^+} R_{\gamma, \lambda} \geq  \lim_{\gamma \to +\infty} R_{\gamma, \lambda}\,,
	\end{equation}
	and equality holds if and only if \arev{$\lSNR \diff H$ is supported on a single eigenvalue}.
\end{prop}
However, as we show in Appendix \ref{sec:supp_linear_miss}, as soon as the statistical properties of the validation set differ from that of the training set, overparameterization might actually yield the best performances.

\section{Quadratic decision statistic}\label{sec:quadratic}

In this section, we consider the decision statistic \eqref{eq:def_T_quadratic}. In this case, asymptotic means and variances under $\cH_0$ and $\cH_1$, denoted by $\hat \mu_{j} :=\E[ \hat T(x) \mid \hat X, \cH_{j}]$ and $\hat \sigma_{j}^2 := \E[(\hat T(x) - \hat \mu_{j})^2 \mid \hat X, \cH_{j}]$, \arev{$j=0, 1$}, can be computed as follows. 
\begin{prop}\label{prop:mean_var_quadratic}
	Let $\hat T$ be the decision statistic defined in \eqref{eq:def_T_quadratic}, and $\hat \mu_{j}$, $\hat \sigma_{j}^2$ be its mean and variance \arev{under $\cH_j$, $j =0, 1$}. Then, $\hat \mu_{j}/p$ and $\hat \sigma_{j}^2/p$ converge almost surely \arev{to the following values:}
	\begin{equation}\label{eq:mean_var_quadratic_H0}
	\begin{aligned}
	&\mu_{0, \infty} := \frac{\kappa(\lambda)}{\lambda } \m_1(\kappa(\lambda))\,, \quad \sigma_{0, \infty}^2 := \frac{\kappa(\lambda)^2}{\lambda^2} \frac{2 \m_2(\kappa(\lambda))}{1- \gamma \m_2(\kappa(\lambda))}\,, \\
	& \mu_{1, \infty}:= \mu_{0, \infty} + \frac{\kappa(\lambda)}{\lambda} \n_1(\kappa(\lambda))  \,, \quad \sigma_{1, \infty}^2:= \sigma^2_{0, \infty}+ \frac{\kappa(\lambda)^2}{\lambda^2 } \frac{ 4 \n_2(\kappa(\lambda))}{1- \gamma \m_2(\kappa(\lambda))}\,.
	\end{aligned}
	\end{equation}
\end{prop}

The main challenge in computing the performances in the sense of the LDP in this case is that $\hat T$ is no longer Gaussian and the sole knowledge of \eqref{eq:mean_var_quadratic_H0} does not allow us to derive the rate as in Section \ref{sec:linear}. Instead, we leverage the G\"artner--Ellis Theorem \cite{gartner1977}, which states that under certain conditions, \arev{the exponential decay of error probabilities is governed by} the \emph{Fenchel--Legendre conjugate} of the limiting log-moment generating function (LMGF):
\begin{equation}\label{eq:def_Lambda}
\Lambda^{(0, 1)}(\tau) = \lim_{p\to + \infty} \frac{1}{p} \log \E\big[ \exp\{\tau \hat T(x)\} \mid \hat X, \cH_{0, 1} \big]\,.
\end{equation}
Using an integration trick proposed by Bai and Silverstein in \cite{bs10spectral}, we show that $\Lambda^{(0)}$ can be written as an integral function with respect to the \emph{first branch} of the Stieltjes transform of the limiting spectral distribution of \arev{$\hat W^{-1}:= \tfrac{1}{n}\hat Z \hat Z^T + \lambda \Sigma^{-1}$, with $\hat Z:= \Sigma^{-1/2}\hat X$}, which we denote by $\nu_{\lambda, \gamma}$. Specifically, in Corollary \ref{cor:supp_lmgf_integral_formula}, we show that, \arev{under an additional \emph{no-spectral-outliers} assumption (Assumption \ref{ass:A3}),}
\begin{equation}\label{eq:def_first_branch_of_s}
\Lambda^{(0)}(\tau) = \frac{1}{2}\int_{0}^{2\tau}\underline{\mathfrak{s}}(z) \diff z\,,  \quad \text{for all} \ \arev{\tau \neq \frac{T}{2}}\,, \quad \text{where} \quad \underline{\mathfrak{s}}(\tau) := \begin{cases}
\mathfrak{s}(\tau) & \text{if} \ \tau \leq T\,,\\
+\infty & \text{else}\,,
\end{cases}
\end{equation}
and $T := \inf \supp \nu_{\lambda, \gamma}$, while $\mathfrak{s}\colon \C_+ \to \C_+$ is its Stieltjes transform, characterized by \cite[Theorem 1.1]{sb95empirical} as being the unique solution to the following fixed-point equation:
\begin{equation}\label{eq:silverstein_fpe_lmgf}
\mathfrak{s}(z) = \mathfrak{s}_{\dRNR}\bigg(z - \frac{1}{1 + \gamma \mathfrak{s}(z)}\bigg)\,, \quad z \in \C_+\,,
\end{equation}
where $\mathfrak{s}_{\dRNR}$ is the Stieltjes transform of $\dRNR$, i.e., the limiting spectral distribution of $\lambda \Sigma^{-1}$. Under $\cH_1$, the function $\Lambda^{(1)}$ can be computed from $\Lambda^{(0)}$ with a shift, see Lemma \ref{lem:supp_lmgf_H1}. To compute the Fenchel--Legendre conjugate $I^{(0)}:=(\Lambda^{(0)})^*$, we can leverage its explicit expression and write the rate function in terms of the (pseudo) inverse Stieltjes transform of $\underline{\mathfrak{s}}$, which is defined by
\begin{equation}\label{eq:pseudo_inverse_Stieltjes}
\underline{\mathfrak{s}}^{[-1]}(u) := \inf \big\{ z \in \R \mid \underline{\mathfrak{s}}(z) \geq u \big\}\,, \quad \text{for all} \ u \in \R\,.
\end{equation}
\arev{The explicit rate-function $I^{(0)}$ is given in Proposition~\ref{prop:quadratic_rate_function_H0}. This yields an explicit bound on the decay of the error probabilities $P_{FA}:= \P\big(\hat T(x) > \tau_p \mid \hat X, \cH_0\big)$ or $P_{MD}:= \P\big(\hat T(x) \leq  \tau_p \mid \hat X, \cH_1\big)$ when the alternative is (asymptotically) fixed:
\begin{thm}\label{thm:quadratic_error_decay}
Suppose Assumptions~\ref{ass:1} and~\ref{ass:A3} hold. For $\beta \in (0, 1)$, choose the threshold $\tau_p = \hat \tau_1 := \hat \mu_1 + \hat \sigma_1 Q^{-1}(1-\beta)$. Then, almost surely, $P_{MD} \to \beta$ and
\begin{equation}\label{eq:quadratic_np_fa_rate}
\limsup_{p\to\infty}\frac1p \log P_{FA}
\leq -R^{(0)}_{\gamma,\lambda}\,,\qquad
R^{(0)}_{\gamma,\lambda}:=I^{(0)}(\mu_{1,\infty})
=\frac12\int_{\mu_{0,\infty}}^{\mu_{1,\infty}}\underline{\mathfrak s}^{[-1]}(u) \diff u \,,
\end{equation}
where $\hat \mu_1, \hat \sigma_1, \mu_{0, \infty}$ and $\mu_{1, \infty}$ are given in Proposition \ref{prop:mean_var_quadratic}.
\end{thm}}\EC{Rearranged the results here. Explicit expression for $I^{(0)}$ is moved into a separate prop. Theorem is about decay of error probabilities. Same style of Thm \ref{thm:rate_linear}}
Whenever $\dRNR$ is \emph{sharp}, i.e., $\mathfrak{s}_{\dRNR}(\tau)\to +\infty$ for $\tau\uparrow \inf \supp \dRNR$, which is the case, e.g., for isotropic data, we can easily compute $\underline{\mathfrak{s}}^{[-1]}$ in \eqref{eq:pseudo_inverse_Stieltjes} leveraging \eqref{eq:silverstein_fpe_lmgf}, obtaining i) explicit expressions for isotropic data, and ii) accurate numerical approximations of \arev{$I^{(0)}$} for non-isotropic data. We refer to Appendix \ref{sec:supp_computing_the_rate_function} for further details, and to Remark \ref{rmk:supp_decay_under_H1} \arev{for the $\cH_1$ case, i.e., the decay of $P_{MD}$ with fixed $P_{FA}$, which follows from the $\cH_0$ case}. The lower bound is notoriously trickier \cite{dz10ld}, and has to be analyzed on a case-by-case basis, see Proposition \ref{prop:supp_rate_quadratic_isotropic} for a case where the LDP holds, i.e., the limsup is a limit and the inequalities are equalities. Interestingly, even in this case, the rate function exhibits multiple-descent behavior, cf.~Figure \ref{fig:supp_quadratic_multidescent}. 

\section{Neyman--Pearson detection in foundation models’ latent space}\label{sec:neyman_pearson_latent}

Detection techniques designed for \eqref{eq:intro_hypotheses_testing} are not directly applicable to real structured data. To deal with this issue and motivated by \cite{bl25}, we instead map the observations into a latent space using encoders from foundation models and apply the decision statistic in the resulting representations, leading to the framework depicted in Figure \ref{fig:proposed_model}. The key properties we establish are as follows:
\begin{itemize}
	\item \textbf{Near-optimal detection at fixed false alarm:} As $\hat T$ approximates the Bayes's decision rule under Gaussian data, we show that our approach outperforms the more standard fine-tuning of logistic regression \rev{and linear SVM} classifiers when we seek to maximize the probability of detection subject to a fixed probability of false alarm constraint.
	\item \textbf{Predictable performances:} We show that both the LDP rate as well as the probability of detection closely match their theoretical predictions, thereby providing solid theoretical foundations for our detection methodology.
	\item \textbf{Data and model agnostic approach:} We validate our approach across multiple foundation models and real-world datasets, showing consistent performance across vision, text, and audio modalities, and across DINOv3, Llama-Nemotron, and BEATs encoders. 
\end{itemize}

\begin{figure}[t]
	\centering
	\begin{subfigure}[t]{.44\linewidth}
		\includegraphics[width=\linewidth]{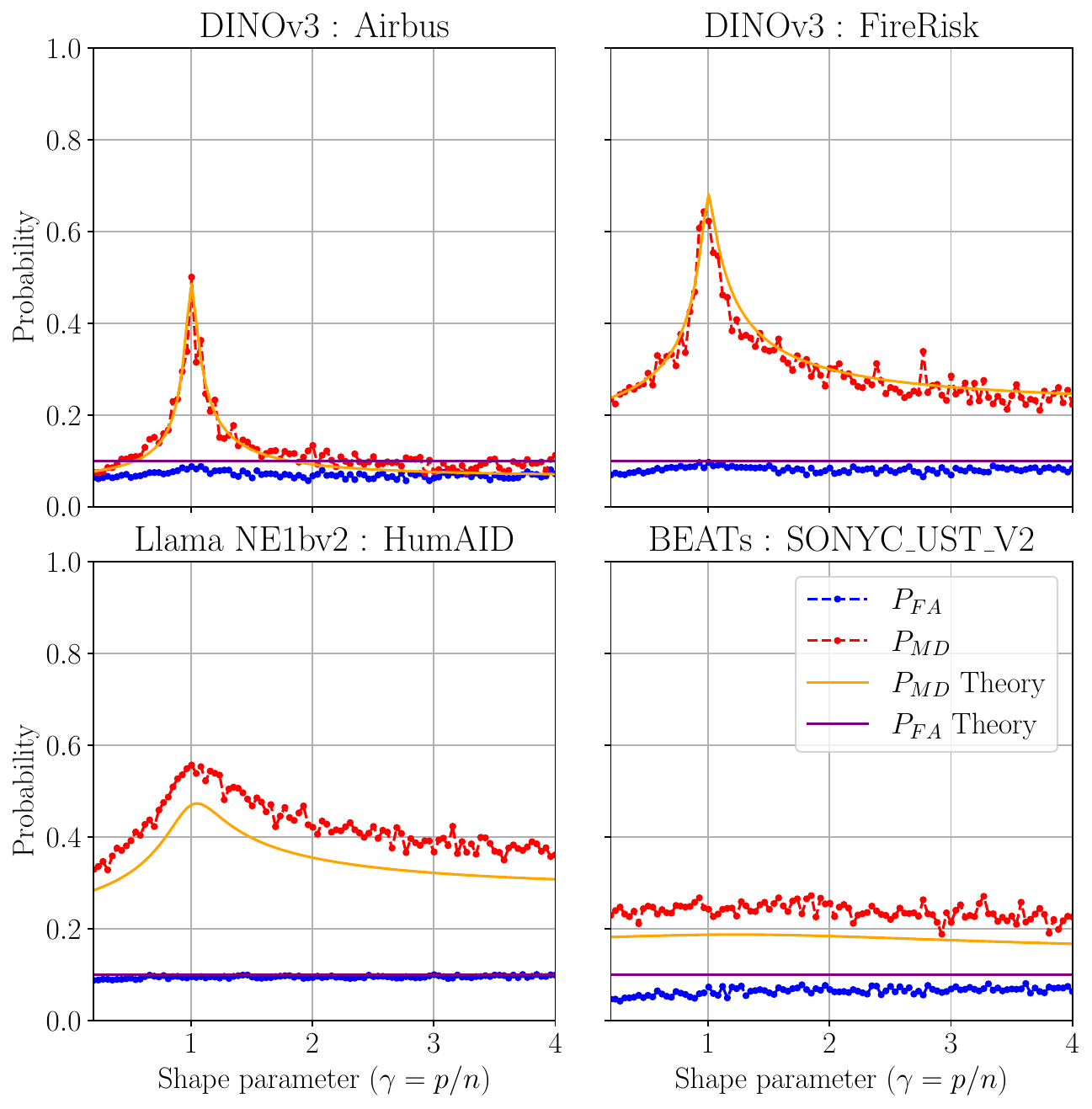}
		\caption{Theoretical and empirical $P_{FA}$ and $P_{MD}$.}
		\label{fig:nebula_probability_of_detection}
	\end{subfigure}
	\begin{subfigure}[t]{.45\linewidth}
		\includegraphics[width=\linewidth]{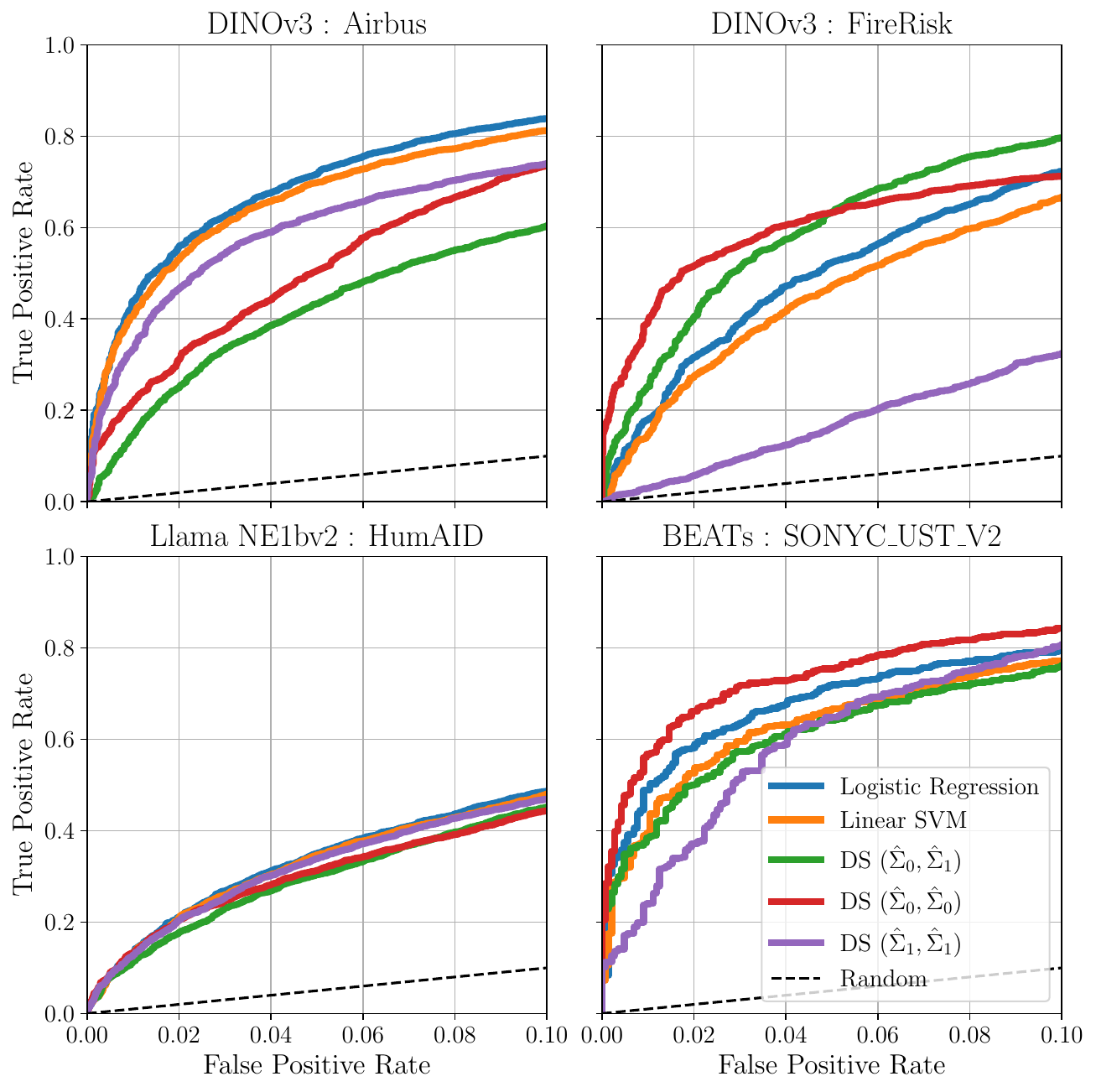}
		\caption{ROC curves.}
		\label{fig:nebula_roc}
	\end{subfigure}
	\caption{\textbf{Left:} Comparison between theoretical and empirical $P_{FA}$ and $P_{MD}$ with desired $P_{FA} =0.1$. \textbf{Right:} ROC curve comparison on small $P_{FA}$ values with benchmark classifier based on logistic regression \rev{and linear SVM}. Please refer to Appendix \ref{sec:supp_experimental_design} for details on the legend and to Figure \ref{fig:supp_nebula_roc} for the full ROC curves both in linear and logarithmic scale.}
	\label{fig:results_nebula_nips}
\end{figure}

\subsection{Experimental design}\label{sec:experimental_design} 

We now summarize our main experimental pipeline. \rev{Although our analysis in Section \ref{sec:linear} assumes the same covariance matrix under the two hypotheses, the experimental data exhibit covariance mismatch. Our theoretical predictions can be extended to account for this mismatch and are used to fit the empirical results, as detailed in Appendix \ref{sec:supp_covmismatch_1}, together with further implementation details. See also Appendix \ref{sec:supp_covmismatch_2} for an alternative approach to tackle the covariance mismatch.}

First, we employ a pre-trained foundation model with latent space of dimension $p\in \N$ and a balanced dataset for binary classification divided into a training set of size $\hat n\in \N$ and a validation set. We range across different values of $\gamma$ by truncating the training set to a given parameter $n < \hat n$. This way, by decreasing $n$ we reach potentially large $\gamma = p / n$, at the cost of i) losing statistical meaningfulness for very large $\gamma$s and ii) accepting to not be able to inspect $\gamma$ smaller than $\hat \gamma := p / \hat n$. We use the full dataset to estimate the \arev{population spectral distribution} $H$ and $\lSNR$.

Once these are fixed, for each foundation model and each dataset, we run the following pipeline. We fix $\lambda > 0$; then, we vary $\gamma > \hat \gamma$ up to a fixed threshold, and, for each $\gamma$: We utilize the truncated training set to compute the covariance matrix $\hat \Sigma$ and the estimate of the signal $\hat s$ on $\cH_1$ training data, which we then use to form $\hat T(x):=\hat s^T \hat \Sigma^{-1}_\lambda x$. Next:
\begin{enumerate}
	\item\label{item:nebula_compute_mean_rate} We use Proposition \ref{prop:mean_var_linear} to compute $\mu_{\infty}$, $\sigma^2_{\infty}$ and the rate function, and compare them to the corresponding empirical ones estimated on the validation set. We adapt our formulas to account for covariance mismatch between the two hypotheses, as detailed in Appendix~\ref{sec:supp_experimental_design}, \rev{at essentially no additional computational cost}. The results are presented in Figure \ref{fig:proposed_model} for the rate and Figure \ref{fig:supp_nebula_aggregate_moments} for mean and variance.
	\item\label{item:nebula_gauss_prob_of_detection_and_histograms} We derive, for Gaussian data, the probability of detection with fixed false alarm and compare it with its empirical counterpart, which strongly depends on the underlying distribution. The results are provided in Figure \ref{fig:nebula_probability_of_detection}.
	\item\label{item:nebula_roc} We compare the decision statistic \eqref{eq:def_That_linear_unknown} with logistic regression \rev{and linear SVM} classifiers. We further consider the decision statistics obtained by approximating the LLR assuming different covariance matrices, as specified in Appendix \ref{sec:supp_covmismatch_2}. The results are presented in Figure \ref{fig:nebula_roc}.
\end{enumerate}

For each point, we discuss the results and how they support our claims regarding the key properties introduced at the beginning of Section~\ref{sec:neyman_pearson_latent}. Since the experiment is identical across models and datasets, we report aggregate results only. We fix $\lambda=10^{-3}$ across all different cases, although the behavior is consistent with other choices. 

\paragraph{Foundation models and data} We evaluate our methodology across diverse modalities and models. First, we utilize two optical earth observation datasets: The Airbus Ship Detection Challenge (Airbus) \cite{airbus_dataset} and a fire risk assessment dataset (FireRisk) \cite{fire_risk_dataset}, both paired with DINOv3 \cite{dinov3} for binary classification. For textual data, we employ the HumAID \cite{humaid_dataset} dataset of human-annotated disaster incidents from Twitter, using Llama-Nemotron-Embed-1B-v2 \cite{llama_nemotron} for feature extraction. Finally, we assess audio data through the SONYC-UST-V2 \cite{SONYC-UST-V2} dataset, utilizing the BEATs \cite{BEATs} model for feature extraction.  We refer the reader to Appendix \ref{sec:supp_models_and_data} for a description of the datasets and models.

\subsection{Results and discussion}

In this section, we discuss the results of our numerical experiments listed in Section \ref{sec:experimental_design}. First, in Appendix \ref{sec:supp_experimental_design}, in Figure \ref{fig:supp_nebula_marginals}, we present four randomly selected marginals of the training $\cH_0$ data mapped into the latent space. As we can observe, the data appear to perfectly match Gaussian distributions, supporting the insights of \cite{bl25}, and strengthening the foundations of our approach.

\textbf{Point \ref{item:nebula_compute_mean_rate}:} In Figure~\ref{fig:proposed_model}, we show a comparison of the theoretical and empirical rate functions across all different cases. This figure demonstrates the close agreement between the theoretical formulas derived in Section~\ref{sec:linear} and the corresponding empirical behaviors, even when $H$ and $\lSNR$ are only approximately estimated with $\gamma \neq 0$. In particular, it shows an overall good match, with only a small vertical shift observed for DINOv3 applied to the FireRisk dataset, which we believe might be due to an inaccurate estimation of $H$ or $\lSNR$, \rev{cf.~Figure \ref{fig:supp_stability_gamma}}.

\textbf{Point \ref{item:nebula_gauss_prob_of_detection_and_histograms}:} Figure~\ref{fig:nebula_probability_of_detection} illustrates that the probabilities of detection at fixed false-alarm levels---which depend strongly on the Gaussianity of the underlying distribution---closely follow the theoretical predictions. Again, the only noticeable mismatches (e.g., for Llama and BEATs) appear as vertical shifts, while the overall shape of the curves remains well captured. Overall, Figure \ref{fig:nebula_probability_of_detection} shows that this does not significantly compromise the detection probability estimate with fixed false alarm levels.

\textbf{Point \ref{item:nebula_roc}:} Eventually, we conduct a systematic comparison \rev{with standard baselines}. In Figure \ref{fig:nebula_roc}, we present the ROC curves for the three approximate LLRs and \rev{logistic regression as well as linear SVM classifiers}. We can observe in particular that the approximate LLRs always achieve excellent performances (with basically no fine-tuning), except when the covariance matrices under $\cH_0$ and $\cH_1$ are assumed equal and estimated from $\cH_1$ training data. Notably, we can observe that the LLRs are systematically better than \rev{the baselines} for small false alarm levels and achieve comparable (or better) performance for larger values.

Finally, our numerical experiments demonstrate the three key properties of our detection methodology, yielding a theoretically grounded near-optimal detector in the sense of Neyman--Pearson (Point \ref{item:nebula_roc}), with predictable performances (Points \ref{item:nebula_compute_mean_rate} and \ref{item:nebula_gauss_prob_of_detection_and_histograms}), applicable to a broad range of problems. 

\paragraph{Limitations} The main limitation of our approach is the potential difficulty in estimating $H$ and $\lSNR$, which may be intractable in some settings. Moreover, cases beyond Assumption \ref{ass:1}, particularly those featuring degenerate covariance matrices, are not covered by the current analysis. \rev{Extending the results to this practically relevant setting, as conventional ML architectures may produce degenerate Gaussian distributions, appears feasible provided that $\lambda$ remains bounded away from zero, although further theoretical developments would be required.}

\section{Conclusions}

Our findings bridge the gap between high-dimensional probability theory and the practical use of foundation models for detection and classification tasks. We develop a high-dimensional statistical framework that accurately predicts the performance of data-driven decision rules applied to foundation-model representations across diverse modalities, including vision, text, and audio. These results support high-dimensional asymptotics as a useful tool for understanding and engineering robust detection systems based on modern foundation models. \rev{Among other promising directions for future research, we aim to characterize the LDP rate for a broader family of non-Gaussian statistics, derive tighter error bounds in the multi-class setting, and investigate the asymptotic regime in which the number of classes grows jointly with $p$ and $n$.}

\begin{ack}
This work was supported by NATO Allied Command Transformation (ACT) under the Seabed-to-Space Situational Awareness (S3A) research programme. We would like to thank Samantha Dugelay, Rosario Di Carlo, and Davide Cortesi for fruitful discussions related to this work.
\end{ack}

\bibliographystyle{plain}
\bibliography{biblio}

\begin{thebibliography}{10}

\bibitem{humaid_dataset}
Firoj Alam, Umair Qazi, Muhammad Imran, and Ferda Ofli.
\newblock Humaid: Human-annotated disaster incidents data from twitter.
\newblock In {\em Proceedings of the Fifteenth International AAAI Conference on
  Web and Social Media}, https://huggingface.co/datasets/QCRI/HumAID-all, 2021.
  AAAI.

\bibitem{bach24high}
Francis Bach.
\newblock High-dimensional analysis of double descent for linear regression
  with random projections.
\newblock {\em SIAM Journal on Mathematics of Data Science}, 6(1):26--50, 2024.

\bibitem{by93limit}
Z.~D. Bai and Y.~Q. Yin.
\newblock {Limit of the Smallest Eigenvalue of a Large Dimensional Sample
  Covariance Matrix}.
\newblock {\em The Annals of Probability}, 21(3):1275--1294, 1993.

\bibitem{bs10spectral}
Zhidong Bai and Jack~W. Silverstein.
\newblock {\em Spectral Analysis of Large Dimensional Random Matrices}.
\newblock Springer Series in Statistics. Springer, New York, 2 edition, 2010.

\bibitem{bl25}
Randall Balestriero and Yann LeCun.
\newblock {LeJEPA}: Provable and scalable self-supervised learning without the
  heuristics.
\newblock 2025.
\newblock arxiv.org/abs/2511.08544.

\bibitem{bc2017convex}
Heinz~H. Bauschke and Patrick~L. Combettes.
\newblock {\em Convex Analysis and Monotone Operator Theory in Hilbert Spaces}.
\newblock CMS Books in Mathematics. Springer, Cham, 2 edition, 2017.

\bibitem{bhmm19}
Mikhail Belkin, Daniel Hsu, Siyuan Ma, and Soumik Mandal.
\newblock Reconciling modern machine-learning practice and the classical
  bias–variance trade-off.
\newblock {\em Proceedings of the National Academy of Sciences of the United
  States of America}, 116(32):15849--15854, 2019.

\bibitem{bhx20}
Mikhail Belkin, Daniel Hsu, and Ji~Xu.
\newblock Two models of double descent for weak features.
\newblock {\em SIAM Journal on Mathematics of Data Science}, 2(4):1167--1180,
  2020.

\bibitem{brt19}
Mikhail Belkin, Alexander Rakhlin, and Alexandre~B. Tsybakov.
\newblock Does data interpolation contradict statistical optimality?
\newblock In Kamalika Chaudhuri and Masashi Sugiyama, editors, {\em Proceedings
  of the Twenty-Second International Conference on Artificial Intelligence and
  Statistics}, volume~89 of {\em Proceedings of Machine Learning Research},
  pages 1611--1619. PMLR, 16--18 Apr 2019.

\bibitem{bishop2006pattern}
Christopher Bishop.
\newblock {\em Pattern Recognition and Machine Learning}.
\newblock Springer, January 2006.

\bibitem{bmasdw22}
Paolo Braca, Leonardo~M. Millefiori, Augusto Aubry, Stefano Marano, Antonio
  De~Maio, and Peter Willett.
\newblock Statistical hypothesis testing based on machine learning: Large
  deviations analysis.
\newblock {\em IEEE Open Journal of Signal Processing}, 3:464--495, 2022.

\bibitem{brown2020language}
Tom~B. Brown, Benjamin Mann, Nick Ryder, et~al.
\newblock Language models are few-shot learners.
\newblock In {\em Advances in Neural Information Processing Systems},
  volume~33, pages 1877--1901, 2020.

\bibitem{SONYC-UST-V2}
Mark Cartwright, Ana Elisa~Mendez Mendez, Jason Cramer, Vincent Lostanlen,
  Graham Dove, Hao-Hsiang Wu, Justin Salamon, Oded Nov, and Juan~Pablo Bello.
\newblock {SONYC} urban sound tagging ({SONYC-UST}): A multilabel dataset from
  an urban acoustic sensor network.
\newblock In {\em Proceedings of the Workshop on Detection and Classification
  of Acoustic Scenes and Events (DCASE)}, pages 35--39, 2019.

\bibitem{BEATs}
Sanyuan Chen, Yu~Wu, Chengyi Wang, Shujie Liu, Daniel Tompkins, Zhuo Chen,
  Wanxiang Che, Xiangzhan Yu, and Furu Wei.
\newblock Beats: audio pre-training with acoustic tokenizers.
\newblock In {\em Proceedings of the 40th International Conference on Machine
  Learning}. JMLR.org, 2023.

\bibitem{chernoff1952}
Herman Chernoff.
\newblock A measure of asymptotic efficiency for tests of a hypothesis based on
  the sum of observations.
\newblock {\em The Annals of Mathematical Statistics}, 23(4):493--507, 1952.

\bibitem{cramer1938}
Harald Cram{\'e}r.
\newblock Sur un nouveau th{\'e}or{\`e}me--limite de la th{\'e}orie des
  probabilit{\'e}s.
\newblock In {\em Actualit{\'e}s Scientifiques et Industrielles}, number 736,
  pages 5--23. Hermann, Paris, 1938.
\newblock Colloque consacr{\'e} \`a la th{\'e}orie des probabilit{\'e}s.

\bibitem{dz10ld}
Amir Dembo and Ofer Zeitouni.
\newblock {\em Large Deviations Techniques and Applications}.
\newblock Stochastic Modelling and Applied Probability. Springer, Berlin,
  Heidelberg, 2 edition, 2010.

\bibitem{deng2022model}
Zeyu Deng, Abla Kammoun, and Christos Thrampoulidis.
\newblock A model of double descent for high-dimensional binary linear
  classification.
\newblock {\em Information and Inference: A Journal of the IMA},
  11(2):435--495, 06 2022.

\bibitem{dobriban2018high}
Edgar Dobriban and Stefan Wager.
\newblock High-dimensional asymptotics of prediction: Ridge regression and
  classification.
\newblock {\em The Annals of Statistics}, 46(1):247--279, 2018.

\bibitem{llama3herd}
Abhimanyu Dubey et~al.
\newblock The llama 3 herd of models.
\newblock {\em arXiv preprint arXiv:2407.21783}, 2024.

\bibitem{ellis84}
Richard~S. Ellis.
\newblock Large deviations for a general class of random vectors.
\newblock {\em The Annals of Probability}, 12(1):1--12, 1984.

\bibitem{gartner1977}
J{\"u}rgen G{\"a}rtner.
\newblock On large deviations from the invariant measure.
\newblock {\em Theory of Probability and Its Applications}, 22(1):24--39, 1977.

\bibitem{audioset}
Jort~F. Gemmeke, Daniel P.~W. Ellis, Dylan Freedman, Aren Jansen, Wade
  Lawrence, R.~Channing Moore, Manoj Plakal, and Marvin Ritter.
\newblock Audio set: An ontology and human-labeled dataset for audio events.
\newblock In {\em 2017 IEEE International Conference on Acoustics, Speech and
  Signal Processing (ICASSP)}, pages 776--780, 2017.

\bibitem{hbmw24}
Ryan Harvey, Paolo Braca, Leonardo~M. Millefiori, and Peter Willett.
\newblock Sequential hypothesis testing based on machine learning.
\newblock In {\em 2024 27th International Conference on Information Fusion
  (FUSION)}, pages 1--8, 2024.

\bibitem{hmrt22high}
Trevor Hastie, Andrea Montanari, Saharon Rosset, and Ryan~J. Tibshirani.
\newblock Surprises in high-dimensional ridgeless least squares interpolation.
\newblock {\em The Annals of Statistics}, 50(2):949--986, 2022.

\bibitem{airbus_dataset}
Kaggle.
\newblock Airbus ship detection challenge.
\newblock \url{https://www.kaggle.com/competitions/airbus-ship-detection},
  2018.
\newblock Accessed: 2025-11-10.

\bibitem{kmhbccgrwa20}
Jared Kaplan, Sam McCandlish, Tom Henighan, Tom~B. Brown, Benjamin Chess, Rewon
  Child, Scott Gray, Alec Radford, Jeffrey Wu, and Dario Amodei.
\newblock Scaling laws for neural language models, 2020.

\bibitem{Kay:B93_2}
Steven~M. Kay.
\newblock {\em Fundamentals of Statistical Signal Processing, Detection
  Theory}.
\newblock Prentice Hall PTR, Upper Saddle River, NJ, USA, 1998.

\bibitem{nv_embed}
Chankyu Lee, Rajarshi Roy, Mengyao Xu, Jonathan Raiman, Mohammad Shoeybi, Bryan
  Catanzaro, and Wei Ping.
\newblock Nv-embed: Improved techniques for training llms as generalist
  embedding models.
\newblock In {\em International Conference on Learning Representations (ICLR)},
  2025.

\bibitem{lehmann2022testing}
Erich~L. Lehmann and Joseph~P. Romano.
\newblock {\em Testing Statistical Hypotheses}.
\newblock Springer Texts in Statistics. Springer Cham, 4 edition, 2022.

\bibitem{lpjbt24}
Daniel LeJeune, Pratik Patil, Hamid Javadi, Richard~G. Baraniuk, and Ryan~J.
  Tibshirani.
\newblock Asymptotics of the sketched pseudoinverse.
\newblock {\em SIAM Journal on Mathematics of Data Science}, 6(1):199--225,
  2024.

\bibitem{lvmkt20}
M.~Loog, T.~Viering, A.~Mey, J.~H. Krijthe, and D.~M.~J. Tax.
\newblock A brief prehistory of double descent.
\newblock {\em Proceedings of the National Academy of Sciences of the United
  States of America}, 117(20):10625--10626, 2020.

\bibitem{marchenko1967distribution}
Vladimir~Alexandrovich Marchenko and Leonid~Andreevich Pastur.
\newblock Distribution of eigenvalues for some sets of random matrices.
\newblock {\em Matematicheskii Sbornik}, 114(4):507--536, 1967.

\bibitem{mm22}
Song Mei and Andrea Montanari.
\newblock The generalization error of random features regression: Precise
  asymptotics and the double descent curve.
\newblock {\em Communications on Pure and Applied Mathematics}, 75(4):667--766,
  2022.

\bibitem{llama_nemotron}
Gabriel de Souza~P. Moreira, Radek Osmulski, Mengyao Xu, Ronay Ak, Benedikt
  Schifferer, and Even Oldridge.
\newblock Improving text embedding models with positive-aware hard-negative
  mining.
\newblock In {\em Proceedings of the 34th ACM International Conference on
  Information and Knowledge Management}, page 2169–2178. Association for
  Computing Machinery, 2025.

\bibitem{nakkiran2020deep}
Preetum Nakkiran, Gal Kaplun, Yamini Bansal, Tristan Yang, Boaz Barak, and Ilya
  Sutskever.
\newblock Deep double descent: Where bigger models and more data hurt.
\newblock In {\em International Conference on Learning Representations (ICLR)},
  2020.

\bibitem{okkn90}
M.~Opper, W.~Kinzel, J.~Kleinz, and R.~Nehl.
\newblock On the ability of the optimal perceptron to generalise.
\newblock {\em Journal of Physics A: Mathematical and General}, 23:L581--L586,
  1990.
\newblock Letter to the Editor.

\bibitem{patil2024asymptotically}
Pratik Patil and Daniel LeJeune.
\newblock Asymptotically free sketched ridge ensembles: Risks,
  cross-validation, and tuning.
\newblock In {\em The Twelfth International Conference on Learning
  Representations}, 2024.

\bibitem{ps75}
Detlef Plachky and Josef Steinebach.
\newblock A theorem about probabilities of large deviations with an application
  to queuing theory.
\newblock {\em Periodica Mathematica Hungarica}, 6(4):343--345, 1975.

\bibitem{rm11spectral}
Francisco Rubio and Xavier Mestre.
\newblock Spectral convergence for a general class of random matrices.
\newblock {\em Statistics \& Probability Letters}, 81(5):592--602, 2011.

\bibitem{salehi2019impact}
Fariborz Salehi, Ehsan Abbasi, and Babak Hassibi.
\newblock The impact of regularization on high-dimensional logistic regression.
\newblock In H.~Wallach, H.~Larochelle, A.~Beygelzimer, F.~d\textquotesingle
  Alch\'{e}-Buc, E.~Fox, and R.~Garnett, editors, {\em Advances in Neural
  Information Processing Systems}, volume~32. Curran Associates, Inc., 2019.

\bibitem{s97linear}
Shayle~R. Searle.
\newblock {\em Linear Models}.
\newblock Wiley Classics Library. John Wiley \& Sons, 1997.

\bibitem{fire_risk_dataset}
Shuchang Shen, Sachith Seneviratne, Xinye Wanyan, and Michael Kirley.
\newblock Firerisk: A remote sensing dataset for fire risk assessment with
  benchmarks using supervised and self-supervised learning.
\newblock In {\em 2023 International Conference on Digital Image Computing:
  Techniques and Applications (DICTA)}, pages 189--196, 2023.

\bibitem{sb95empirical}
Jack~W. Silverstein and Zhidong~D. Bai.
\newblock On the empirical distribution of eigenvalues of a class of large
  dimensional random matrices.
\newblock {\em Journal of Multivariate Analysis}, 54(2):175--192, 1995.

\bibitem{dinov3}
Oriane Siméoni, Huy~V. Vo, Maximilian Seitzer, Federico Baldassarre, Maxime
  Oquab, Cijo Jose, Vasil Khalidov, Marc Szafraniec, Seungeun Yi, Michaël
  Ramamonjisoa, Francisco Massa, Daniel Haziza, Luca Wehrstedt, Jianyuan Wang,
  Timothée Darcet, Théo Moutakanni, Leonel Sentana, Claire Roberts, Andrea
  Vedaldi, Jamie Tolan, John Brandt, Camille Couprie, Julien Mairal, Hervé
  Jégou, Patrick Labatut, and Piotr Bojanowski.
\newblock Dinov3, 2025.

\bibitem{sur2019modern}
Pragya Sur and Emmanuel~J. Cand{\`e}s.
\newblock A modern maximum-likelihood theory for high-dimensional logistic
  regression.
\newblock {\em Proceedings of the National Academy of Sciences},
  116(29):14516--14525, 2019.

\bibitem{av04}
Antonia Tulino and Sergio Verdú.
\newblock Random matrix theory and wireless communications.
\newblock {\em Foundations and Trends in Communications and Information
  Theory}, 1, 01 2004.

\bibitem{vcr89}
F.~Vallet, J.-G. Cailton, and Ph~Refregier.
\newblock Linear and nonlinear extension of the pseudo-inverse solution for
  learning boolean functions.
\newblock {\em Europhysics Letters}, 9(4):315, jun 1989.

\end{thebibliography}

\newpage
\appendix

\section*{Supplementary Material}
{\hypersetup{linkcolor=black}
\startcontents[appendix]
\printcontents[appendix]{}{1}{\setcounter{tocdepth}{2}}
}
\newpage

\section{Preliminaries}

\subsection{Notation}\label{sec:supp_notation}
$\R$ is the real line, $\R_+$ and $\R_{++}$ are non-negative and positive real numbers. $\C$ is the complex plane and $\C_+ := \{ z \in \C \mid \mathfrak{Im}(z) > 0\}$. We denote by $\mathbf{1}:=(1, \dots, 1)\in \R^d$, with $d\in \N$ depending on the context. We use the notation $\lesssim$ to indicate that the inequality holds up to a positive constant. For a matrix $A\in \R^{p\times p}$ with real eigenvalues $\xi_1, \dots, \xi_p$, we call the \emph{spectral measure of $A$} the probability measure defined as the sum of $\delta_{\xi_i}$, i.e., the Dirac deltas centered on $\xi_i$, with uniform weights. The set \arev{$\mathcal{P}(D)$} denotes the family of probability measures on $D\subset \R^d$, $d\in \N$.  The set $C_b(\R)$ the family of continuous and bounded functions on $\R$. We use the notation of \emph{asymptotic equivalents}, i.e., for two sequences $(a_n)_n$ and $(b_n)_n$ we say that $a_n \sim b_n$ as $n\to + \infty$ if $\lim_{n\to +\infty} \frac{a_n}{b_n} = 1$. We denote by $\E$ the \emph{expectation} with respect to the underlying probability measure $\mathbb{P}$ defined on a suitable probability space $(\Omega, \mathfrak{F}, \mathbb{P})$ where $\mathfrak{F}$ is the $\sigma$-algebra, and by $\Var$ the \emph{variance}. We use $\E_m$ to refer to the expectation taken with respect to the measure $m$.

\subsection{Random matrix theory}\label{sec:supp_rmt}

We start out by showing how to solve the Silverstein equation \eqref{eq:silverstein_fpe}.

\paragraph{Solving the Silverstein Equation} The Silverstein equation is a nonlinear fixed-point equation that does not admit a closed-form solution in general.  Let us first propose an approximation strategy based on fixed-point iterations. To solve \eqref{eq:supp_silverstein_on_k} for a fixed $\lambda>0$, we need to find a fixed point of the operator $\mathcal{F}\colon [0, +\infty)\to [0, +\infty)$ defined by
\begin{equation}
\mathcal{F}(\kappa):=\lambda + \gamma \kappa \int_0^{+\infty}\frac{\xi}{\kappa + \xi} \diff H(\xi)\,, \quad \text{for all} \ \kappa \geq 0\,,
\end{equation}
which can be easily seen to be an increasing and concave function in $[0, +\infty)$, due to Assumption \ref{ass:A1}. Therefore, for each fixed $\lambda$, we can approximate efficiently $\kappa(\lambda)$ by $\kappa_{\lambda}^{K}$ obtained by fixed-point iterations $\kappa_{\lambda}^{i+1} = \mathcal{F}(\kappa_{\lambda}^{i})$ for all $i =0, \dots, K-1$. 

An alternative approximation strategy used, e.g., in \cite{bach24high}, consists in taking the almost-sure limit of
\begin{equation}\label{eq:supp_companion_transform}
\hat \varphi(z) := \frac{1}{n} \trace\bigg[\bigg( \frac{1}{n}\hat X^T \hat X - z I\bigg)^{-1} \bigg]\,,
\end{equation}
since indeed for all $z\in \C \setminus \R_+$, $\hat \varphi(z) \to \varphi(z)$ almost surely. However, this requires inverting high-dimensional matrices and the results are subject to randomness.

\paragraph{Isotropic data} In the case of isotropic data, the solution to the Silverstein equation is explicit.  Let $\Sigma =  \sigma^2 I$, for some $\sigma > 0$, then $H = \delta_{\sigma^2}$, and the Silverstein fixed-point equation \eqref{eq:silverstein_fpe} turns into
\begin{equation}
\frac{1}{\varphi(z)} + z = \gamma \frac{\sigma^2}{1 + \sigma^2 \varphi(z)}\, \quad \text{for} \ z \in \C\setminus \R_+\,.
\end{equation}
This admits the following closed-form solution: Setting $\gamma_\pm=\sigma^2(1\pm\sqrt{\gamma})^2$, 
\begin{equation}\label{eq:supp_phi_isotropic}
\varphi(z) = \frac{-(z+\sigma^2-\gamma\sigma^2)+q(z)}{2\sigma^2 z}\,, \quad q(z)^2:=(z-\gamma_-)(z-\gamma_+)\,,
\end{equation}
\arev{where $q$ denotes the analytic branch satisfying $q(z)/z\to 1$ as $|z|\to\infty$. In particular, $q(z)<0$ for real $z<0$.} Plugged into (A.5) of \cite{bach24high}, i.e.,
\begin{equation}\label{eq:relation_psi_phi}
\gamma \psi(z) = \varphi(z) + \frac{1 - \gamma}{z}\,, \quad \text{for all} \ z \in \C_+\,,
\end{equation}
yields an explicit formula for $\psi$, and, by Stieltjes inversion, yields the celebrated Marchenko--Pastur distribution \cite{marchenko1967distribution}, which is given by:
\begin{equation}\label{eq:supp_marchenko_pastur_distribution}
\nu_{\mathrm{MP}} := \max\Big\{0, 1 - \frac{1}{\gamma}\Big\} \delta_0 + f_\gamma \diff x, \quad \text{where} \quad f_\gamma(x):=\frac{1}{2\pi \sigma^2} \frac{\sqrt{(\gamma_+ - x)(x-\gamma_-)}}{\gamma x}\,.
\end{equation} 

\paragraph{Trace limits} \arev{We recall below a technical result due to Rubio and Mestre \cite{rm11spectral} and Bai and Silverstein \cite{bs10spectral}, which we use repeatedly. See \cite{bach24high} for a comprehensive, self-contained account.}
\begin{thm}\label{thm:rmb}
	Let $\hat X =  \Sigma^{1/2} \hat Z$ with $\hat Z \in \R^{p \times n}$ having sub-Gaussian i.i.d.~components with mean zero and unit variance and suppose that Assumption \ref{ass:1} holds. Let $A, B$ bounded in operator norm and such that $\sum_{i=1}^p (v_i^T A v_i) \delta_{\sigma_i}$ and $\sum_{i=1}^p (v_i^T B v_i) \delta_{\sigma_i}$ converge to measures $\nu_A$ and $\nu_B$ with bounded total variation. Then, for $z \in \C \setminus\R_+$, as $p, n \to +\infty$ with $\gamma = p/n\geq 0$ constant, we get
	\begin{align}
	& \trace\big[A (\hat \Sigma - z I)^{-1}\big] \sim -\frac{1}{z \varphi(z)} \trace\bigg[ A \bigg(\Sigma + \frac{1}{\varphi(z)}I\bigg)^{-1}\bigg]\,, \label{eq:asymtotic_equivalent_mu}\\
	& \trace\big[A (\hat \Sigma - z I)^{-1}B (\hat \Sigma - z I)^{-1}\big] \sim \frac{1}{z^2 \varphi(z)^2} \trace\bigg[A \bigg(\Sigma + \frac{1}{\varphi(z)} I\bigg)^{-1}B \bigg( \Sigma +  \frac{1}{\varphi(z)} I\bigg)^{-1}\bigg]  \label{eq:asymtotic_equivalent_sig} \\
	& \qquad + \frac{1}{z^2 \varphi(z)^2} \trace\bigg[A \bigg(\Sigma + \frac{1}{\varphi(z)} I\bigg)^{-2}\Sigma\bigg]\trace\bigg[B \bigg(\Sigma + \frac{1}{\varphi(z)} I\bigg)^{-2}\Sigma\bigg]\frac{1}{n - \diff f_2(1/\varphi(z))}\,.\nonumber
	\end{align}
	where $\varphi$ is the unique solution to \eqref{eq:silverstein_fpe} and $\diff f_2(\kappa):= \trace\big(\Sigma^2(\Sigma + \kappa I)^{-2}\big)$ for all $\kappa \in \C\setminus \R_-$.
\end{thm} 

Observe that the Gaussianity assumption is not necessary to get the results of Theorem \ref{thm:rmb}, \arev{see also \cite{lpjbt24, patil2024asymptotically} for weaker assumptions on $A$ and $B$.}

\subsection{Large deviation theory}\label{app:supp_preliminaries_LDP}

The theory of large deviations investigates the asymptotic behavior of the tails of sequences of probability distributions. Before delving into the general framework, we discuss a simple example:

\begin{example}\label{example:supp_ldp_gaussian}
	Suppose that $(\hat T (x))_p$ is a sequence of Gaussian random variables with $\E[\hat T (x) \mid \cH_0] = 0$, $\E[\hat T (x) \mid \cH_1] = \hat \mu$ and $\Var[\hat T (x) \mid \cH_0] = \Var[\hat T (x) \mid \cH_1]=\hat \sigma^2$. The probability of \emph{miss detection}, and of the \emph{false alarm}, only depend on the first two moments:
	\begin{equation}\label{eq:supp_pmd_and_pfa}
	\begin{aligned}
	&P_{FA} :=  \P(\hat T (x) \geq \tau \mid \cH_0) = Q\big( \tau / \hat \sigma \big) = Q\big(Q^{-1}(P_{D}) + \hat d \big)\,,\\
	&P_{MD} := \P(\hat T(x) \leq \tau \mid \cH_1) = 1 - Q\Big(\frac{\tau - \hat \mu}{\hat\sigma}\Big) = 1 - Q\big(Q^{-1}(P_{FA}) - \hat d\big)\,,
	\end{aligned}
	\end{equation}
	where $Q$ is the standard $Q$-function and $\hat d^2 : = \hat \mu^2 / \hat \sigma^2$.	In this setting, using the well-known asymptotic behavior of the $Q$ function, if $\hat d^2 \sim p$ as $p\to +\infty$, the probability of miss detection with fixed false alarm $P_{FA}=\alpha > 0$, i.e., with $\tau = \hat \sigma Q^{-1}(\alpha)$, can be expressed in terms of $\hat d$ as:
	\begin{equation}\label{eq:supp_def_rate_linear}
	\rev{\frac{1}{p} \log} \ P_{MD} = \rev{\frac{1}{p} \log} \big(1 - Q\big(Q^{-1}(\alpha) - \hat d\big) \big) = -\frac{\hat d^2}{2p} + o(1)\,, \quad \text{as} \ p \to +\infty\,.
	\end{equation}
	Clearly, the same result holds switching the roles of $P_{FA}$ and $P_{MD}$. Observe that the decay of $P_{MD}$ is governed by the left tail of $\hat T(x)$ under $\cH_1$, and is by no means a surprise that in the Gaussian setting it can indeed be controlled by only its scale parameters.
\end{example}

In a different yet simple setting, the large deviation theory provides the rate of convergence in the weak law of large numbers. To be precise, consider any i.i.d.~random variables $z_1, \dots, z_n$ with mean $\mu$ and LMGF $\Lambda(s) = \log \E(e^{sz_1}) < +\infty$ for all $s\in \R$, and let $S_n := n^{-1}\sum_{i=1}^n z_i$. Then for any $\tau > \mu$, \arev{we have the following decay of the tail probabilities:}
\begin{equation}
    \lim_{n \to +\infty} \frac{1}{n} \log \P(S_n > \tau) = - \Lambda^*(\tau)\,,
\end{equation}
where $\Lambda^*$ is the Fenchel--Legendre conjugate of $\Lambda$ \cite{cramer1938}, see \cite[Theorem 2.2.30]{dz10ld}. In this context, we say that the LDP holds with rate function $\Lambda^*$. More generally, we refer to the following definition:
\begin{defn}[Large Deviation Principle]\label{def:supp_ldp}
	A sequence of random variables $(\mathsf{t}_p)_p$ is said to satisfy the \emph{Large Deviation Principle} with \emph{rate function} $I\colon \R \to \R \cup \{+\infty\}$ if for all Borel sets $\Gamma \subset \R$
	\begin{equation}\label{eq:supp_intro_def_ldp}
	\begin{aligned}
	-\inf_{\tau \in \Int \Gamma} I(\tau) &\leq \liminf_{p\to +\infty} \frac{1}{p} \log \P(\mathsf{t}_p \in \Gamma )\\
	&\leq \limsup_{p\to +\infty} \frac{1}{p} \log \P(\mathsf{t}_p  \in \Gamma ) \leq -\inf_{\tau \in \bar{\Gamma}} I(\tau)\,.
	\end{aligned}
	\end{equation}
\end{defn} 
The following results, due to G\"artner \cite{gartner1977}, Ellis \cite{ellis84} and Plachky and Steinebach \cite{ps75}, provides a broad class of settings in which the LDP holds. It requires that the following (possibly infinite) limit exists:
\begin{equation}\label{eq:supp_intro_limit_lmgf}
\Lambda(s) := \lim_{p\to +\infty} \frac{1}{p} \Lambda_p(ps) := \lim_{p\to +\infty} \frac{1}{p} \log \E(e^{p s  \mathsf{t}_p}) \,, \quad \text{for all} \ s \in \R\,.
\end{equation}
We provide below a simplified statement in one dimension and refer, for example, to \cite[Theorem 2.3.6]{dz10ld} for a complete discussion. In particular, the lower bound depends on the concept of \emph{exposed points} of a convex function, which formally captures points of strict convexity; see \cite[Definition 2.3.3]{dz10ld} for a precise definition.
\begin{thm}[G\"artner--Ellis]\label{thm:supp_gartner_ellis}
	Let $(\mathsf{t}_p)_p$ be a sequence of random variables such that $\Lambda \colon \R\to\R\cup \{+\infty\}$ defined in \eqref{eq:supp_intro_limit_lmgf} exists and is such that $0 \in \Int \Dom \Lambda$. Then:
	\begin{enumerate}
		\item For every closed set $C\subset \R$,
		\begin{equation}
			\limsup_{p\to +\infty} \frac{1}{p} \log \P(\mathsf{t}_p  \in C ) \leq -\inf_{\tau \in C} \Lambda^*(\tau)\,.
		\end{equation}
		\item For every open set $O\subset \R$, 
		\begin{equation}
			\liminf_{p\to +\infty} \frac{1}{p} \log \P(\mathsf{t}_p \in O) \geq -\inf_{\tau \in O \cap \mathcal{E} } \Lambda^*(\tau)\,,
		\end{equation}
		where $\mathcal{E}$ is the set of \emph{exposed points} for $\Lambda^*$ whose exposing line belongs to $\Int \Dom \Lambda$.
	\end{enumerate}
	Suppose in addition that $\Lambda$ is lower semicontinuous in $\R$, differentiable in $\Int \Dom \Lambda$, and such that 
	\begin{equation}\label{eq:supp_def_steepness}
	\lim_{s\to \bar s} |\Lambda'(s)| = +\infty\,, \quad \text{for all} \ \bar s \in \partial \Dom \Lambda\,.
	\end{equation}
	Then, $( \mathsf{t}_p )_p$ satisfies the LDP with rate function $\Lambda^*$.
\end{thm}

\section{Missing proofs and further insights of Section \ref{sec:linear}}\label{sec:linear_missing_proofs}

The following remark is going to be useful to lighten the upcoming proofs.
\begin{fact}\label{fact:supp_observables_as_trace_limits}
	Suppose that Assumption \ref{ass:1} holds. The observables $\mathfrak{m}_q$ and $\mathfrak{n}_q$ can be written as limits of trace functionals:
	Let $v_1, \dots, v_p$ be the basis of eigenvectors of $\Sigma$ and $\xi_1, \dots, \xi_p$ the corresponding eigenvalues. By the properties of the trace operator, as $p\to +\infty$,
	\begin{equation*}
	\begin{aligned}
	\trace\bigg[ \frac{s s^T}{p} \big(\Sigma + \kappa I\big)^{-1}\bigg] = \frac{1}{p}\sum_{i=1}^p  \frac{\left(v_i^T s\right)^2}{\xi_i+ \kappa} \to  & \int_{\R_+} \frac{1}{\xi + \kappa }\diff \nu_s(\xi) \\
	&=   \int_{\R_+} \left(\frac{\xi}{\xi + \kappa }\right)\lSNR(\xi)\diff H(\xi) = \mathfrak{n}_1(\kappa)\,,
	\end{aligned}
	\end{equation*}
	where in the last step we have used Assumption \ref{ass:A2} with $\theta^*=s$. Similarly, 
	\begin{equation*}
	\trace\bigg[ \frac{s s^T}{p} \big(\Sigma + \kappa I\big)^{-2}\Sigma \bigg] \sim \trace\bigg[ \frac{s s^T}{p} \big(\Sigma + \kappa I\big)^{-1}\Sigma \big(\Sigma + \kappa I\big)^{-1} \bigg] \sim \mathfrak{n}_2(\kappa)\,.
	\end{equation*}
	While, for $\mathfrak{m}_q$ we have, using the same approach $\frac{1}{p}\trace\big[ \Sigma \big(\Sigma +\kappa  I\big)^{-1}\big] \sim \m_1(\kappa )$ and 
	\begin{equation*}
	\frac{1}{p}\trace\big[ \Sigma \big(\Sigma +\kappa I\big)^{-2} \Sigma\big] \sim 	\frac{1}{p}\trace\big[ \Sigma \big(\Sigma +\kappa  I\big)^{-1} \Sigma  \big(\Sigma +\kappa  I\big)^{-1} \big] \sim  \m_2(\kappa)\,.
	\end{equation*}
\end{fact}

We now discuss how Assumption \ref{ass:1} as well as the observables defined in Definition \ref{def:observables} translate in two notable cases:

\begin{example}[Isotropic data]\label{example:supp_isotropic_data}
	Suppose that for each $p\in \N$, $\Sigma = \sigma^2 I$ with $\sigma > 0$. Then Assumptions \ref{ass:A1} and \ref{ass:A2} are satisfied. Indeed, the spectral measure of $\Sigma$ is constant $\delta_{\sigma^2}$, hence converges to itself, showing Assumption \ref{ass:A1} with $H= \delta_{\sigma^2}$ \arev{and $a=b=\sigma^2$}. \arev{Moreover, pick a signal $\theta^*=(\theta^*_1, \dots, \theta^*_p) \in \R^p$ such that $\frac{1}{p}\|\theta^*\|^2$ converges. Then, $\|\theta^*\|^2=\mathcal{O}(p)$ as $p\to +\infty$, and 
	\begin{equation}
	\frac{1}{p}\sum_{i=1}^p (v_i^T \theta^*)^2 \delta_{\xi_i} = \frac{1}{p}\sum_{i=1}^p (\theta_i^*)^2 \delta_{\sigma^2} = \frac{1}{p}\|\theta^*\|^2 \delta_{\sigma^2} \rightharpoonup \Big( \lim_{p\to +\infty} \frac{1}{p} \|\theta^*\|^2\Big) \delta_{\sigma^2} =: \nu_{\theta^{*}}\,.
	\end{equation}
    Furthermore, $\nu_s \ll H$, and in this case} $\lSNR$ is a function defined only on $\sigma^2$ with value that indeed coincides with $\lim_{p\to +\infty}\frac{1}{p}\|s\|^2 / \sigma^2$, i.e., the $\SNR$; \arev{thus, Assumption \ref{ass:A2} holds}. Therefore, $\lSNR \diff H = \SNR\delta_{\sigma^2}$, and
	\begin{equation}\label{eq:supp_observables_isotropic}
	\mathfrak{m}_q(\kappa) = \Big( \frac{\sigma^2}{\sigma^2 + \kappa}\Big)^q\,, \quad \text{and} \quad \mathfrak{n}_q(\kappa) = \Big( \frac{\sigma^2}{\sigma^2 + \kappa}\Big)^q \SNR\,.
	\end{equation}
\end{example}

\begin{example}\label{example:supp_k_0}
	Suppose that Assumptions \ref{ass:A1} and \ref{ass:A2} hold. \arev{Then, $\lim_{p\to +\infty} \frac{1}{p} s^T\Sigma^{-1}s$ exists. Additionally,} for all $q\in \N$, we have, clearly $\mathfrak{m}_q(0) = 1$, and
	\begin{equation}
	\begin{aligned}
	\mathfrak{n}_q(0) = \int_{\R_+} \lSNR(\xi) \diff H(\xi) &=  \int_{\R_+} \frac{1}{\xi} \diff \nu_s(\xi)\\
	& =\lim_{p\to +\infty} \frac{1}{p}\sum_{i=1}^p \frac{(v_i^T s)^2}{\xi_i} =  \lim_{p\to +\infty} \frac{1}{p} s^T\Sigma^{-1}s = \SNR\,.
	\end{aligned}
	\end{equation}
	This shows that $\mathfrak{m}_q$ and $\mathfrak{n}_q$ can be seen as regularized moments of the limiting spectral distribution of $\Sigma$ and the signal-to-noise ratio, respectively.
\end{example}

\begin{proof}[Proof of Proposition \ref{prop:mean_var_linear}]
	Start from the mean and let us first consider the decision statistic $\hat T(x) = s^T \hat \Sigma_{\lambda}^{-1}x$. We have:
	\begin{equation}\label{eq:supp_mu_hat_overp_1}
	\begin{aligned}
	\frac{\hat \mu}{p} & =   \frac{1}{p}\E [\hat T(x) \mid \hat X, {\cal H}_1 ] = \frac{1}{p} s^T \hat \Sigma_\lambda^{-1}s  = \frac{1}{p}\trace \big[s^T \hat \Sigma_\lambda^{-1}s\big] = \trace \bigg[\frac{ss^T}{p} \hat \Sigma_\lambda^{-1}\bigg] \\
	& = \trace \bigg[\frac{ss^T}{p} \big( \hat \Sigma + \lambda I \big)^{-1}\bigg] \sim \frac{\kappa(\lambda)}{\lambda }\trace\bigg[ \frac{s s^T}{p} \big(\Sigma + \kappa(\lambda)I\big)^{-1}\bigg] \sim   \frac{\kappa(\lambda)}{\lambda } \n_1(\kappa(\lambda))\,,
	\end{aligned}
	\end{equation}
	where in the last line we have used Theorem \ref{thm:rmb} with $A:=\frac{ss^T}{p}$, for which $\|A\|=\|s\|^2/p$ and is then bounded (in $p$), and Fact \ref{fact:supp_observables_as_trace_limits}. Thus the convergence of $\hat \mu / p$.
	
	We argue analogously for $\hat \sigma^2$ applying Theorem \ref{thm:rmb} to
	\begin{equation}\label{eq:supp_avar_1_proof}
	\begin{aligned}
	\hat \sigma^2 & = \E[(\hat T(x) - \hat \mu )^2 \mid  \hat X, \cH_1] = \E[(s^T \hat \Sigma_\lambda^{-1} (x - s))^2 \mid  \hat X, \cH_1]\\
	& = s^T \hat \Sigma_\lambda^{-1} \E\big[(x - s)(x - s)^T \mid  \hat X, \cH_1 \big] \hat \Sigma_\lambda^{-1}s = s^T \hat \Sigma_\lambda^{-1} \Sigma \hat \Sigma_\lambda^{-1}s = \trace \big[ ss^T \hat \Sigma_\lambda^{-1} \Sigma \hat \Sigma_\lambda^{-1}\big]\,.
	\end{aligned}
	\end{equation}
	Therefore, we can now use \eqref{eq:asymtotic_equivalent_sig} with $A:= ss^T/p$ and $B:=\Sigma$, to get that, almost surely, as $n, p\to +\infty$ with $p / n= \gamma \in (0, \infty)$,
	\begin{equation*}
	\begin{aligned}
	\frac{\hat \sigma^2}{p} &\sim \frac{\kappa(\lambda)^2}{\lambda^2 } \trace\bigg[ \frac{s s^T}{p} \big(\Sigma + \kappa(\lambda) I\big)^{-1}\Sigma \big(\Sigma + \kappa(\lambda) I\big)^{-1}\bigg] \\
	& \qquad + \frac{\kappa(\lambda)^2}{\lambda^2 } \trace\bigg[ \frac{s s^T}{p} \big(\Sigma + \kappa(\lambda) I\big)^{-2}\Sigma\bigg]\trace\big[\Sigma \big(\Sigma + \kappa(\lambda) I\big)^{-2}\Sigma\big]\frac{1}{n - \diff f_2(\kappa(\lambda))}\\
	& \sim  \frac{\kappa(\lambda)^2 }{\lambda^2 } \n_2(\kappa(\lambda))  + \frac{\kappa(\lambda)^2}{\lambda^2  } \n_2(\kappa(\lambda)) \m_2(\kappa(\lambda)) \frac{p}{n - \trace\big(\Sigma^2(\Sigma + \kappa(\lambda) I)^{-2}\big)}\\
	& \sim  \frac{\kappa(\lambda)^2}{\lambda^2 } \n_2(\kappa(\lambda))  + \frac{\kappa(\lambda)^2}{\lambda^2 } \n_2(\kappa(\lambda)) \m_2(\kappa(\lambda)) \frac{\gamma}{1 - \gamma \m_2(\kappa(\lambda))}\,,
	\end{aligned}
	\end{equation*}
	where we have used once again Fact \ref{fact:supp_observables_as_trace_limits}. Rearranging we obtain \eqref{eq:mean_var_linear}. 
	
	If $s$ is estimated from the training set, we can reduce the analysis to the case in which the mean is known by employing the following argument. As in \eqref{eq:supp_mu_hat_overp_1} we obtain
	\begin{equation}\label{eq:supp_proof_mean_2}
	\begin{aligned}
	\frac{\hat \mu}{p} & =  \trace \bigg[\frac{s\hat s^T}{p} \big( \hat \Sigma + \lambda I \big)^{-1}\bigg]  = \underbrace{\trace \bigg[\frac{s s^T}{p} \big( \hat \Sigma + \lambda I \big)^{-1}\bigg]}_{:=A} + \underbrace{\trace \bigg[\frac{s(\hat s - s)^T}{p} \big( \hat \Sigma + \lambda I \big)^{-1}\bigg]}_{:=B}\,.
	\end{aligned}
	\end{equation}
	The asymptotic behavior of $A$ is discussed in the point above, the only difference being that in this case $\hat \Sigma$ is now centered with the sample mean $\hat s$, which does not change the results, see \cite[Chapter 3]{bs10spectral}. We now show that $B\to 0$ almost surely. Denote by $\hat \varepsilon := \hat s - s$ and note that $\hat \varepsilon \sim \cN(0, \frac{1}{n_1} \Sigma)$, hence $\hat z : = \Sigma^{-1/2}\hat \varepsilon \sim \cN(0, \frac{1}{n_1} I)$, thus
	\begin{equation}\label{eq:supp_proof_norm_of_z}
	\| \hat z \|^2 = \sum_{i=1}^p \hat z_i^2 = \frac{1}{n_1} \sum_{i=1}^p ( \sqrt{n_1} \hat z_i )^2 = \frac{p}{n_1} \frac{1}{p}\sum_{i=1}^p  ( \sqrt{n_1} \hat z_i )^2 \to \gamma_1\,, \quad \text{a.s.} \ \text{as} \ p\to +\infty\,,
	\end{equation}
	by the classical strong law of large numbers, where $\gamma_1>0$ is the limit of $p/n_1$. Hence, $\|\hat \varepsilon \|^2$ is almost surely bounded as $p\to +\infty$. Therefore, using Cauchy--Schwarz,
	\begin{equation*}
	|B|:=   \Big| \frac{1}{p} \hat \varepsilon^T \big( \hat \Sigma + \lambda I \big)^{-1} s \Big| \leq \frac{1}{p}\frac{\|\hat \varepsilon\| \|s\|}{\lambda} = \mathcal{O}(p^{-1/2})\,,\quad  \text{a.s. as} \ p\to +\infty\,,
	\end{equation*}
	
	Let us now deal with the variance. Following \eqref{eq:supp_avar_1_proof} and using the same approach as in \eqref{eq:supp_proof_mean_2}, we get
	\begin{equation}
	\frac{\hat \sigma^2}{p} = \frac{1}{p}\trace \big[ \hat s \hat s^T \hat \Sigma_\lambda^{-1} \Sigma \hat \Sigma_\lambda^{-1}\big] = \frac{1}{p} \hat s^T M \hat s =  \underbrace{\frac{1}{p} s^T M s}_{:= A} + \underbrace{\frac{2}{p} \hat \varepsilon^T M s}_{:= B} + \underbrace{\frac{1}{p} \hat \varepsilon^T M \hat \varepsilon}_{:= C}\;,
	\end{equation}
	where we have denoted by $M:= \hat \Sigma_\lambda^{-1} \Sigma \hat \Sigma_\lambda^{-1}$. As above, $\|M\|$ is uniformly bounded (in $p$) as $p\to +\infty$. Therefore, Cauchy--Schwarz yields again $B = \mathcal{O}(p^{-1/2})$ and $C = \mathcal{O}(p^{-1})$ almost surely as $p\to +\infty$. The only term remaining is $A$, whose limit is computed above.
\end{proof}

\begin{rmk}
	From \eqref{eq:supp_proof_norm_of_z}, we see that if $p / n_1$ grows faster than $\sqrt{p}$, a bias appears in the asymptotic mean of $\hat T$. This highlights the role of a balanced dataset.	
\end{rmk}

\rev{

\begin{rmk}\label{rmk:beyond_gaussianity_prop}
Proposition \ref{prop:mean_var_linear} extends to non-Gaussian data with i.i.d. sub-Gaussian coordinates. In fact, Theorem \ref{thm:rmb} does not require Gaussianity and the almost sure boundedness of $\hat\varepsilon$ also extends to this setting by replacing the strong law of large numbers in \eqref{eq:supp_proof_norm_of_z} with Bernstein's inequality and the Borel--Cantelli lemma.
\end{rmk}

}

\rev{
\subsection{Calibrating error probabilities}\label{app:supp_calibrate_errors_nongaussian}

Since $\hat T$ is Gaussian, the error probabilities can be calibrated directly from \eqref{eq:supp_pmd_and_pfa}. In particular, choosing $\tau=\hat\sigma Q^{-1}(\alpha)$
yields $P_{FA}=\alpha$, whereas choosing $\tau=\hat\mu+\hat\sigma Q^{-1}(1-\beta)$ yields $P_{MD}=\beta$.

We can asymptotically calibrate the error probabilities beyond the Gaussian case, relying on a central limit theorem for triangular arrays under an additional condition excluding dominant coordinates. Given the practical relevance of this extension, we establish the corresponding result below.

\begin{prop}\label{prop:supp_clt_linear_subgaussian}
Suppose that the coordinates of $z$ in \eqref{eq:intro_hypotheses_testing} are i.i.d.~sub-Gaussian random variables. Let $\hat T$ be defined as in \eqref{eq:def_That_linear_unknown} and set $\hat w:= \Sigma^{1/2}\hat\Sigma_\lambda^{-1}\hat s$, $\hat\mu_0=0$ and $\hat\mu_1=\hat\mu$. Assume that, almost surely,
\begin{equation}
    \hat\sigma = \|\hat w\|>0\,, \quad \text{and} \quad \lim_{p\to +\infty} \frac{\|  \hat w\|_\infty}{\|\hat w\|}=0\,.
\end{equation}
Then, conditionally on the training data, almost surely as $p \to +\infty$ 
\begin{equation}
    \frac{\hat T(x)-\hat\mu_i}{\hat\sigma} \xrightarrow{d}\mathcal N(0,1)\,, \qquad i=0,1\,.
\end{equation}
\end{prop}
\begin{proof}
For $i=0, 1$, under $\cH_i$ we can write $\hat T(x)-\hat\mu_i =\hat w^T z = \hat w_1 z_1+\dots +\hat w_p z_p$. The summands are independent, centered, and have total variance $\hat\sigma^2$. Since the coordinates of $z$ are sub-Gaussian, $\E[|z_1|^3]<+\infty$. Moreover,
\begin{equation*}
    \frac{1}{\hat\sigma^3} \sum_{j=1}^p \E\left[|\hat w_j z_j|^3 \mid \hat X\right] = \E[|z_1|^3] \frac{\sum_{j=1}^p |\hat w_j|^3}{\|\hat w\|^3}  \leq \E[|z_1|^3] \frac{\|\hat w\|_\infty}{\| \hat w\|}
    \to 0\,.
\end{equation*}
Thus, Lyapunov's condition is satisfied, and the claim follows from the CLT for triangular arrays.
\end{proof}
Consequently, for any $\alpha,\beta\in(0,1)$, choosing $\tau=\hat\sigma Q^{-1}(\alpha)$ yields $P_{FA}\to\alpha$, whereas choosing $\tau=\hat\mu+\hat\sigma Q^{-1}(1-\beta)$ yields $P_{MD}\to\beta$, almost surely.
}

\subsection{Isotropic data}\label{sec:supp_linear_isotropic} We now investigate a setting in which computations can be performed in closed form---the case of isotropic data, i.e., with $\Sigma=\sigma^2 I$, with $\sigma^2>0$, \arev{cf.~Example \ref{example:supp_isotropic_data}. In this case, it turns out that the rate only depends on the $\SNR$ and the \emph{regularization-to-noise ratio} $\RNR:=\lambda/\sigma^2$.}
\begin{prop}\label{prop:supp_linear_rate_isotropic}
	\arev{Under the assumptions of Theorem \ref{thm:rate_linear}, assume further that $\Sigma = \sigma^2 I$ and let $r:= \RNR = \lambda/\sigma^2$}. Then, \EC{New cool simplification in the isotropic regime.}
    \begin{equation}\label{eq:supp_isotropic_rate}
    R_{\gamma,\lambda} = \SNR   \frac{ \sqrt{(1+\gamma+r)^2-4\gamma}}{1+\gamma+r+
    \sqrt{(1+\gamma+r)^2-4\gamma}}\,.
    \end{equation}
\end{prop}
\begin{proof}
    Recall from Example \ref{example:supp_isotropic_data} that $H= \delta_{\sigma^2}$ and $\lSNR \diff H  = \SNR \delta_{\sigma^2}$. Thus, the observables in \eqref{eq:def_observables} read as in \eqref{eq:supp_observables_isotropic}. Additionally, the Silverstein equation admits an explicit solution given in \eqref{eq:supp_phi_isotropic}, \arev{which yields
    \begin{equation}
    \kappa(\lambda) = \frac{ \lambda+\sigma^2(\gamma-1) +\sqrt{\bigl(\lambda+\sigma^2(\gamma-1)\bigr)^2 +4\lambda\sigma^2}}{2}\,.
    \end{equation}}
    Altogether, \eqref{eq:mean_var_linear} simplifies significantly resulting in the following: 
\begin{equation}\label{eq:supp_asymptotic_mean_variance_linear}
\mu_\infty =  \frac{\kappa(\lambda)}{\lambda  } \frac{\sigma^2 \SNR}{(\sigma^2 + \kappa(\lambda))}\,, \quad \text{and} \quad \sigma_\infty^2 = \frac{\kappa(\lambda)^2}{\lambda^2  }  \frac{\SNR \sigma^4}{\big((\sigma^2 + \kappa(\lambda))^2 - \gamma \sigma^4\big)} \,,
\end{equation}
By computing $R_{\gamma, \lambda} = \frac{1}{2}\mu_\infty^2 / \sigma_{\infty}^2$, we obtain the claim \arev{after simple algebraic manipulations.}
\end{proof}

\subsection{Ridgeless case}\label{sec:supp_linear_ridgeless}

\arev{We now investigate what we refer to as the \emph{ridgeless regime}, i.e., the case $\lambda=0$, yielding $\hat T_0(x) := \hat s^T \hat \Sigma^{-1}x$. Here, we restrict our analysis to $\gamma\in(0,1)$, since for $\gamma>1$ the sample covariance matrix is singular. We leverage the analysis of the ridge-regularized detector by observing that $\hat T_0(x)=\lim_{\lambda \to 0^+} \hat s^T\hat\Sigma_\lambda^{-1}x$. To study the corresponding error probability decay as $p\to +\infty$, we compute the limit of $R_{\gamma, \lambda}$ as $\lambda \to 0^+$ and show, using the Moore--Osgood theorem, that this indeed coincides with the rate for $\hat T_0$.}

\arev{We first recall the following consequence of the
Silverstein fixed-point equation.}
\begin{lem}\label{lem:supp_limits_phi}
	Let $\varphi$ be a function satisfying the Silverstein equation \eqref{eq:silverstein_fpe} and suppose that Assumption \ref{ass:1} holds. Then
	\begin{enumerate}
		\item If $\gamma > 1$, we have $\lim_{\lambda \to 0^+} \lambda \varphi(-\lambda) = 0$.
		\item If $\gamma \in (0, 1)$ it holds $\lim_{\lambda \to 0^+} \varphi(-\lambda) = +\infty$ and $\lim_{\lambda \to 0^+} \lambda \varphi(-\lambda) = \arev{1 - \gamma}$. 
	\end{enumerate}
\end{lem}
\arev{We use this to establish the following result. We state the result with fixed false alarm level, but the same holds by switching the roles of $P_{MD}$ and $P_{FA}$.}\EC{I removed the case $\gamma > 1$ which was much more delicate than what I wrote.}

\begin{prop}\label{prop:supp_linear_rate_ridgeless}
Suppose that Assumption \ref{ass:1} holds. Then, in the ridgeless regime, as $p, n\to +\infty$ with $\gamma = p / n \in (0, 1)$, we get, almost surely, 
	\begin{equation}
		\lim_{p\to +\infty} \frac{1}{p} \log P_{MD} = - R_{\gamma, 0}\,, \quad R_{\gamma, 0}  :=
	\frac{\SNR}{2} (1-\gamma) \quad \text{for} \ \gamma \in (0, 1)\,.
	\end{equation}
\end{prop}
\begin{proof}
	\arev{First, observe that the rate for the ridgeless decision statistic $T_0$ coincides with
    \begin{equation*}
        R_{\gamma, 0}:=\lim_{p \to +\infty} \lim_{\lambda\to 0^+} \hat d^2 / (2p)\,.
    \end{equation*}
    To characterize this, we compute the limit $\lim_{\lambda\to 0^+} \lim_{p \to +\infty} \hat d^2 / (2p) = \lim_{\lambda\to 0^+} R_{\gamma, \lambda}$, and show, using the Moore--Osgood theorem, that we can exchange the two limits.} From Lemma \ref{lem:supp_limits_phi} and Example \ref{example:supp_k_0}, we get $\kappa(\lambda) \to 0$, and
	\begin{equation}\label{eq:supp_limlimmeanlinear}
	\begin{aligned}
	\lim_{\lambda\to 0^+}\lim_{p\to +\infty} \frac{\hat \mu}{p} & = \lim_{\lambda \to 0^+} \frac{\kappa(\lambda)}{\lambda  } \n_1(\kappa(\lambda)) = \frac{\SNR}{1-\gamma}\,.
	\end{aligned}
	\end{equation}

	For all $p$ large enough, the convergence of $\hat \mu / p$ to its limit as $\lambda \to 0^+$ is uniform in $p$ since, using \eqref{eq:supp_proof_norm_of_z} and standard
    extreme-eigenvalue results for sample covariance
    matrices \cite{by93limit} for $\gamma \in (0, 1)$, we get $\|\hat \Sigma_{\lambda}^{-1}\|\leq C^{-1}$, and $\|\hat \Sigma_{\lambda}^{-2}\| \leq C^{-2}$ almost surely for $C$ independent on $p$ and $\lambda$, and
	\begin{equation*}
	\left|\frac{d}{d\lambda} \frac{\hat \mu}{ p} \right| =\left| \frac{d}{d\lambda} \frac{1}{p} \hat s^T \hat \Sigma_{\lambda}^{-1}s \right| = \left|  \frac{1}{p} \hat s^T \hat \Sigma_{\lambda}^{-2} s \right|   \leq \frac{1}{C^2} \frac{1}{p} \|s\|^2 = \mathcal{O}(1)\,, \quad \text{as} \ p\to +\infty\,.
	\end{equation*}
	Let us now discuss the asymptotic behavior of the variance as $p\to+\infty$ in the ridgeless regime. In this case, we get, by dominated convergence theorem, 
	\begin{equation}\label{eq:supp_limlimsiglinear}
	\begin{aligned}
	&\lim_{\lambda\to 0^+}\lim_{p\to +\infty} \frac{\hat \sigma^2}{p} = \lim_{\lambda \to 0^+} \frac{\kappa(\lambda)^2}{\lambda^2 } \frac{\n_2(\kappa(\lambda))}{(1 - \gamma \m_2(\kappa(\lambda)))}  = \frac{\SNR}{(1-\gamma)^3}\,.
	\end{aligned}
	\end{equation}
	Similarly to the step above, we have
    \begin{equation*}
    \left|\frac{d}{d\lambda}\frac{\hat\sigma^2}{p}\right|
    =\frac{1}{p}\left|
    \hat s^T\left(
    \hat\Sigma_\lambda^{-2}\Sigma\hat\Sigma_\lambda^{-1}
    +\hat\Sigma_\lambda^{-1}\Sigma\hat\Sigma_\lambda^{-2}
    \right)\hat s
    \right|
    \leq \frac{2\|\Sigma\|}{C^3}
    \frac{\|\hat s\|^2}{p}
    =\mathcal{O}(1)\,,  \quad \text{as} \ p\to +\infty\,.
    \end{equation*}
    Thus, the convergence of $\hat \sigma^2 /p$ as $\lambda\to 0^+$ is uniform in $p$ and the Moore--Osgood allows us to exchange the two limits in  \eqref{eq:supp_limlimsiglinear}. Taking half the ratio between \eqref{eq:supp_limlimmeanlinear} and \eqref{eq:supp_limlimsiglinear} yields the claim.
\end{proof}

\subsection{Asymptotic values}\label{sec:supp_rate_limits}

The following result follows from a systematic inspection of the Silverstein equation in \eqref{eq:supp_silverstein_on_k}.

\begin{lem}\label{lem:limits_of_k_gamma}
	Suppose that Assumption \ref{ass:1} holds. Let $\kappa(\lambda)$ be the unique solution to the Silverstein equation \eqref{eq:supp_silverstein_on_k}. Then, for all $\lambda > 0$, the following limits hold:
	\begin{itemize}
		\item $\kappa(\lambda) \to \lambda$ as $\gamma \to 0^+$;
		\item $\kappa(\lambda) \sim \lambda + \gamma \E_H(\xi)$ as $\gamma \to +\infty$.
	\end{itemize}
\end{lem} 
\begin{proof}
First, we start with the limit in $\gamma \to 0^+$. Let $\gamma_h \to 0^+$ as $h\to +\infty$ and $\kappa_h$ the corresponding $\kappa(\lambda)$. Then, from the Silverstein equation \eqref{eq:supp_silverstein_on_k} we get that, for some $\bar \gamma \geq \gamma_h$ for all $h\in \N$, 
\begin{equation}
\kappa_h \leq \lambda + \bar \gamma \E_H(\xi) < +\infty\,.
\end{equation}
Thus, $(\kappa_h)_h$ is bounded. Passing to a subsequence if necessary, we shall assume that $\kappa_h \to \bar \kappa\geq 0$. Passing to the limit in the Silverstein equation \eqref{eq:supp_silverstein_on_k} we get
\begin{equation}
\bar \kappa = \lim_{h \to +\infty} \kappa_h =\lambda + \lim_{h \to +\infty} \gamma_h \int_{\R_+} \frac{\kappa_h \xi}{\kappa_h + \xi} \diff H =  \lambda\,. 
\end{equation}

Let us now show the asymptotic behavior for $\gamma\to +\infty$. Suppose now that $\gamma_h \to +\infty$. It is easy to see that assuming that $\kappa_h$ has a cluster point leads to an immediate contradiction (pass to the limit in the Silverstein equation as above). Then, it must be that $\kappa_h \to +\infty$ as $h \to +\infty$. Now, from the Silverstein equation we get
\begin{equation}
\frac{\kappa_h}{\gamma_h} = \frac{\lambda}{\gamma_h} + \int_{\R_+} \frac{\kappa_h \xi}{\kappa_h + \xi} \diff H \sim \frac{\lambda}{\gamma_h} + \E_H(\xi)\,, \quad \text{as} \ h \to +\infty\,,
\end{equation}
which leads to the claim.
\end{proof}

From Lemma \ref{lem:limits_of_k_gamma} we find that the rate \eqref{eq:rate_linear_general} for $\gamma \to 0$ and $\gamma \to +\infty$ has the following limits
\begin{equation}\label{eq:supp_def_R0_RInf}
\lim_{\gamma \to 0^+} R_{\gamma, \lambda} = \frac{1}{2}\frac{ \E_{ \lSNR \diff H}\big[f_\lambda(\xi)\big]^2}{ \E_{\lSNR \diff H}\big[f_\lambda(\xi)^2\big]}\,, \quad \text{and} \quad \lim_{\gamma \to +\infty } R_{\gamma, \lambda} = \frac{1}{2}\frac{ \E_{\lSNR \diff H}\big[\xi\big]^2}{ \E_{\lSNR \diff H}\big[\xi^2\big]}\,.
\end{equation}
where $f_\lambda(\xi) := \frac{\xi}{\xi + \lambda}$ for all $\xi \geq 0$. We are now ready to prove Proposition \ref{lem:dispersion_index}.

\begin{proof}[Proof of Proposition \ref{lem:dispersion_index}] Let us denote by $\mathsf{s} H := \lSNR \diff H$ and by $R_{0, \lambda}:= \lim_{\gamma \to 0^+}R_{\gamma, \lambda}$, i.e., the right-hand side of the first limit in \eqref{eq:supp_def_R0_RInf}. Take the derivative in $\lambda$ and differentiate under the integral sign to obtain
	\begin{equation*}
	\begin{aligned}
		\frac{d}{d\lambda} R_{0, \lambda} & = \frac{1}{2}\frac{2\E_{\mathsf{s} H}\big[f_\lambda(\xi)\big] \E_{\mathsf{s} H}\big[\frac{d}{d\lambda} f_\lambda(\xi)\big]\E_{\mathsf{s} H}\big[f_\lambda(\xi)^2\big] - \E_{\mathsf{s} H}\big[f_\lambda(\xi)\big]^2 \E_{\mathsf{s} H}\big[2 f_\lambda(\xi)\frac{d}{d\lambda} f_\lambda(\xi) \big]}{\E_{\mathsf{s} H}\big[f_\lambda(\xi)^2\big]^2}\\
	& =  -\frac{ \E_{\mathsf{s} H}\big[f_\lambda(\xi) \big]}{\E_{\mathsf{s} H}\big[f_\lambda(\xi)^2\big]^2} \Big( \E_{\mathsf{s} H}\big[f_\lambda(\xi)\tfrac{1}{\xi + \lambda}\big]\E_{\mathsf{s} H}\big[f_\lambda(\xi)^2\big] - \E_{\mathsf{s} H}\big[f_\lambda(\xi)\big] \E_{\mathsf{s} H}\big[f_\lambda(\xi)^2 \tfrac{1}{\xi + \lambda}\big]  \Big)\,.
	\end{aligned}
	\end{equation*}
    Now, let us introduce the auxiliary probability measure
    \begin{equation*}
    \mathbb{Q} := \frac{f_\lambda(\xi)^2}{\E_{\mathsf{s}H}\big[f_\lambda(\xi)^2\big]} \lSNR(\xi) \diff H(\xi) \,,  
    \end{equation*}
    and denote by $C_{\xi}:= \frac{1}{\xi +\lambda}$ for brevity. With this notation, we can write:
    \begin{equation*}
	\begin{aligned}
		\frac{d}{d\lambda} R_{0, \lambda} & =  -\frac{ \E_{\mathsf{s} H}\big[f_\lambda(\xi) \big]}{\E_{\mathsf{s} H}\big[f_\lambda(\xi)^2\big]} \Big( \E_{\mathsf{s} H}\big[f_\lambda(\xi)C_\xi\big] - \E_{\mathsf{s} H}\big[f_\lambda(\xi)\big] \frac{\E_{\mathsf{s} H}\big[f_\lambda(\xi)^2 C_\xi\big]}{\E_{\mathsf{s} H}\big[f_\lambda(\xi)^2\big]} \Big)\\
        &  =  -\frac{ \E_{\mathsf{s} H}\big[f_\lambda(\xi) \big]}{\E_{\mathsf{s} H}\big[f_\lambda(\xi)^2\big]} \Big( \E_{\mathsf{s} H}\big[f_\lambda(\xi)^2\big] \E_{\mathbb{Q}}\big[ C_\xi / f_\lambda(\xi)\big]  - \E_{\mathsf{s} H}\big[f_\lambda(\xi)\big]\E_{\mathbb{Q}}\big[ C_\xi\big] \Big)\\
        &  =  - \E_{\mathsf{s} H}\big[f_\lambda(\xi) \big]\Big( \E_{\mathbb{Q}}\big[ C_\xi / f_\lambda(\xi)\big]  - \frac{\E_{\mathsf{s} H}\big[f_\lambda(\xi)\big]}{{\E_{\mathsf{s} H}\big[f_\lambda(\xi)^2\big]} }\E_{\mathbb{Q}}\big[ C_\xi\big] \Big)\\
         &  =  - \E_{\mathsf{s} H}\big[f_\lambda(\xi) \big]\Big( \E_{\mathbb{Q}}\big[ C_\xi / f_\lambda(\xi)\big]  -  \E_{\mathbb{Q}}\big[1 / f_{\lambda}(\xi)\big]\E_{\mathbb{Q}}\big[ C_\xi\big] \Big)\,.
	\end{aligned}
	\end{equation*}
	This yields that for all $\lambda > 0$
	\begin{equation}\label{eq:supp_decrease_R0lam}
		\frac{d}{d\lambda} R_{0, \lambda} = - \E_{\mathsf{s} H}\big[f_\lambda(\xi) \big] \,\mathrm{Cov}_{\mathbb{Q}}\big(1/f_\lambda(\xi),\, C_{\xi}\big)\,.
	\end{equation}
	Since $\E_{\mathsf{s} H}\big[f_\lambda(\xi) \big] \geq 0$ due to the fact that $H$ is supported on $\R_+$, and $1/f_\lambda(\xi)$ and $C_{\xi}$ are both decreasing in $\xi$, by Chebyshev’s association inequality, we deduce that $\tfrac{d}{d\lambda}R_{0, \lambda} \leq 0$. In particular, $R_{0, \lambda}$ is monotonically non-increasing in $\lambda$. Thus, for each $\lambda>0$, 
	\begin{equation}
	\lim_{\gamma \to 0^+} R_{\gamma, \lambda} = R_{0, \lambda}  \geq \lim_{\lambda\to +\infty} R_{0, \lambda}  =  \frac{1}{2}\frac{ \E_{\mathsf{s} H}\big[\xi\big]^2}{ \E_{\mathsf{s} H}\big[\xi^2\big]} = \lim_{\gamma \to +\infty} R_{\gamma, \lambda}\,,
	\end{equation}
	where the second identity follows immediately by inspection and the first inequality is indeed an equality if and only if $\mathsf{s} H$ is supported on a single eigenvalue.
\end{proof}

\subsection{Linear decision statistic with miss-specified mean}\label{sec:supp_linear_miss}

In this section, we assume that the signal in the training set, denoted by $\mathsf{s}$, does not correspond to the signal in the test set, denoted by $s$---we say that it is \emph{miss-specified}. We thus consider $\hat T(x) =\hat{\mathsf{s}}^T \hat \Sigma_{\lambda}^{-1} x$, for which
\begin{equation}
\begin{aligned}
\hat \mu &=  \E \left[\hat T(x) \mid \hat X , {\cal H}_1\right] =  \hat{\mathsf{s}}^T \hat \Sigma_\lambda^{-1}s = \trace \big[s \hat{\mathsf{s}}^T \hat \Sigma_\lambda^{-1} \big]\,,\\
\hat \sigma^2 & = \E \left[ (\hat T(x) - \hat \mu)^2 \mid \hat X , \cH_1\right] =  \hat{\mathsf{s}}^T \hat \Sigma_\lambda^{-1} \Sigma \hat \Sigma_\lambda^{-1} \hat{\mathsf{s}} = \trace \big[ \hat{\mathsf{s}} \hat{\mathsf{s}}^T \hat \Sigma_\lambda^{-1} \Sigma \hat \Sigma_\lambda^{-1}\big]\,.
\end{aligned}
\end{equation}
Normalizing by $p$ and passing to the limit as $p, n\to +\infty$ with $p/n =\gamma$ we can obtain, using the same arguments of Proposition \ref{prop:mean_var_linear}, closed form expressions for the limits. Specifically, we need to adapt Assumption \ref{ass:A2} to this specific setting, requiring, in addition to the fact that Assumption \ref{ass:1} holds for both $s$ and $\mathsf{ s}$, the following coupling condition. 
\begin{assumption*}[Mismatched means]\label{ass:mismatch}
	\arev{The norm of the signals} $s, \mathsf{s} \in \R^p$ grow as $\mathcal{O}(p^{1/2})$ as $p\to +\infty$ and there exists a signed measure $\nu_{s,  \mathsf{s}}$ such that
	\begin{equation}
	\frac{1}{p}\sum_{i=1}^p (v_i^T s)(v_i^T  \mathsf{s} ) \delta_{\xi_i} \rightharpoonup \nu_{s,  \mathsf{s} }\,, \quad \text{as} \ p\to +\infty\,,
	\end{equation}
	where $v_i$ is the eigenvector of $\Sigma$ with unit norm associated to $\xi_i$. Additionally, $\nu_{s,  \mathsf{s} } \ll H$ with Radon--Nikodym derivative $\xi \mapsto \xi \mlSNR(\xi)$.
\end{assumption*}
Similarly to Definition \ref{def:snr}, we refer to $\mlSNR(\xi) = \frac{1}{\xi} \frac{\partial \nu_{s,  \mathsf{s}} }{\partial H}(\xi)$ as the \emph{mismatched directional signal-to-noise} ratio, which is a new not necessarily positive function that models the limiting behavior of $(v_i^T s)(v_i^T  \mathsf{s} ) / \xi_i$. Obviously, the further away from $\lSNR$, the more the model is mismatched. 

\begin{figure}[t]
	\centering
	\begin{subfigure}[t]{1\textwidth}
		\centering
		\includegraphics[width=\textwidth]{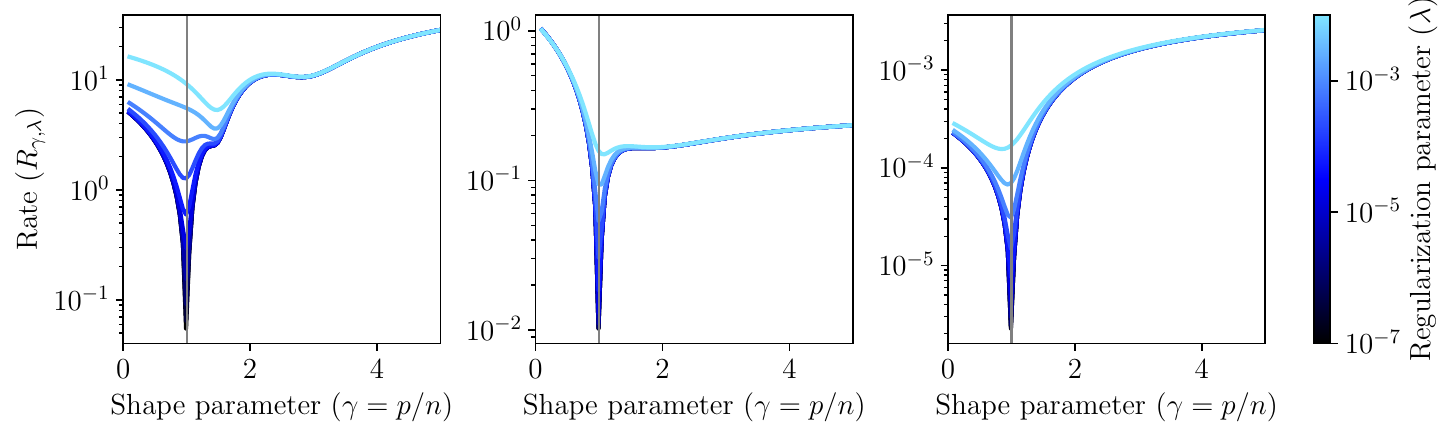}
		\caption{Performances as a function of $\gamma$.}
		\label{fig:supp_linear_miss_general_gamma}
	\end{subfigure}
	\hfill
	\begin{subfigure}[t]{0.8\textwidth}
		\centering
		\includegraphics[width=\textwidth]{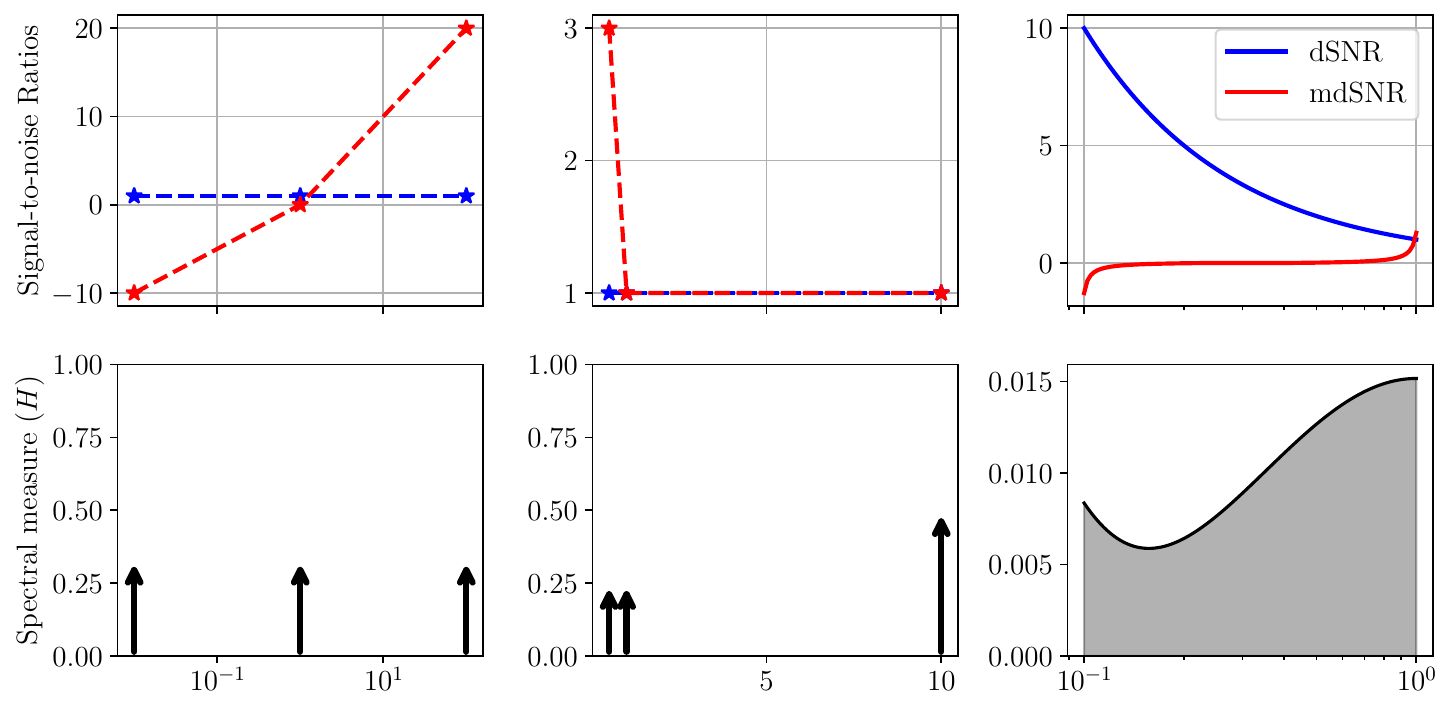}
		\caption{Geometric setting considered to produce the results in Figure \ref{fig:supp_linear_miss_general_gamma}. First row: signal-to-noise ratios $\lSNR$ and its miss-specified variant $\mlSNR$. Second row: Population spectral distribution $\diff H$.}
		\label{fig:supp_linear_miss_general_data}
	\end{subfigure}
	\caption{Rate for the linear decision statistic defined in \eqref{eq:def_That_linear_unknown} under miss-specification of the mean.}
	\label{fig:supp_linear_miss_general}
\end{figure}

\paragraph{Non-isotropic setting} Using the same techniques as in the proof of Proposition \ref{prop:mean_var_linear} we can see that the asymptotic variance of $\hat T$ coincides with $\sigma_{\infty}^2$ with $\lSNR$ computed with respect to $ \mathsf{s} $, while the mean reduces to
\arev{
\begin{equation}
\begin{aligned}
\frac{\hat\mu}{p} &= \trace\left[ \frac{s\hat{\mathsf{s}}^T+\hat{\mathsf{s}}s^T}{2p} \hat\Sigma_\lambda^{-1} \right] = \frac{\kappa(\lambda)}{\lambda } \int_{\R_+} \frac{\xi}{\xi+\kappa(\lambda)} \frac{(v_i^Ts)(v_i^T\hat{\mathsf{s}})}{\xi} \diff H_p(\xi)+o(1)\\
&= \frac{\kappa(\lambda)}{\lambda} \int_{\R_+} \frac{\xi}{\xi+\kappa(\lambda)} \mlSNR(\xi) \diff H(\xi)+o(1)\,,
\end{aligned}
\end{equation}}%
almost surely as $p\to +\infty$. \arev{Since $\mu_{\infty}$ is not necessarily positive in this setting, from the asymptotic values $\mu_{\infty}$ and $\sigma^2_{\infty}$, we compute the rate as $R_{\gamma, \lambda} = \frac{1}{2} \max\{\mu_\infty, 0\}^2 / \sigma_\infty^2$}. We show in Figure \ref{fig:supp_linear_miss_general} three notable examples. \arev{In these cases, we can see that, as soon as $s \neq  \mathsf{s} $ and $\Sigma$ is non-isotropic, the best rate can be achieved in the overparameterized regime}.

\paragraph{Isotropic case} In the isotropic case, $\mlSNR$ is defined only on $\sigma^2$ and thus can be identified with a number, which is $\mSNR := \lim_{p\to + \infty} \frac{1}{p} \mathsf{s}^T s / \sigma^2$, \arev{which exists by Assumption \ref{ass:mismatch}}. Therefore, mean and variance now read as
\begin{equation}\label{eq:supp_asymptotic_mean_variance_miss_linear}
\mu_\infty = \frac{\mSNR \kappa(\lambda)}{\lambda } \frac{\sigma^2}{\sigma^2 + \kappa(\lambda)}\,, \quad \text{and} \quad \sigma_\infty^2 =  \frac{\tilde \SNR \kappa(\lambda)^2}{\lambda^2}  \frac{\sigma^4}{(\sigma^2 + \kappa(\lambda))^2 - \gamma \sigma^4} \,,
\end{equation}
where \arev{$\tilde \SNR:= \lim_{p\to +\infty} \tfrac{1}{p}\|\mathsf{s}\|^2/\sigma^2$, which exists by Assumption \ref{ass:1} for $\mathsf{s}$}. We deduce that the asymptotic rate has the same shape as in \eqref{eq:supp_isotropic_rate}, but with a different constant
\begin{equation}\label{eq:supp_rate_linear_miss}
R_{\gamma, \lambda} =  \frac{\mSNR^2}{2 \tilde \SNR}  \bigg(1 - \frac{\gamma \sigma^4}{ (\sigma^2 +  \kappa(\lambda))^2} \bigg) =  \mathcal{K} \frac{\SNR}{2}  \bigg(1 - \frac{\gamma \sigma^4}{ (\sigma^2 + \kappa(\lambda))^2} \bigg)\,,
\end{equation}
where we have denoted by $\mathcal{K}$ the constant
\begin{equation}
\mathcal{K} :=  \frac{\mSNR^2}{  \SNR \cdot \tilde \SNR} =  \lim_{p\to +\infty} \Big[\frac{\mathsf{s}^T s}{\|\mathsf{s}\| \|s\|}\Big]^2 \in [0, 1]\,.
\end{equation}
First, observe that $\mathcal{K}$ coincides with the squared limiting \emph{cosine similarity} between $\mathsf{s}$ and $s$, quantifying the degree of miss-specification. Secondly, compared to \eqref{eq:supp_isotropic_rate}, we now have a loss in performances quantifiable by $\mathcal{K}$.

\begin{figure}
	\centering
	\includegraphics[width=\textwidth]{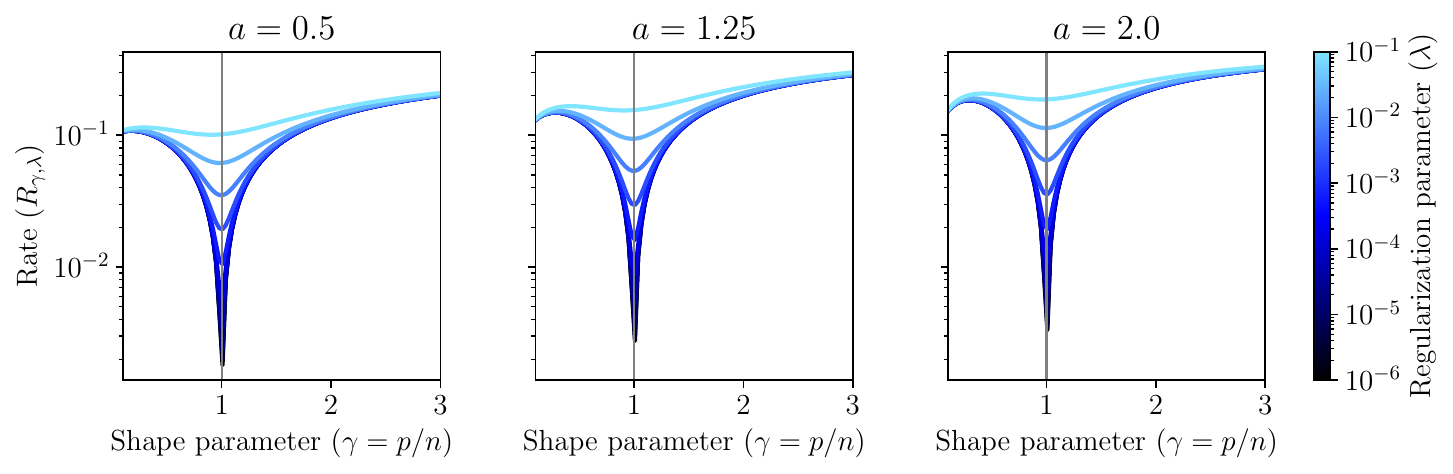}
	\caption{ Isotropic case: Rate for the linear decision statistic defined in \eqref{eq:def_That_linear_unknown} assuming a miss-specified mean. The \emph{artificial miss-specification} degree is polynomially decreasing in $\gamma$ as in \eqref{eq:supp_decay_miss} with $c=0.9$ and three different choices of $a$. Confer Appendix \ref{sec:supp_linear_miss} for details.}
	\label{fig:supp_linear_miss}
\end{figure}

Following \cite{hmrt22high}, if we assume that the degree of dissimilarity decays as the resolution of the model increases, e.g., supposing that $\mathcal{K}$ is a function of $\gamma$ as
\begin{equation}\label{eq:supp_decay_miss}
\mathcal{K}(\gamma) :=  1 - \frac{c}{(1 + \gamma)^a} \,, \quad \text{for all} \ \gamma > 0\,,
\end{equation}
and two constants $a > 0$ and $c \in (0, 1)$, we obtain the curves shown in Figure \ref{fig:supp_linear_miss}. In Figure \ref{fig:supp_linear_miss}, we observe a full \emph{double-descent}-type behavior, where $R_{\gamma, \lambda}$ exhibits a local maximum in $(0, 1)$ and increases again after $\gamma=1$.

\begin{figure}[t]
	\centering
	\begin{subfigure}{0.22\linewidth}
		\includegraphics[width=1\linewidth]{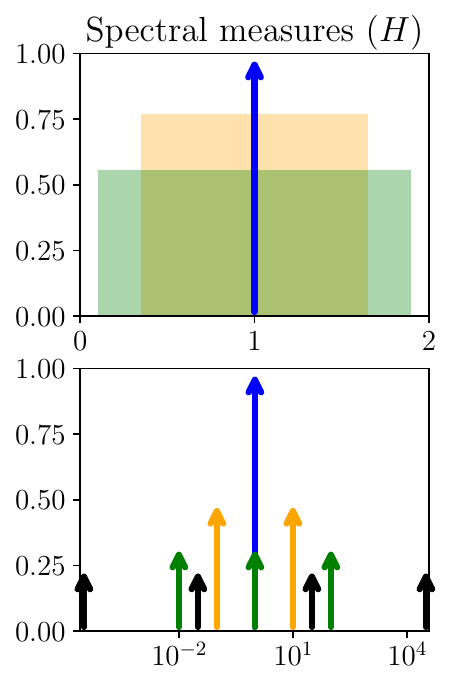}
	\end{subfigure}
	\begin{subfigure}{0.24\linewidth}
		\includegraphics[trim={15cm 0cm 0cm 0cm}, clip, width=1\textwidth]{Figures/rate_linear_vs_lam.pdf}
	\end{subfigure}
	\caption{Population spectral distribution $H$ and $\lSNR$ for the three examples shown in Figure \ref{fig:simulated_data}. \textbf{Left:} Spectral measures for the two cases shown on the left of Figure \ref{fig:simulated_data}. \textbf{Right:} Spectral measure for the case on the right of Figure \ref{fig:simulated_data}. Vertical arrows represent Dirac masses, while shaded rectangles indicate uniform distributions. Color-coding is consistent with the corresponding curves in Figure \ref{fig:simulated_data}.}
	\label{fig:supp_simulated_data_geometry}
\end{figure}

To conclude this section, let us make the population spectral distribution $H$ and the $\lSNR$ of Figure \ref{fig:simulated_data} explicit in Figure \ref{fig:supp_simulated_data_geometry}.

\subsection{Extension to the multi-hypotheses setting}\label{sec:multi_hypotheses}

Let us now consider the multi-hypotheses setting of \eqref{eq:intro_multi_hypotheses_testing} and the decision statistic
\begin{equation}
\hat k (x) := \argmax_{0 \le i \le M-1} \hat{p}_i(x)\,, \quad \text{where} \quad \hat p(x) := \softmax(\hat \Omega x + \hat b)\,,
\end{equation}
and $\hat \Omega_i : = \hat s_i^T \hat \Sigma^{-1}_\lambda$,  $\hat b_i := \log \pi_i - \frac{1}{2} \hat s_i^T \hat \Sigma_{\lambda}^{-1}\hat s_i$ for all $i= 0, \dots, M-1$.

\paragraph{Error probability decay} Assume for simplicity that $\pi_i = 1/M$ for all $i= 0, \dots, M-1$. In this case, the probability of error for class $\cH_k$ can be bounded as follows. Let $\hat T_i (x) := \hat s_i^T \hat \Sigma^{-1}_\lambda x + \hat b_i $ for all $i = 0, \dots, M-1$. Then, using the union bound and that $x \sim s_k + \Sigma^{1/2}z$ under $\cH_k$, 
\begin{equation*}
\begin{aligned}
&\P(\hat k (x) \neq k \mid  \hat X, \cH_k) = \P(\text{There exists} \ j \neq k : \hat T_j(x) > \hat T_k(x) \mid \hat X, \cH_k) \\
& = \P\bigg( \bigcup_{j \neq k} \{ \hat T_j(x) \geq \hat T_k(x) \}   \mid  \hat X, \cH_k \bigg) \leq \sum_{j\neq k} \P\big(\hat T_j(x) \geq \hat T_k(x) \mid \hat X, \cH_k \big)\\
& = \sum_{j\neq k} \P\Big(\hat s_j^T \hat \Sigma^{-1}_\lambda (s_k + \Sigma^{1/2}z)  - \frac{1}{2} \hat s_j^T \hat \Sigma_{\lambda}^{-1}\hat s_j \geq \hat s_k^T \hat \Sigma^{-1}_\lambda (s_k + \Sigma^{1/2}z)  - \frac{1}{2} \hat s_k^T \hat \Sigma_{\lambda}^{-1}\hat s_k \mid \hat X, \cH_k \Big)\\
& \leq \sum_{j\neq k} \P\Big( \hat s_{jk} ^T \hat \Sigma^{-1}_\lambda \Sigma^{1/2}z \geq \frac{1}{2}\hat s_{jk}^T \hat \Sigma^{-1}_\lambda \hat s_{jk}  +  \hat s_{jk}^T \hat \Sigma^{-1}_\lambda (\hat s_k - s_k) \mid \hat X, \cH_k \Big)\,,
\end{aligned}
\end{equation*}
where $\hat s_{jk} := \hat s_{j} - \hat s_k$ for all $j\neq k$. Using now that $\frac{1}{p}\hat s_{jk}^T \hat \Sigma^{-1}_\lambda (\hat s_k - s_k)  \to 0$ almost surely, and the same bounds as in Section \ref{sec:linear} with $\hat s$ replaced by $\hat s_{jk}$, we get as $p\to +\infty$
\begin{equation*}
\begin{aligned}
&\P(\hat k (x) \neq k \mid \hat X, \cH_k)\\
&\leq \sum_{j\neq k} \P\Big( \frac{1}{p} \hat s_{jk}^T \hat \Sigma^{-1}_\lambda \Sigma^{1/2}z \geq   \frac{1}{2p} \hat s_{jk}^T \hat \Sigma^{-1}_\lambda \hat s_{jk} +  \frac{1}{p}\hat s_{jk}^T \hat \Sigma^{-1}_\lambda (\hat s_{k} - s_k) \mid \hat X, \cH_k \Big)\\
& = \sum_{j\neq k}  \exp \bigg\{ -\frac{p}{2} \left(\frac{\mu_{jk, \infty}}{2\sigma_{jk, \infty}} \right)^2 + o(p) \bigg\}  \lesssim \exp\Big\{ - p \left(R_{\gamma, \lambda}^{(k)} +o(1)\right) \Big\}\,, 
\end{aligned}
\end{equation*}
where we have denoted by
\begin{equation}
    \arev{R_{\gamma, \lambda}^{(k)}: =  \frac{1}{8}\min_{j \neq k} \left(  \frac{\mu_{jk, \infty}}{\sigma_{jk, \infty}} \right)^2}\,,
\end{equation}
and $\mu_{jk, \infty}$ and $\sigma_{jk, \infty}^2$ are the almost sure limits of mean and variance (normalized by $p$) of the decision statistic $\hat T_{jk}(x) := \hat s_{jk}^T \hat \Sigma_{\lambda}^{-1} x$ with $x \sim \cN(0, \Sigma)$ under $\cH_0$ and $x \sim \cN(s_{jk}, \Sigma)$ under $\cH_1$.

\paragraph{Comparison with baseline} We evaluate the classification performance of the proposed multi-hypothesis decision rule against a fully connected network (FCN) baseline. 
Using the entire FireRisk dataset with 7 distinct classes, we first perform feature extraction using the DINOv3 model. To investigate model behavior across varying complexity levels, we systematically downsample the training set to span shape parameter values $\gamma = p/n$ from $0.1$ to $4.0$ to cover both the underparameterized and overparameterized regimes.

The FCN baseline consists of a single linear layer with softmax output, equivalent to multi-class logistic regression. The proposed decision rule is based on a multi-class Gaussian, and consists of a linear layer (different weight compared to the FCN) and softmax output. The decision for both the classifiers is the MAP rule, which selects the hypothesis with the largest posterior probability.   

The FCN is trained with stochastic gradient descent (SGD) and a weighted cross-entropy loss to address class imbalance. Training is performed in full-batch mode for 300 epochs with a learning rate of $0.01$ and ridge regularization ($\lambda_{L2} = 10^{-3}$). We observed that the FCN performance was insensitive to the choice of $\lambda_{L2}$ within the tested range ($10^{-5}, 10^{-3}, 10^{-1}$).

\begin{figure}
	\centering
	\includegraphics[width=0.5\textwidth]{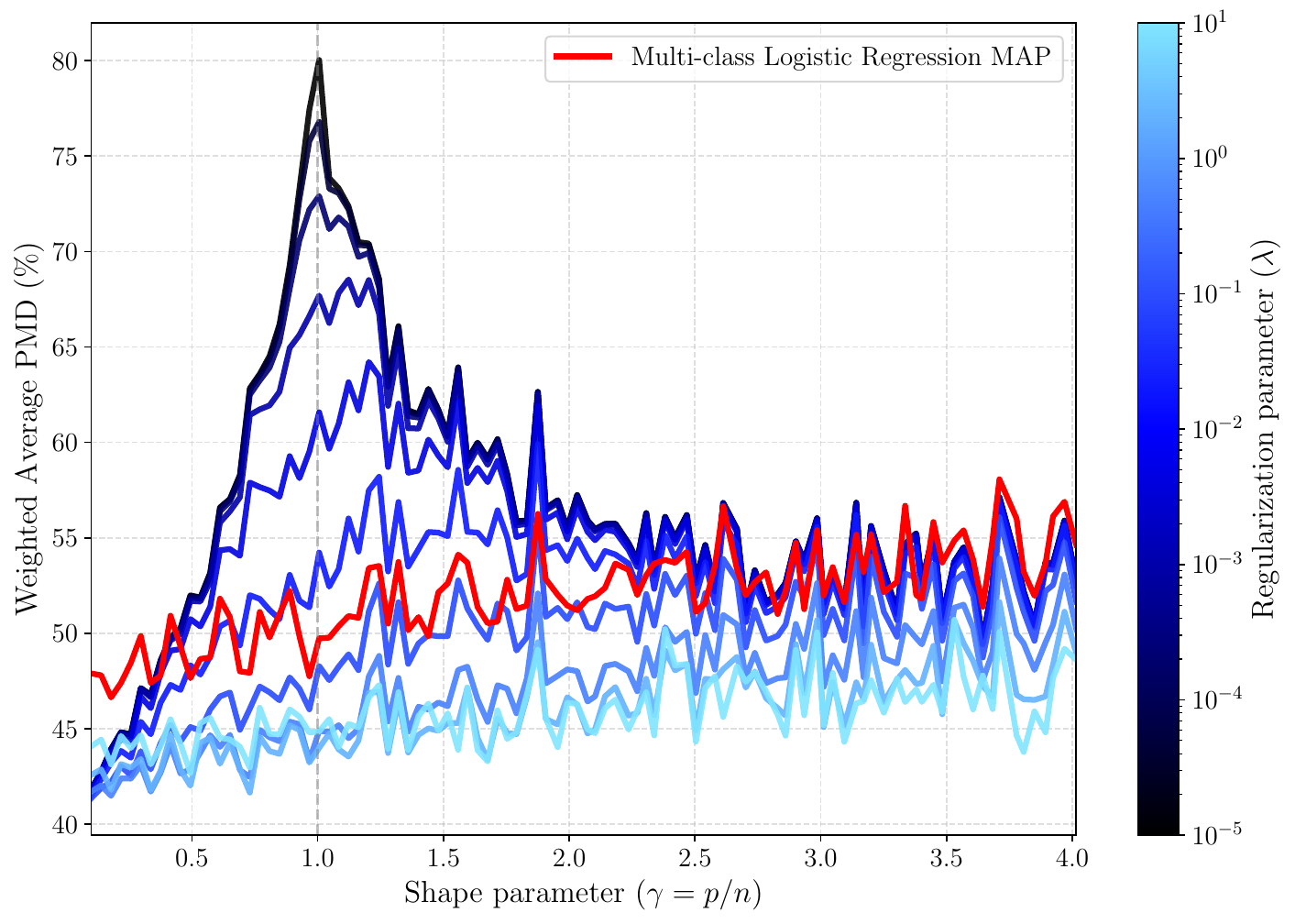}
	\caption{Classification performance on multi-hypotheses setting. We compare the weighted probability of miss detection of the proposed multi-class Gaussian MAP framework against the multi-class logistic regression baseline as a function of the shape parameter $\gamma = p/n$ on the FireRisk dataset ($M=7$). The Gaussian MAP framework with regularization parameters $\lambda \in \{0.1, 10, 100\}$ (bold solid lines) achieves significantly lower error rates. The results demonstrate the robustness of the multi-hypothesis approach compared to standard linear baseline.}
	\label{fig:multi_class_results}
\end{figure}

Performance is evaluated using the weighted probability of miss-classification, which accounts for class imbalances by weighting the error rate of each class by its respective prior probability We show the results in Figure \ref{fig:multi_class_results}. Much like in Figure \ref{fig:nebula_roc}, compared to the baseline, our approach exhibits comparable or better performances across all values of $\gamma > 0$ yet requiring essentially no fine-tuning.

\section{Missing proofs and further insights on Section \ref{sec:quadratic}}\label{sec:supp_quadratic}

In this section, we consider the case in which no $\cH_1$ data is available in the training set, and we aim solving \eqref{eq:intro_hypotheses_testing} by means of the decision statistic defined in \eqref{eq:def_T_quadratic}.

\subsection{High-dimensional mean and variance}

\arev{The following classical identities for Gaussian quadratic forms can be found in \cite[Chapter 2]{s97linear}.}
\begin{lem}\label{lem:supp_moment_of_quadratic_form}
	Let $x \sim \mathcal{N}(\mu, \Sigma)$ where $\mu \in \mathbb{R}^p$ and $\Sigma \in \mathbb{R}^{p \times p}$ is symmetric positive definite.  
	Let $A \in \mathbb{R}^{p \times p}$ be a symmetric positive semidefinite matrix. Then
	\begin{equation}
	\E[x^T A x] =  \trace (A\Sigma) + \mu^T A \mu \,, \quad \text{and} \quad \Var(x^T A x) = 2\, \trace(A\Sigma A\Sigma) + 4 \mu^T A\Sigma A \mu\,.
	\end{equation}
\end{lem}
\arev{We employ these formulas to establish Proposition \ref{prop:mean_var_quadratic} below:}

\begin{proof}[Proof of Proposition \ref{prop:mean_var_quadratic}]
	Using Lemma \ref{lem:supp_moment_of_quadratic_form}, we obtain
	\begin{equation}
	\hat \mu_{0, 1} =  \trace (\hat \Sigma^{-1}_\lambda \Sigma) + \mu_{0, 1}^T \hat \Sigma^{-1}_\lambda \mu_{0, 1} \,, \quad \text{and} \quad \hat \sigma_{0, 1}^2 = 2\, \trace(\hat \Sigma^{-1}_\lambda\Sigma \hat \Sigma^{-1}_\lambda\Sigma) + 4 \mu_{0, 1}^T\hat \Sigma^{-1}_\lambda \Sigma \hat \Sigma^{-1}_\lambda \mu_{0, 1}\,,
	\end{equation}
	with $\mu_{0}=0$ under $\cH_0$ and $\mu_{1}=s$ under $\cH_1$. Arguing as in Proposition \ref{prop:mean_var_linear} and using systematically Fact \ref{fact:supp_observables_as_trace_limits}, Theorem \ref{thm:rmb} yields the following. For the mean under $\cH_0$:
	\begin{equation}\label{eq:supp_quadratic_form_mu_inf_0}
	\frac{\hat \mu_{0}}{p}  \sim \frac{\kappa(\lambda)}{\lambda  }\frac{1}{p}\trace\big[ \Sigma \big(\Sigma +\kappa(\lambda) I\big)^{-1}\big] \sim \frac{\kappa(\lambda)}{\lambda } \m_1(\kappa(\lambda))\,.
	\end{equation}
	For the variance under $\cH_0$, instead, we get
	\begin{equation}\label{eq:supp_quadratic_form_sig2_inf_0}
	\begin{aligned}
	\frac{\hat \sigma_{0}^2}{p} & \sim \frac{2\kappa(\lambda)^2}{\lambda^2 } \frac{1}{p} \trace\big[ \Sigma \big(\Sigma + \kappa(\lambda) I\big)^{-1}\Sigma \big(\Sigma +  \kappa(\lambda) I\big)^{-1}\big] \\
	& \quad + \frac{2\kappa(\lambda)^2}{\lambda^2 }  \frac{1}{p} \trace\big[ \Sigma \big(\Sigma + \kappa(\lambda) I\big)^{-2}\Sigma\big]^2 \frac{1}{n - \diff f_2(\kappa(\lambda))} \\
	& \sim \frac{2\kappa(\lambda)^2}{\lambda^2 } \m_2(\kappa(\lambda)) + \frac{2 \kappa(\lambda)^2}{\lambda^2  } \m_2(\kappa(\lambda))^2 \frac{p}{n - \trace\big( \Sigma^2(\Sigma + \kappa(\lambda))^{-2} \big)}\\
	& \sim \frac{2\kappa(\lambda)^2}{\lambda^2 } \m_2(\kappa(\lambda)) + \frac{2\kappa(\lambda)^2}{\lambda^2 } \m_2(\kappa(\lambda))^2 \frac{\gamma}{1 - \gamma \m_2(\kappa(\lambda))}\,,
	\end{aligned}
	\end{equation}
	which eventually yields \eqref{eq:mean_var_quadratic_H0}. While under $\cH_1$, applying once again Theorem \ref{thm:rmb}, we get
	\begin{equation}\label{eq:supp_quadratic_form_mu_inf_1}
	\frac{\hat \mu_{1}}{p} \sim \frac{\hat \mu_{0}}{p}  + \frac{\kappa(\lambda)}{\lambda  }\trace\bigg[ \frac{s s^T}{p} \big(\Sigma + \kappa(\lambda) I\big)^{-1}\bigg] \sim \frac{\kappa(\lambda)}{\lambda } \m_1(\kappa(\lambda)) + \frac{\kappa(\lambda)}{\lambda } \n_1(\kappa(\lambda))\,.
	\end{equation}
	Regarding the variance, we get,
	\begin{equation}\label{eq:supp_quadratic_form_sig2_inf_1}
	\begin{aligned}
	&\frac{\hat \sigma_{1}^2}{p}  \sim \frac{\hat \sigma_{0}^2}{p} + \frac{4\kappa(\lambda)^2}{\lambda^2  } \trace\bigg[ \frac{s s^T}{p} \big(\Sigma + \kappa(\lambda) I\big)^{-1}\Sigma \big(\Sigma + \kappa(\lambda) I\big)^{-1}\bigg]\\
	& \qquad + \frac{4\kappa(\lambda)^2}{\lambda^2 } \trace\bigg[ \frac{s s^T}{p} \big(\Sigma + \kappa(\lambda) I\big)^{-2}\Sigma\bigg]\trace\big[\Sigma \big(\Sigma + \kappa(\lambda) I\big)^{-2}\Sigma\big]\frac{1}{n - \diff f_2(\kappa(\lambda))}\\
	& \sim \frac{\hat \sigma_{0}^2}{p} + \frac{4\kappa(\lambda)^2}{\lambda^2 }  \n_2(\kappa(\lambda)) + \frac{4\kappa(\lambda)^2}{\lambda^2 } \n_2(\kappa(\lambda)) \m_2(\kappa(\lambda)) \frac{p}{n - \trace\big( \Sigma^2(\Sigma + \kappa(\lambda))^{-2}\big)}\\
	& \sim \frac{\hat \sigma_{0}^2}{p} + \frac{4\kappa(\lambda)^2}{\lambda^2 }  \n_2(\kappa(\lambda)) + \frac{4 \kappa(\lambda)^2}{\lambda^2  } \n_2(\kappa(\lambda)) \m_2(\kappa(\lambda)) \frac{\gamma}{1 - \gamma \m_2(\kappa(\lambda))}\,.
	\end{aligned}
	\end{equation}
	Rearranging everything and plugging in the limits in the $\cH_0$ case, we obtain the claim. 
\end{proof}

\subsection{Rate in the LDP sense}\label{app:rate_quadratic}

Since $\hat T$ is not Gaussian in this quadratic setting, the performances of our detection mechanism cannot be explained simply by the ratio between asymptotic mean and asymptotic variance. We shall resort on the G\"artner--Ellis Theorem \cite{gartner1977, ellis84}. See Appendix \ref{app:supp_preliminaries_LDP} for non-exhaustive introduction to the topic, which nonetheless contains the most important tools for our analysis.

\subsubsection{Computing the limiting LMGF}

We first discuss the $\cH_0$ case, as the $\cH_1$ case can be deduced from the former. In this setting, it turns out that the ratio between regularization and noise plays a central role in determining the performances in the sense of LDP. We introduce the measure $\dRNR$ as the push-forward of $H$ under $f(x)=\frac{\lambda}{x}$, which, in relation to Assumption \ref{ass:1}, can be understood as the limiting spectral distribution of $\lambda \Sigma^{-1}$. Clearly in the isotropic case, $\dRNR = \delta_{\lambda / \sigma^2}$, in which case we denote $\RNR:= \lambda / \sigma^2$. 

\paragraph{The $\cH_0$ Case} \arev{Our objective is to identify the Fenchel--Legendre conjugate of $\Lambda^{(0)}$ according to \eqref{eq:def_Lambda}, which requires studying the (possibly infinite) almost sure limit of 
\begin{equation}\label{eq:supp_def_Lambda_p}
	\hat \Lambda^{(0)}(\tau):=	\frac{1}{p} \log \E\big[ \exp\{\tau \hat T(x)\} \mid \hat X, \cH_0 \big] \,, \quad \tau \in \R\,.
\end{equation}
In contrast to the linear setting of Section \ref{sec:linear}, to obtain a closed form solution for the rate in the sense of the LDP, we need a \emph{no-spectral-outliers} assumption on the \emph{whitened decision matrix}
\begin{equation}\label{eq:def_What}
	\hat W^{-1} := \Sigma^{-\frac{1}{2}}(\hat \Sigma + \lambda I)\Sigma^{-\frac{1}{2}} =  \tfrac{1}{n}\hat Z \hat Z^T  + \lambda \Sigma^{-1}\,,
\end{equation}
where we have denoted $\hat Z := \Sigma^{-\frac{1}{2}} \hat X$, which is a collection of i.i.d.~random variables with zero mean and unit variance. 
\begin{assumption*}[No spectral outliers]\label{ass:A3} Let $\hat W^{-1}$ be defined as in \eqref{eq:def_What}. Assume that its empirical spectral distribution $\hat \nu_{\gamma, \lambda}$ converges weakly to a probability distribution $\nu_{\gamma, \lambda}$ with 
\begin{equation}\label{eq:supp_no_spectral_outliers}
    \lambda_{\min}(\hat W^{-1}) \to T:= \inf \supp \nu_{\gamma, \lambda}\,, \quad  \text{a.s. as $p\to +\infty$}\,.
\end{equation}
\end{assumption*}
Note that Assumption \ref{ass:A3} holds for isotropic data. In terms of the underlying population geometry, \eqref{eq:supp_no_spectral_outliers} is strictly linked to $\|\Sigma\|\to \sup \supp H$. We formulate it as is to simplify the derivation of the following:}
\EC{Rotationally invariant with independent component $\to$ Gaussian (cf. Maxwell characterization of Gaussian distributions)
}
\begin{lem}\label{lem:supp_lmgf_H0}
	Suppose that \arev{Assumptions \ref{ass:1} and \ref{ass:A3} hold}. Then, \arev{for every $\tau \ne T/2$ the conditional almost-sure limit defining $\Lambda^{(0)}(\tau)$ exists and is given by:}
	\begin{equation}\label{eq:supp_limiting_lmgf}
	\Lambda^{(0)}(\tau) = \arev{-\frac{1}{2}} \int_{\R_+} \log\left(1 - \frac{2\tau}{\xi}\right) \diff \nu_{\gamma, \lambda} (\xi)\,, \quad \arev{\text{for all $\tau < \frac{T}{2}$}}\,, 
	\end{equation}
	\arev{and $\Lambda^{(0)}(\tau)= +\infty$ for $\tau > \frac{T}{2}$, where $T:=\inf \supp \nu_{\gamma, \lambda}$}, $\nu_{\gamma, \lambda}$ is the probability measure whose Stieltjes transform $\mathfrak{s}\colon \C_+ \to \C_+$ is the unique solution to the following fixed-point equation:
	\begin{equation}\label{eq:supp_silverstein_fpe_lmgf}
	\mathfrak{s}(z) = \mathfrak{s}_{\dRNR}\bigg(z - \frac{1}{1 + \gamma \mathfrak{s}(z)}\bigg)\,, \quad z \in \C_+\,,
	\end{equation}
	and $\mathfrak{s}_{\dRNR}$ is the Stieltjes transform of $\dRNR$.
\end{lem}
\begin{proof}  Using the definition of $\hat T$, we get, under $\cH_0$, 
	\begin{equation*}
		\hat T(x) = x^T \hat \Sigma_\lambda^{-1} x = x^T \Sigma^{-\frac{1}{2}} \Sigma^{\frac{1}{2}}\hat \Sigma_\lambda^{-1}\Sigma^{\frac{1}{2}}\Sigma^{-\frac{1}{2}} x = z^T \hat W z\;,
	\end{equation*}
	Therefore, denoting by $\tilde v=O z $, where $O$ is the (orthogonal) diagonalizing matrix of $\hat W$, we shall write, \arev{using that under $\cH_0$, the components $|\tilde v_i|^2$ are i.i.d.~$\chi^2$ variables with LMGF $t\mapsto -\frac{1}{2} \log(1-2t)$ for all $t< \frac{1}{2}$,}
	\begin{equation}\label{eq:supp_formula_Lambda0hat}
	\begin{aligned}
	\hat \Lambda^{(0)}(\tau)& = \frac{1}{p} \log \E\big[ \exp\{\tau z^T \hat W z\} \mid \hat X, \cH_0 \big] = \frac{1}{p} \log \E\bigg[ \exp\bigg\{ \sum_{i=1}^p   \frac{\tau}{\xi_i} | \tilde v_i |^2 \bigg\} \,\bigg| \hat X, \cH_0 \bigg]\\
	& = \frac{1}{p} \sum_{i=1}^p \log \E\Big[ \exp\Big\{\frac{\tau}{\xi_i} | \tilde v_i |^2 \Big\} \, \bigg| \hat X, \cH_0 \Big] = -\frac{1}{2}\int \log\bigg(1 - \frac{2\tau}{\xi} \bigg) \diff \hat \nu_{\gamma, \lambda}(\xi)\,,
	\end{aligned}
	\end{equation}
	where $\hat \nu_{\gamma, \lambda}$ is the spectral empirical distribution of \arev{$\hat W^{-1}$}. We can thus invoke \cite[Theorem 1.1]{sb95empirical}, which yields the weak convergence of $\hat \nu_{\gamma, \lambda}$ to the measure $\nu_{\gamma, \lambda}$ whose Stieltjes transform is the unique solution to \eqref{eq:supp_silverstein_fpe_lmgf}. \arev{Under Assumption \ref{ass:A3}, for each $\tau<T/2$, the empirical spectral measures $\hat\nu_{\gamma,\lambda}$ are almost surely eventually supported uniformly away from the singularity of the integrand, allowing us to pass to the limit in \eqref{eq:supp_formula_Lambda0hat} as $p\to+\infty$. Conversely, for each $\tau>T/2$, the corresponding finite-dimensional LMGF is almost surely eventually equal to $+\infty$.}
\end{proof}

\begin{rmk}[Boundary case]
    \arev{We assume that $\hat\Lambda^{(0)}(T/2)$ converges almost surely in the extended real line. Its limiting value depends on the spectral behavior at the boundary and must be determined on a case-by-case basis. However, this value does not affect the remainder of our analysis.}\EC{Pointwise limit must be lsc?}
\end{rmk}

We have obtained a second fixed-point equation \eqref{eq:supp_silverstein_fpe_lmgf}, that this time characterizes the limiting measure $\nu_{\gamma, \lambda}$. Note that $\mathfrak{s}$ is defined on $\C_+$, but can be obviously extended by continuity to $\R\setminus \supp \nu_{\gamma, \lambda}$, where by dominated convergence theorem still admits the integral representation:
\begin{equation}\label{eq:supp_Stieltjes_s_mulam}
\mathfrak{s}(z) = \int_{\R_+} \frac{1}{\xi - z} \diff \nu_{\gamma, \lambda}(\xi)\,, \quad \text{for all} \ z<T\,.
\end{equation}
For notational convenience, we do not rename $\mathfrak{s}$.

Although the integral in \eqref{eq:supp_limiting_lmgf} is somewhat challenging to compute, in practice, we can resort to an integration trick proposed by Silverstein and Bai, see e.g., \cite{bs10spectral}. It consist in writing the integrand as an integral in $\xi$ and applying Fubini's theorem. 
\begin{cor}\label{cor:supp_lmgf_integral_formula}
	Let $\mathfrak{s}\colon \C_+ \to \C_+$ be the unique solution to \eqref{eq:supp_silverstein_fpe_lmgf}. Then,
	\begin{equation}\label{eq:supp_lmgf_H0_fubini}
	\begin{aligned}
	\Lambda^{(0)}(\tau) = \frac{1}{2} \int_{0}^{2\tau} \mathfrak{s}(z) \diff z\,, \quad \text{for all} \ \tau < \frac{T}{2} \,.
	\end{aligned}
	\end{equation}
\end{cor}
\begin{proof}
	Observe that 
	\begin{equation}
	 -\frac{1}{2} \log\left(1 - \frac{2\tau}{\xi}\right) = \frac{1}{2}\int_{0}^{2\tau}\frac{1}{\xi - z} \diff z\,, \quad \text{for all} \  \tau < \frac{\xi}{2}\,, 
	\end{equation}
	and is $+\infty$ otherwise. We now plug this expression into \eqref{eq:supp_limiting_lmgf} and apply Fubini's theorem, which yields for all $\tau < \frac{T}{2}$
	\begin{equation}\label{eq:supp_lmgf_H0_fubini_proof}
	\begin{aligned}
	\Lambda^{(0)}(\tau) &=  \frac{1}{2} \int_{\R_+}\int_{0}^{2\tau}\frac{1}{\xi - z} \diff z \diff \nu_{\gamma, \lambda} (\xi) = \frac{1}{2} \int_{0}^{2\tau}\int_{\R_+}\frac{1}{\xi - z} \diff \nu_{\gamma, \lambda} (\xi) \diff z \,.
	\end{aligned}
	\end{equation}
	Recalling \eqref{eq:supp_Stieltjes_s_mulam} we get the claim.
\end{proof}

Since the fixed-point equation \eqref{eq:supp_silverstein_fpe_lmgf} can be solved efficiently, this expression can be leveraged to compute $\Lambda^{(0)}$ with high accuracy, yet not in closed-form in general.

\paragraph{The $\cH_1$ Case} Computing the limiting LMGF under the $\cH_1$ hypothesis \arev{is more delicate}. We have the following result:
\begin{lem}\label{lem:supp_lmgf_H1}
	Suppose that Assumptions \ref{ass:1} \arev{and \ref{ass:A3}} hold and let $b$ be a uniform upper bound for $\|\Sigma\|$ given by Assumption \ref{ass:A1}. Then, almost surely,
	\begin{equation}\label{eq:supp_limiting_lmgf_H1}
	\Lambda^{(1)}(\tau) = \Lambda^{(0)}(\tau) +  \arev{\tau \Delta(\tau) \,, \quad \text{for all} \ \tau < \tfrac{1}{2}\lambda / b\,,}
	\end{equation}
	where $\Lambda^{(0)}$ is defined as in \eqref{eq:supp_limiting_lmgf} and \arev{$\Delta(\tau)$ is defined by
    \begin{equation}
        \Delta(\tau):= \frac{\kappa_\tau(\lambda)}{\lambda} \int_{\R_+} \frac{\xi}{ \kappa_\tau(\lambda) + (1-2\tau \kappa_\tau(\lambda)/\lambda) \xi} \lSNR(\xi)\diff H(\xi)\,,
    \end{equation}
    where $\kappa_\tau(\lambda)$ is the unique solution to
    \begin{equation}\label{eq:silverstein_k_tau}
        \kappa_\tau(\lambda) = \lambda + \gamma \kappa_\tau(\lambda) \int_{\R_+} \frac{\xi}{ \kappa_\tau(\lambda) + (1-2\tau \kappa_\tau(\lambda)/\lambda) \xi} \diff H(\xi)\,.
    \end{equation}
    }
\end{lem}
\begin{proof}
    \arev{Let $\hat \xi_1, \dots, \hat \xi_p$ the eigenvalues of $\hat W^{-1}$ and $\hat \xi_{\min}$ the minimum. First, note that from \eqref{eq:supp_formula_Lambda0hat},
    \begin{equation}
        \hat \Lambda^{(0)}(\tau)   = -\frac{1}{2p} \log \det(I-2\tau\hW) \,,
    \end{equation}
    for all $\tau < \hat \xi_{\min}/2$ and $\hat \Lambda^{(0)}(\tau)=+\infty$ otherwise.} Under $\cH_1$, we have
	\begin{equation*}
	\begin{aligned}
	&\hat \Lambda^{(1)}(\tau) = \frac{1}{p} \log \E\big[ \exp\{\tau x^T \hat \Sigma_\lambda^{-1} x\} \mid \hat X, \cH_1 \big] \\
	&= \frac{1}{p} \log \E\big[ \exp\{\tau (x - s)^T \hat \Sigma_\lambda^{-1} (x - s)\} \exp\{ 2\tau (x - s)^T \hat \Sigma_\lambda^{-1} s\}\exp\{\tau s^T \hat \Sigma_\lambda^{-1} s\}\mid \hat X, \cH_1 \big]\\
	& = \frac{1}{p} \log \E\big[ \exp\{\tau (x - s)^T \hat \Sigma_\lambda^{-1} (x - s)\} \exp\{2 \tau (x - s)^T \hat \Sigma_\lambda^{-1} s\} \mid \hat X, \cH_1 \big] + \frac{\tau}{p} s^T \hat \Sigma_\lambda^{-1} s\,.
	\end{aligned}
	\end{equation*}
	
    \arev{
    We apply Gaussian integration, see, e.g., \cite[Eq. (2.84)]{bishop2006pattern}, which gives for $\tau<\hat \xi_{\min}/2$ that
    \begin{equation}\label{eq:lam_1_finite}
    \begin{aligned}
        \hat \Lambda^{(1)}(\tau)&= \hat \Lambda^{(0)}(\tau)+\frac{\tau}{p}\,s^T\hSigl^{-1}s+\frac{2\tau^2}{p}\,s^T\hSigl^{-1}\Sigma^{1/2}\big(I-2\tau\hW\big)^{-1}\Sigma^{1/2}\hSigl^{-1}s\\
    &= \hat \Lambda^{(0)}(\tau)+\frac{\tau}{p}\,s^T\big(\hSigl-2\tau\Sigma\big)^{-1}s\,,
    \end{aligned}
    \end{equation}
    and $+\infty$ otherwise.
    }

    \arev{To pass to the limit, we restrict ourselves to $\tau < \tau_0:= \tfrac{1}{2}\lambda / b$ and turn the quadratic form in the format of Theorem \ref{thm:rmb}. First, we define $\Sigma_\tau:=\big(\Sigma^{-1}-\frac{2\tau}{\lambda}I\big)^{-1}$, $\hSig_\tau:=\Sigma_\tau^{1/2}\big(\frac1n \hat Z \hat Z^T\big)\Sigma_\tau^{1/2}$, and $s_\tau:=\Sigma_\tau^{1/2}\Sigma^{-1/2}s$. Now, we have
    \begin{equation}
    \begin{aligned}
    \hSigl-2\tau\Sigma&=\Sigma^{1/2}\big(\hW^{-1}-2\tau I\big)\Sigma^{1/2}
    \\&=\Sigma^{1/2}\Big(\frac1n \hat Z\hat Z^T+\lambda\Sigma_\tau^{-1}\Big)\Sigma^{1/2}
    =\Sigma^{1/2}\Sigma_\tau^{-1/2}\big(\hSig_\tau+\lambda I\big)\Sigma_\tau^{-1/2}\Sigma^{1/2} \,.
    \end{aligned}
    \end{equation}
    Therefore, the quadratic form in the right-hand side of \eqref{eq:lam_1_finite} reads as
    \begin{equation}\label{eq:inproof_quadratic_form}
    \begin{aligned}
         \frac{\tau}{p}s^T\big(\hSigl-2\tau\Sigma\big)^{-1}s&= \frac{\tau}{p}s^T\Sigma^{-1/2}\Sigma_\tau^{1/2}\big(\hSig_\tau+\lambda I\big)^{-1}\Sigma_\tau^{1/2}\Sigma^{-1/2}s\\
         & = \frac{\tau}{p} s_\tau^T \big(\hSig_\tau+\lambda I\big)^{-1} s_\tau \,.
    \end{aligned}
    \end{equation}
    Observe that $\hat \Sigma_\tau$ is sample covariance matrix of the training set $\hat X_\tau:= \Sigma_\tau^{1/2} \hat Z$ with population covariance matrix $\Sigma_\tau$, and $s_\tau$ is a deterministic vector. For $\tau < \tau_0$, we are in the setting of Theorem \ref{thm:rmb}, which yields as $p\to +\infty$, almost surely,
    \begin{equation}\label{eq:inproof_quadratic_form_2}
        \frac{\tau}{p} s_\tau^T \big(\hSig_\tau+\lambda I\big)^{-1} s_\tau = \tau \trace  \Big[ \frac{ s_\tau s_\tau^T}{p} \big(\hSig_\tau+\lambda I\big)^{-1}  \Big] \sim  \frac{\tau \kappa_\tau(\lambda)}{\lambda }\trace  \Big[ \frac{ s_\tau s_\tau^T}{p} \big(\Sigma_\tau + \kappa_\tau(\lambda) I\big)^{-1}  \Big] \,,
    \end{equation}
    where $\kappa_\tau(\lambda)$ is the unique solution to the corresponding Silverstein equation \eqref{eq:silverstein_k_tau}.
    Passing to the limit, after omitted elementary manipulations to write the limit with respect to the original problem geometry, we get as $p\to +\infty$, that the quadratic form in \eqref{eq:inproof_quadratic_form_2} converges almost surely to $\tau \Delta(\tau)$, which yields the claim.
  }
\end{proof}

\arev{Note that, while \eqref{eq:lam_1_finite} is exact and holds for all $\tau < \hat \xi_{\min}/2$, the limit in Lemma \ref{lem:supp_lmgf_H1} is established only for $\tau < \tau_0 := \lambda/(2 b)$. Although these two bounds may coincide, e.g., for isotropic data with $\gamma>1$, extending the argument beyond $\tau_0$ would make $\Sigma_\tau$ indefinite (or even singular), so that it can no longer define a covariance matrix. Nevertheless, as we show in Appendix \ref{sec:supp_rate_quadratic_neumann_pearson}, this result is sufficient to establish the rate function in the Neyman--Pearson setup.}

\subsubsection{Computing the rate function}\label{sec:supp_computing_the_rate_function}

According to G\"artner--Ellis, Theorem \ref{thm:supp_gartner_ellis}, as soon as $0 \in \Int \Dom \Lambda^{(0)}$, the rate in the sense of the LDP \arev{is governed by the} Fenchel--Legendre conjugate function \cite{bc2017convex} of $\Lambda^{(0)}$, which is defined by
\begin{equation}\label{eq:supp_def_fenchel_dual}
I^{(0)}(\omega) := (\Lambda^{(0)})^*(\omega) := \sup_{\tau \in \R} \ \big(\omega \tau - \Lambda^{(0)}(\tau)\big)\,.
\end{equation}
In our case, $\Lambda^{(0)}$ can be written as an integral function with respect to the \emph{first branch} of the Stieltjes transform of $\nu_{\gamma, \lambda}$, cf.~\eqref{eq:def_first_branch_of_s}. Leveraging this specific structure, we derive an explicit formula for the rate function. Once again, the $\cH_1$ case can be deduced from $\cH_0$, so we focus on $\cH_0$ first. \EC{Omitted comment which is useless}

\begin{prop}\label{prop:quadratic_rate_function_H0}
	\arev{$I^{(0)}$ coincides} with
	\begin{equation}\label{eq:rate_function_H0}
	I^{(0)}(\omega) = \frac{1}{2} \int_{\mu_{0, \infty}}^{\omega} \underline{\mathfrak{s}}^{[-1]}(u) \diff u\,,  \quad \text{for all $\omega > 0$ and $I^{(0)}(\omega)=+\infty$ else}\,.
	\end{equation} 
\end{prop}
\begin{proof}
    \arev{We denote by $\underline{\Lambda}^{(0)}$ the lower semicontinuous envelope of $\Lambda^{(0)}$, which might differ from $\Lambda^{(0)}$ only at the boundary of its domain $\tau = T/2 > 0$, and in particular has the same subgradient of $\Lambda^{(0)}$.} We heavily rely on \cite[Theorem 16.29]{bc2017convex}, which yields $\partial I^{(0)} = (\partial \Lambda^{(0)})^{-1}$. In our case, we have $(\Lambda^{(0)})'(\tau) = \underline{\mathfrak{s}}(2\tau)$ for all $\tau < \tfrac{T}{2} $. Since $\underline{\mathfrak{s}}$ is increasing, the following limit exists
    \begin{equation}\label{eq:supp_def_S}
        S:=\lim_{\tau \to T^{-}/2} (\Lambda^{(0)})'(\tau) = \lim_{\tau \to T^{-}} \underline{\mathfrak{s}}(\tau) \in \R \cup \{+\infty\}\,.
    \end{equation}
    We divide the analysis into two cases:
	
	Suppose first that $S=+\infty$. By \cite[Theorem 16.29]{bc2017convex}, $(I^{(0)})'(\omega) = \frac{1}{2}\underline{\mathfrak{s}}^{-1}(\omega)$ for all $\omega \in (0, +\infty)$. Therefore, we get \eqref{eq:rate_function_H0} by integrating with starting point $\mu_{0, \infty}$, since for this choice we have $I^{(0)}(\mu_{0, \infty}) = 0$. Indeed, in this case $I^{(0)}$ has a unique stationary point, which, once again applying \cite[Theorem 16.29]{bc2017convex}, can be seen to be exactly $\mu_{0, \infty} = (\Lambda^{(0)})'(0)$. Thus, it is the unique minimizer of $I^{(0)}$, which has value
	\begin{equation*}
	\inf_{\omega\in \R} I^{(0)}(\omega) = - \sup_{\omega\in \R} -I^{(0)}(\omega) = \arev{\underline{\Lambda}^{(0)}(0) =  \Lambda^{(0)}(0)} = 0\,.
	\end{equation*}
	Suppose now that $S< +\infty$. It must also hold that $\lim_{\tau\to T/2 }\Lambda^{(0)}(\tau) < +\infty$\,. Therefore, we have
	\begin{equation}
	\partial \Lambda^{(0)}(\tau) = \begin{cases}
	\{\underline{\mathfrak{s}}(2\tau)\} & \text{if $\tau < \frac{T}{2} $}\,,\\
	\big[S, +\infty \big) & \text{if $\tau = \frac{T}{2}$}\,,\\
	\emptyset & \text{if $\tau > \frac{T}{2}$}\,.
	\end{cases}
	\end{equation}
	Inverting the subgradient of $\Lambda^{(0)}$ we obtain that of $I^{(0)}$, which turns out to be single-valued and indeed \arev{exactly equal to} $\tfrac{1}{2}\underline{\mathfrak{s}}^{[-1]}$. Integrating once again with starting point $\mu_{0, \infty}$ we obtain \eqref{eq:rate_function_H0}.
	
	To conclude, we need to show that $I^{(0)}(0) = +\infty$. In this edge case, we have, using the fixed-point equation \eqref{eq:supp_silverstein_fpe_lmgf}, 
	\begin{equation*}
	\begin{aligned}
	I^{(0)}(0) &= -\inf_{\tau \in \R} \Lambda^{(0)}(\tau) = \int_{(-\infty, 0]} \underline{\mathfrak{s}}(z) \diff z\\
	& = \int_{(-\infty, 0]}\int_{\R_+} \bigg(\lambda t^{-1} - z + \frac{1}{1 + \gamma \underline{\mathfrak{s}}(z)} \bigg)^{-1} \diff H(t) \diff z\\
	& \geq \int_{(-\infty, 0]}\int_{\R_+} \frac{1}{\lambda t^{-1} - z + 1}  \diff H(t) \diff z = \int_{\R_+} \int_{(-\infty, 0]} \frac{1}{\lambda t^{-1} - z + 1} \diff z \diff H(t) = +\infty\,,
	\end{aligned}
	\end{equation*}
	where in the last identity we have used Tonelli, which is applicable since the integrand is positive.
\end{proof}

\begin{figure}[t]
	\centering
	\includegraphics[width=\textwidth]{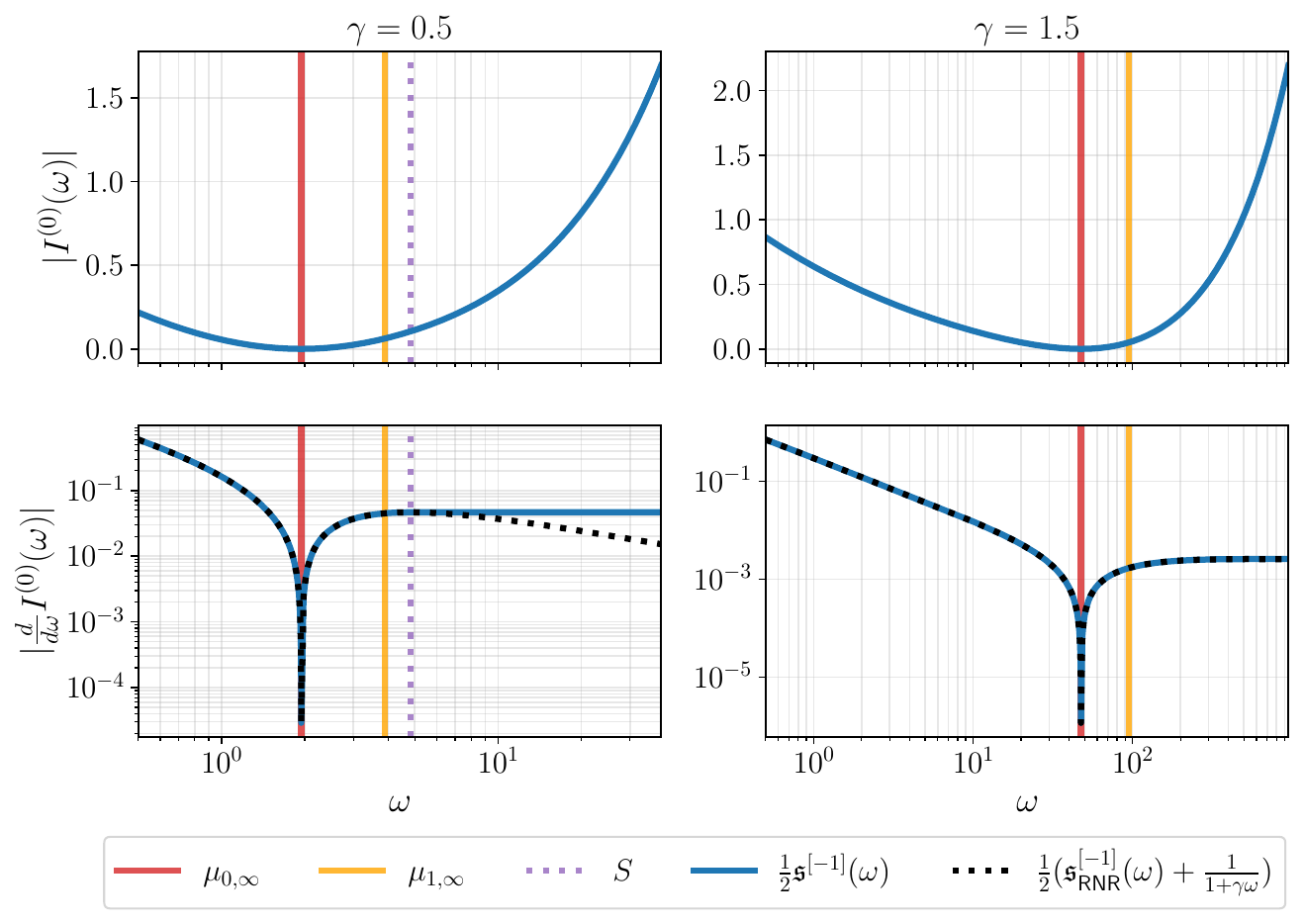}
	\caption{Computing the inverse Stieltjes transform $\underline{\mathfrak{s}}^{[-1]}$ according to \eqref{eq:pseudo_inverse_Stieltjes} and the function $I^{(0)}$ given in Proposition \ref{prop:quadratic_rate_function_H0}, with $\gamma<1$ and $\gamma > 1$. Here, $\lambda =10^{-2}$, $H = \arev{\tfrac{1}{2}}\delta_1 + \arev{\tfrac{1}{2}}\delta_2$ and $\lSNR$ is constant equal to $1$. }
	\label{fig:supp_quadratic_dual_and_Stieltjes}
\end{figure}

\begin{rmk}[Computing $I^{(0)}$]\label{rmk:supp_computing_I0}
	While computing $\mathfrak{s}(z)$ requires solving the fixed-point equation \eqref{eq:supp_silverstein_fpe_lmgf}, identifying the pseudo inverse $\underline{\mathfrak{s}}^{[-1]}$ is indeed much simpler, provided that $\dRNR$ is \emph{sharp}, i.e., $\mathfrak{s}_{\dRNR}(\tau) \to +\infty$ as $\tau \uparrow \inf \supp \dRNR$. In fact, starting from \eqref{eq:supp_silverstein_fpe_lmgf} it is easy to see that, in this case, we must have
	\begin{equation}\label{eq:supp_inverse_Stieltjes_with_rnr}
	\underline{\mathfrak{s}}^{[-1]}(u) = \underline{\mathfrak{s}}_{\dRNR}^{[-1]}(u) + \frac{1}{1 + \gamma u}\,, \quad \text{for all} \ u \in (0, S)\,,
	\end{equation}
	and is extended by a constant for all $u \geq S$. We can leverage this explicit expression to i) Compute exact rates for isotropic data, and ii) Build accurate numerical approximation of the rate function for non-isotropic data. In the latter, we only need to approximate the function $\underline{\mathfrak{s}}_{\dRNR}^{[-1]}$, which does not admit a closed form expression in general: We show an example in Figure \ref{fig:supp_quadratic_dual_and_Stieltjes}. This function as well as its integral admit closed form expressions in the isotropic regime, cf.~Proposition \ref{prop:supp_rate_quadratic_isotropic}.
\end{rmk}

\subsubsection{Exponential decay in the Neyman--Pearson setup}\label{sec:supp_rate_quadratic_neumann_pearson}

\rev{Now that the explicit expression for $I^{(0)}$ \rev{has} been derived, we can use it to bound or characterize the probability of error decay under one hypothesis, while the probability of error under the alternative hypothesis is kept fixed.} This is exactly where we utilize the G\"artner--Ellis Theorem, Theorem \ref{thm:supp_gartner_ellis}.

\begin{proof}[Proof of Theorem \ref{thm:quadratic_error_decay}] 
\arev{We divide the proof into two steps. First, we show that with the selected threshold $\lim_{p\to +\infty} P_{MD} = \beta$, then the exponential decay of $P_{FA}.$

\rev{Observe that by the CLT for triangular arrays, when suitably normalized, $\hat T$ under $\cH_1$ converges in distribution to a unit Gaussian since, conditionally on the training data, we have 
\begin{equation}
    \hat T(x)  = \sum_{i=1}^p \frac{(v_i+a_i)^2}{\hat\xi_i}, \qquad v_i \overset{\mathrm{i.i.d.}}{\sim}\mathcal N(0,1)\,,
\end{equation}
where $a=U^T\Sigma^{-1/2}s$ and $\hat W=U\operatorname{diag}(\hat\xi_1^{-1},\ldots, \hat\xi_p^{-1})U^T$. Thus, $\hat T$ is a sum of independent noncentral chi-square variables. Thus, setting $\tau_p = \hat\tau_1:=\hat\mu_1 + \hat\sigma_1 Q^{-1}(1- \beta)$, we obtain $\lim_{p\to+\infty} P_{MD}=\beta$ almost surely.}

Turning to the decay of $P_{FA}$, using the continuity of $\hat T$, we get that
\begin{equation}
  \limsup_{p\to +\infty} \frac{1}{p} \log P_{FA} = \limsup_{p\to +\infty}\frac{1}{p} \log \P \Big( \tfrac{1}{p}(\hat T(x) - \hat \tau_1) \geq 0 \  \Big| \  \hat X, \cH_0\Big)\,.
\end{equation}
Now, since $ \hat \mu_1/p \to \mu_{1, \infty}$ and $\hat \sigma_1/p \to 0$ almost surely by Proposition \ref{prop:mean_var_quadratic}, the asymptotic LMGF of $ \hat T(x) - \hat \tau_1 $ under $\cH_0$ in the sense of \eqref{eq:def_Lambda} is exactly $\tau \mapsto \Lambda^{(0)}(\tau) - \mu_{1, \infty}\tau$ for all $\tau \in \R$, whose Fenchel--Legendre conjugate is simply $\omega \mapsto I^{(0)}(\omega + \mu_{1, \infty})$. Since $0$ belongs to the interior of the domain of the limiting LMGF, cf.~Lemma \ref{lem:supp_lmgf_H0}, we get using the G\"artner--Ellis Theorem (cf.~Theorem \ref{thm:supp_gartner_ellis} applied to $\mathsf{t}_p := \tfrac{1}{p}(\hat T(x) - \hat \tau_1)$) that
\begin{equation}
     \limsup_{p\to +\infty} \frac{1}{p} \log P_{FA} \leq - \inf_{\omega \in [0, +\infty)} I^{(0)}(\omega +\mu_{1, \infty}) = - I^{(0)}(\mu_{1, \infty}) =:  - R^{(0)}_{\gamma, \lambda}\,,
\end{equation}
where in the last step we have used that $I^{(0)}$ is convex and attains its minimum on $\mu_{0, \infty}$, which is smaller than $\mu_{1, \infty}$, which makes $I^{(0)}$ non-decreasing on $[\mu_{1, \infty}, +\infty)$.
}
\end{proof}

\begin{rmk}[Decay under $\cH_1$]\label{rmk:supp_decay_under_H1}
\arev{Since Lemma \ref{lem:supp_lmgf_H1} does not establish the existence of $\Lambda^{(1)}(\tau)$ for $\tau > \tau_0$, we cannot directly apply Theorem \ref{thm:supp_gartner_ellis}, which requires the existence of $\Lambda^{(1)}(\tau)$ for all $\tau \in \R$. Although local variants of the G\"artner--Ellis theorem could be invoked, the upper bound can be established directly, even in our setting.

Similarly to the $\cH_1$ case, if we pick $\tau_p = \hat \tau_0 := \hat \mu_0 + \hat \sigma_0 Q^{-1}(\alpha)$, we get $\lim_{p\to +\infty} P_{FA} = \alpha$. Now, for every $\tau<0$ (not to be confused with the threshold), Markov inequality yields
\begin{equation}
    P_{MD}
    = \P(\hat T(x) \leq \hat \tau_0 \mid \hat X, \cH_1)
    \leq \E\big[ \exp\big\{\tau \hat T(x) \big\} \ \big| \ \hat X, \cH_1 \big]  e^{-\tau \hat \tau_0}\,.
\end{equation}
Since, almost surely, $\hat\Lambda^{(1)}(\tau)\to\Lambda^{(1)}(\tau)$ for all $\tau < 0 \leq \tau_0$ by Lemma \ref{lem:supp_lmgf_H1}, continuity, and $\hat\tau_0/p\to\mu_{0,\infty}$ almost surely, we obtain
\begin{equation}
\begin{aligned}
     \limsup_{p\to\infty} \frac1p\log P_{MD}
    &\leq \inf_{\tau\leq0}
    \left\{\Lambda^{(1)}(\tau)-\tau\mu_{0,\infty}\right\} \\
    &=  \inf_{\tau\leq0}
    \left\{\Lambda^{(0)}(\tau) + \tau \Delta(\tau)-\tau\mu_{0,\infty}\right\}
    =: - R^{(1)}_{\gamma, \lambda}\,,
\end{aligned}
\end{equation}
Thus, $R^{(1)}_{\gamma, \lambda}$ is the \emph{infimal convolution} between $I^{(0)}$ and the sum of the Fenchel--Legendre conjugate of $g\colon \tau \mapsto \tau \Delta(\tau)$ and the indicator function of the positive line at $\mu_{0, \infty}$, see, e.g., \cite[Proposition 15.2]{bc2017convex}.
}
\end{rmk}

In general, $R^{(0)}_{\gamma, \lambda}$ and $R^{(1)}_{\gamma, \lambda}$ only represent upper bounds for the decay rates of the corresponding error probabilities. Next, we show that in some cases the LDP indeed holds.

\paragraph{Isotropic regime}
Suppose we have isotropic data, i.e., $\Sigma = \sigma^2I$. Then, $\dRNR = \delta_{\RNR}$, with $\RNR:= \lambda / \sigma^2$. In this case, $\mathfrak{s}_{\dRNR}(z)=\frac{1}{\RNR - z}$, hence $\dRNR$ is sharp. \arev{Furthermore, the no-spectral-outlier assumption, Assumption \ref{ass:A3}, is satisfied.} We can thus utilize \eqref{eq:supp_inverse_Stieltjes_with_rnr} to turn Proposition \ref{prop:quadratic_rate_function_H0} into an explicit expression for the rate function, at least under $\cH_0$:\EC{Only stylistic changes here}

\begin{prop}\label{prop:supp_rate_quadratic_isotropic}
	Let $\Sigma= \sigma^2I$, $r=\RNR = \lambda/\sigma^2>0$. Let $\hat T$ be defined as in \eqref{eq:def_T_quadratic}. \arev{Define
    \begin{equation}
    F_\gamma(\omega)=\frac r2(\omega-\mu_{0,\infty})
+\frac1{2\gamma}\log\frac{1+\gamma\omega}{1+\gamma\mu_{0,\infty}}
-\frac12\log\frac{\omega}{\mu_{0,\infty}}\,, \quad \text{for} \ \omega>0\,.        
    \end{equation}
    Then, the following hold:}
	\begin{itemize}
		\item If $\gamma \in (0, 1)$, \arev{then $T=r+(1-\sqrt\gamma)^2$, $S=(\sqrt\gamma-\gamma)^{-1}$ and 
		\begin{equation}\label{eq:supp_rate_function_isotropic_smallgamma}
        I^{(0)}(\omega)=
        \begin{cases}
        F_\gamma(\omega)\,, & \text{for} \ 0<\omega\le S\,,\\
        F_\gamma(S) + \tfrac T2(\omega-S)\,,& \text{for} \ \omega>S\,.
        \end{cases}
        \end{equation}
		\item If $\gamma > 1$, then $T=r$, $S=+\infty$, and
        \begin{equation}\label{eq:supp_rate_function_isotropic_largegamma}
        I^{(0)}(\omega)=F_\gamma(\omega)\,, \quad \text{for all} \ \omega>0\,.
        \end{equation}
		and the LDP holds.}
	\end{itemize}
\end{prop}
\begin{proof}
	First, observe that $\underline{\mathfrak{s}}_{\dRNR}^{[-1]}(u):= r - \frac{1}{u}$ and from \eqref{eq:supp_inverse_Stieltjes_with_rnr} that $S = +\infty$ if and only if $\gamma > 1$. To see this, note that  by a simple calculation
	\begin{equation}\label{eq:supp_inproof_sminus1}
	u \mapsto \underline{\mathfrak{s}}_{\dRNR}^{[-1]}(u) + \frac{1}{1 + \gamma u} = r -\frac{1}{u} + \frac{1}{1 + \gamma u}\,,
	\end{equation}
	has a stationary point (a maximum indeed) in $\R_+$ if and only if $\gamma \in (0, 1)$. Therefore, if $\gamma > 1$, we must have that $\underline{\mathfrak{s}}^{[-1]}$ can be inverted on $(0, +\infty)$, yielding $S=+\infty$ \arev{and, by inspection, $T=r$}. If $\gamma \in (0, 1)$, this maximum $\frac{1}{\sqrt{\gamma} - \gamma}$, should correspond exactly to $S$, otherwise we can either extend $\underline{\mathfrak{s}}$ beyond $T$, or have infinite derivative at $z = \frac{1}{\sqrt{\gamma} - \gamma}$, both of which are not possible, so $S = \frac{1}{\sqrt{\gamma} - \gamma} < +\infty$, \arev{and the value there is $T=r + (1-\sqrt{\gamma})^2$}. Moreover, 
	\begin{equation}
	0=(I^{(0)})'(\mu_{0, \infty}) = \frac{1}{2}\underline{\mathfrak{s}}^{[-1]}(\mu_{0, \infty})\,.
	\end{equation}
	Thus, $\arev{\underline{\mathfrak{s}}^{[-1]}(\mu_{0, \infty})  =0}$, which implies $\mu_{0, \infty} \leq S$. Altogether, we conclude that $\underline{\mathfrak{s}}^{[-1]}$ coincides with \eqref{eq:supp_inproof_sminus1} in $(0, S)$, passes through zero at $\mu_{0, \infty} \leq S$ and is defined as its unique constant continuation for $u \geq S$.
	
	Once the exact formula of the integrand is identified in both the cases $\gamma \in (0, 1)$ and $\gamma > 1$, we can now derive \eqref{eq:supp_rate_function_isotropic_smallgamma} and \eqref{eq:supp_rate_function_isotropic_largegamma} by integrating it. As $S=+\infty$ in the case $\gamma > 1$, relying on Theorem \ref{thm:supp_gartner_ellis}, the LDP must hold since
    \begin{equation}
        \lim_{\tau \to T/2} (\Lambda^{(0)})'(\tau) = \lim_{\tau \to T/2}\underline{\mathfrak{s}}(2\tau) = S = +\infty\,,
    \end{equation}
    and thus the steepness condition of Theorem \ref{thm:supp_gartner_ellis} is satisfied.
\end{proof}

\begin{rmk}
	Combining the explicit expression of the rates given by Proposition \ref{prop:supp_rate_quadratic_isotropic} with the explicit expression of the moments given by Proposition \ref{prop:mean_var_quadratic} and \eqref{eq:supp_observables_isotropic}, we obtain an explicit expression of the rates in the Neyman--Pearson setup as detailed in Appendix \ref{sec:supp_rate_quadratic_neumann_pearson}.
\end{rmk}

\begin{figure}[t]
	\centering
	\begin{subfigure}[t]{0.47\textwidth}
		\centering
		\includegraphics[width=\textwidth]{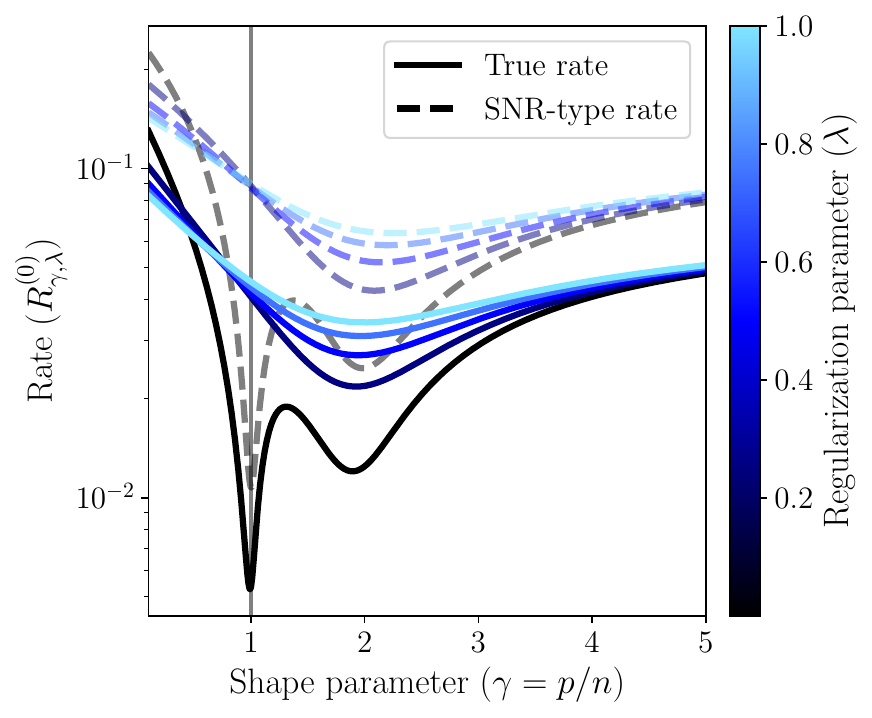}
	\end{subfigure}
	\begin{subfigure}[t]{0.47\textwidth}
		\centering
		\includegraphics[width=\textwidth]{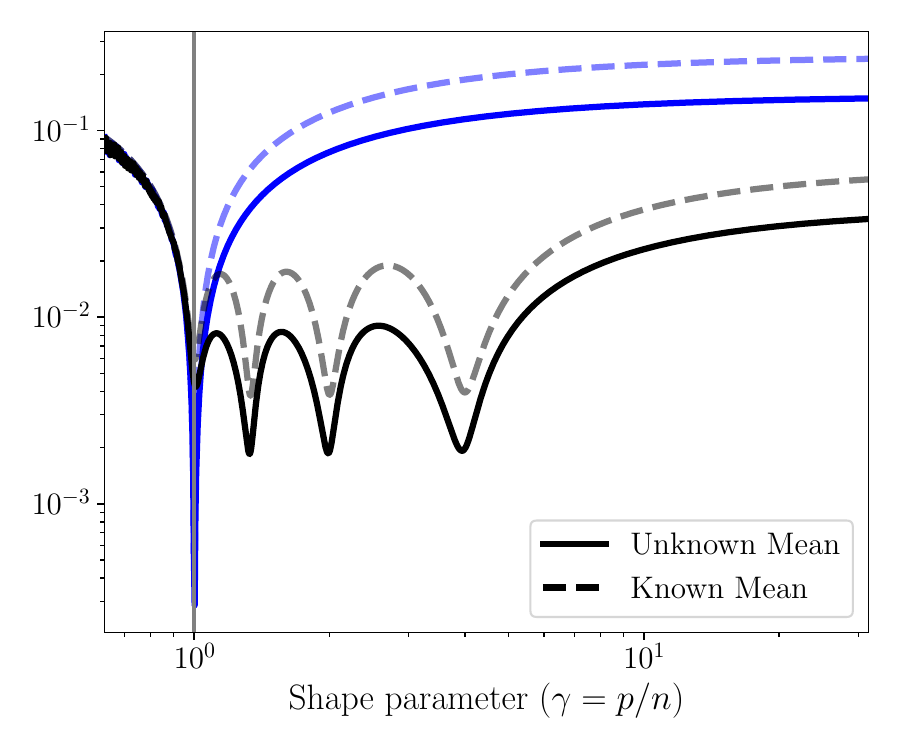}
	\end{subfigure}
	\caption{ Multiple-descent in the quadratic case. \textbf{Right:} False alarm rate function of the quadratic decision statistic compared to the same for the linear decision statistic \eqref{eq:def_That_linear_unknown} in the geometric setting of Figure \ref{fig:rate_linear_examples}. \textbf{Left:} comparing, for different choices of $\lambda$, the true rate function with the \emph{SNR-type rate} that one obtains by assuming $\hat T$ to be Gaussian.}
	\label{fig:supp_quadratic_multidescent}
\end{figure}

We are now ready to compute the rate function in this quadratic case and compare it with its linear counterpart discussed in Section \ref{sec:linear}. To avoid repetitions, we consider the same geometric setting of Section \ref{sec:linear} and show the results in Figure \ref{fig:supp_quadratic_multidescent}. Specifically, we also compare the rate function $R^{(0)}_{\gamma, \lambda}$ with a naive approximation that one gets assuming $\hat T$ to be Gaussian, yielding $$\text{SNR-type rate} \approx (\mu_{1, \infty} - \mu_{0, \infty})^2/ (2 \sigma_{0, \infty}^2)\,.$$ We refer to this approximation as the \emph{SNR-type rate}. From Figure \ref{fig:supp_quadratic_multidescent}, we can observe an expected loss in performances compared to the case where the mean is known or approximable from the training set, with an overall shapes that exhibit the same behaviors. Moreover, we observe that approximating the rate of $\hat T$ with its SNR-type rate would incur into a severe mistake.

\section{Further insights on Section \ref{sec:neyman_pearson_latent}}\label{sec:supp_neyman_pearson_latent}

\subsection{Models and data}\label{sec:supp_models_and_data}

We now describe the foundation models as well as the datasets considered for this work. The experiments use an NVIDIA A100 (80GB) GPU and 216GB of RAM.

\subsubsection{Optical images} 

We first consider optical Earth observation imagery, where detection under controlled false-alarm constraints is critical for operational monitoring and surveillance. Satellite imagery plays a central role in maritime domain awareness, as well as critical undersea infrastructure monitoring and strategic intelligence. It provides a representative high-dimensional regime in which pre-trained visual foundation models generate semantically structured latent representations. We evaluate whether Neyman--Pearson detection applied directly to these latent features achieves reliable performance across distinct remote sensing tasks that vary in semantic content, spatial structure, image resolution, and sensor characteristics.

\paragraph{Feature extraction model}
For the task, we employ the DINOv3 model, which is a vision foundation model based on a self-supervised Vision Transformer architecture, capable of generating high-quality dense features \cite{dinov3}. Specifically, we use the publicly available DINOv3 ViT-L architecture with 300M parameters that is pre-trained on a dataset of 493 million Maxar satellite images of 512$\times$512 resolution (SAT-493M). The CLS token, which is the final 1024-dimensional representation generated by the model, is utilized as the global image feature to apply the decision statistic. We evaluate the decision statistics of foundation model representations across two Earth observation tasks: ship detection and fire risk assessment.

\paragraph{Airbus ship detection challenge dataset}

First, we use the publicly available Airbus Ship Detection Challenge dataset released as part of a public competition on Kaggle aimed at automated ship detection in high resolution satellite imagery \cite{airbus_dataset}. It provides around 192,000 optical satellite images at 768$\times$768 resolution. Each image is paired with a ground-truth ship segmentation mask. Among these, approximately 42,000 contain at least one ship with a variety of sizes, orientations, and environmental conditions. The remaining $\approx$150,000 images contain only background scenes of ocean, shoreline, or land without vessels. From the dataset, we consider images containing at least one ship as positive samples ($\cH_1$) and background-only images as negative samples ($\cH_0$). Figure~\ref{fig:supp_airbus_dataset} shows example images for each hypothesis. We construct a balanced dataset with 37,376 training images ($\approx$81\%) and 8,000 validation images ($\approx$19\%) per class. The dataset occupies approximately 13 GB of storage, and feature extraction takes about 4 minutes with a batch size of 32.

\paragraph{A remote sensing dataset for fire risk assessment}

The FireRisk dataset includes seven fire risk classes with a total of 91,872
labeled images of 320$\times$320 resolution derived from high-resolution aerial imagery provided by the U.S. Department of Agriculture National Agriculture Imagery Program for fire risk assessment. The labels are determined by the U.S. Department of Agriculture based on several factors such as the wildfire likelihood, potential intensity and landscape conditions \cite{fire_risk_dataset}. From the dataset, we consider images with \textit{very low risk} label as positive samples ($\cH_1$), and images with \textit{non-burnable} label as negative samples ($\cH_0$). Figure~\ref{fig:supp_fire_risk_dataset} presents a number of scenes for each hypothesis. We sample 23,050 images per class to have a balanced dataset, where 17,959 ($\approx$78\%) of the images are used for training, and 5,091 ($\approx$22\%) are for testing.  The dataset occupies approximately 7 GB of storage, and feature extraction takes about 2 minutes and 20 seconds with a batch size of 32.

\subsubsection{Text}
We next evaluate our framework on textual data, which serves as a critical source of actionable information for operational monitoring and anomaly detection. Textual reports, alerts, and communications can provide early indications of incidents or anomalies that may not be directly observable via physical sensors. Effectively extracting high-level semantic features from such text is therefore critical for timely automated decision-making. Semantic abstraction in language is inherently high-dimensional, and foundation models induce richly structured embedding spaces to capture these nuances. Unlike vision representations learned through self-supervised invariance objectives, text embeddings are explicitly optimized for semantic retrieval via contrastive query--document supervision. This training paradigm directly shapes the geometry of the latent space to reflect semantic similarity and ranking structure rather than task-agnostic feature invariance. We evaluate whether Neyman--Pearson detection applied directly to these latent features achieves reliable performance across textual inputs that vary in semantic content, linguistic style, length, and contextual origin.

\paragraph{Feature extraction model}
For the text modality, we employ the Llama-Nemotron-Embed-1B-v2 model, a text foundation model optimized for multilingual semantic retrieval \cite{llama_nemotron}. Its backbone is based on the Llama 3.2 1B decoder-only architecture \cite{llama3herd}. To adapt the model for embedding extraction, the standard causal attention mask is replaced with a bidirectional attention mechanism to repurpose the transformer blocks as a non-causal encoder \cite{nv_embed}. The model was developed by NVIDIA using a two-stage contrastive training process: an initial pre-training phase on a proprietary mixture of 12 million document--passage pairs curated from public datasets, followed by supervised fine-tuning on 1 million high-fidelity, instruction-wrapped query--document pairs. To obtain a global semantic representation, we perform masked mean pooling across the final hidden states by excluding padding tokens from the average, which yields a 2048-dimensional feature vector that serves as the global text feature for applying the decision statistic.

\paragraph{Disaster incidents data from Twitter}
We use the publicly available HumAID: Human-Annotated Disaster Incidents Data from Twitter dataset, comprising annotated tweets collected from multiple real-world disaster events \cite{humaid_dataset}. Each tweet is labeled with one of 10 humanitarian categories based on its context. We consider tweets reporting injured or dead people, missing or found people, infrastructure and utility damage, requests or urgent needs, and rescue, volunteering or donation efforts as positive samples ($\cH_1$), which represents urgent or actionable information. The remaining tweets with labels caution and advice, displaced people and evacuations, other relevant information, sympathy and support, and non-humanitarian content are considered negative samples ($\cH_0$). Table~\ref{tab:supp_humaid_dataset} presents a number of example tweets and their corresponding humanitarian categories for each hypothesis. To construct a balanced binary dataset, we first combine the validation and test splits to form a larger test set. Then, we sample 36,764 tweets per class to create a balanced dataset, where 25,732 ($\approx$70\%) are used for training, and 11,032 ($\approx$30\%) are used for testing. The dataset occupies approximately 120 MB of storage, and feature extraction takes about 8 minutes and 40 seconds with a batch size of 32.

\subsubsection{Audio}
Finally, we investigate acoustic event detection in audio signals, which is fundamental for surveillance and early-warning systems. Reliable detection under false-alarm constraints is particularly important in safety-critical acoustic sensing applications. Audio representations differ from vision and text as they encode temporal dependencies and spectral structure inherent to raw waveforms. Pre-trained audio foundation models produce compact global embeddings that summarize long temporal contexts. In contrast to retrieval-optimized text embeddings and invariance-driven visual features, foundation audio representations are shaped by self-supervised predictive objectives that model masked acoustic units. This training paradigm emphasizes temporal consistency and contextual prediction rather than semantic ranking. The audio modality therefore provides an additional test of whether Neyman–Pearson detection in latent space remains effective when the underlying geometry is induced by sequence prediction objectives rather than semantic ranking or invariance-driven training.

\paragraph{Feature extraction model}
For this task, BEATs is used as the backbone architecture to extract features. As a self-supervised audio foundation model built on a Transformer encoder, it learns high-level acoustic representations directly from raw waveforms \cite{BEATs}. We use the publicly available BEATs\_iter3+ model having 90M parameters and pre-trained on 2 million 10-second audio clips sampled at 16 kHz from YouTube videos (AudioSet-2M) \cite{audioset}. The frame-level embeddings produced by the final Transformer layer are mean-pooled over the temporal dimension to obtain a 768-dimensional global representation of the audio signal, which is then used to compute the decision statistic.

\begin{figure}[t]
	\centering
	\begin{subfigure}{.45\textwidth}
		\includegraphics[width=\linewidth]{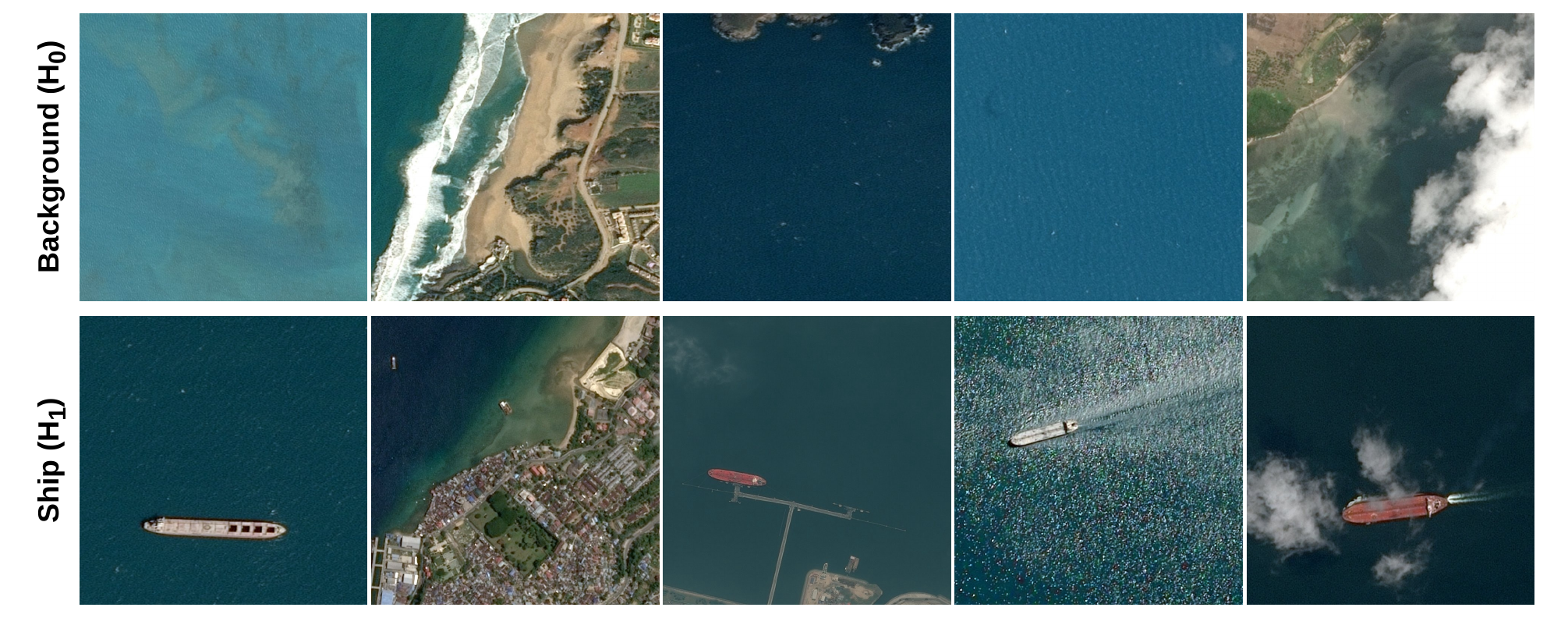}
		\caption{Airbus Ship Detection dataset.}
		\label{fig:supp_airbus_dataset}
	\end{subfigure}
	\begin{subfigure}{.52\textwidth}
		\includegraphics[width=\linewidth]{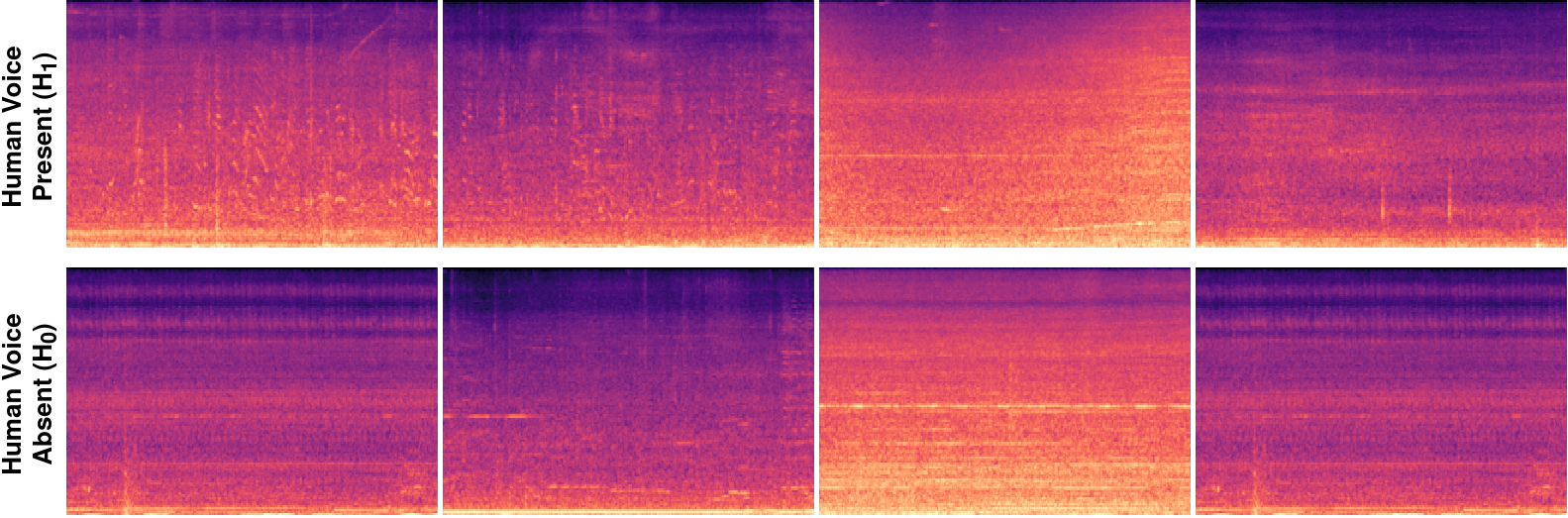}
		\caption{SONYC-UST-V2 dataset.}
		\label{fig:supp_sonyc_dataset}
	\end{subfigure}\\
	\begin{subfigure}{.45\textwidth}
		\includegraphics[width=\linewidth]{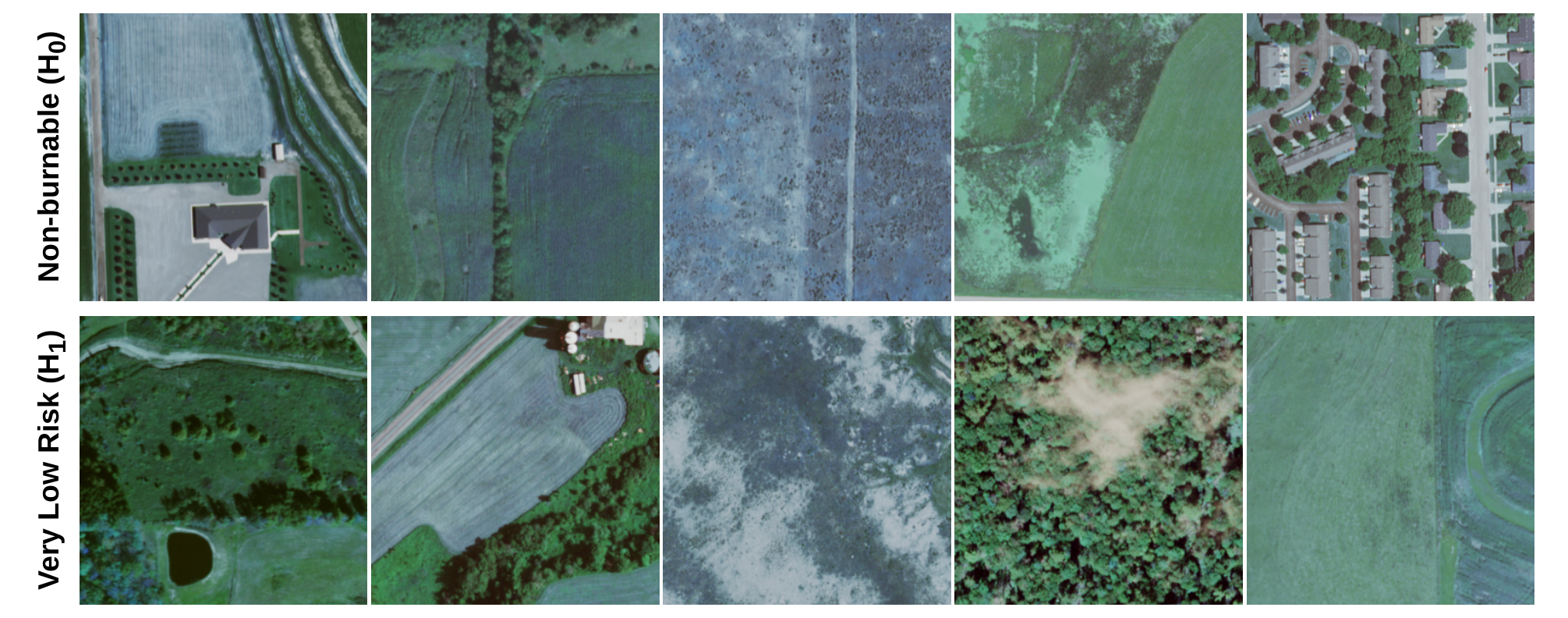}
		\caption{FireRisk dataset. }
		\label{fig:supp_fire_risk_dataset}
	\end{subfigure}
	\caption{Samples extracted from three different image datasets, including satellite images \ref{fig:supp_airbus_dataset} and \ref{fig:supp_fire_risk_dataset}, and a spectrogram \ref{fig:supp_sonyc_dataset}.}
\end{figure}

\paragraph{SONYC urban sound tagging dataset}
We use the SONYC Urban Sound Tagging v2 dataset (SONYC-UST-V2), a multilabel urban acoustic dataset collected from the SONYC sensor network deployed across New York City \cite{SONYC-UST-V2}. The dataset provides 10-second recordings annotated across 23 fine-grained sound classes grouped into 8 coarse categories, including engine noise, machinery impact, powered saws, alert signals, music, human voice, and dog sounds. We consider the human voice coarse class as the target and formulate a one-versus-rest binary detection task. Recordings in which human voice is present form positive samples ($\mathcal{H}_1$), while recordings in which human voice is entirely absent form negative samples ($\mathcal{H}_0$). Since the dataset is multilabel, negative samples may contain other sound classes, but contain no human voice activity. Annotations from all available annotators and all dataset splits are pooled prior to class assignment. We prioritize the verified ground-truth labels provided by the SONYC team when available; otherwise, we consider the class present if at least one crowd annotator labeled it as such. Figure~\ref{fig:supp_sonyc_dataset} illustrates representative mel spectrograms for each hypothesis. We sample an equal number of audio recordings per class to obtain a balanced dataset, where 11,520 (80\%) are used for training and 2,880 (20\%) are used for testing. The dataset occupies approximately 17 GB of storage, and feature extraction takes about 14 minutes with a batch size of 32.

\begin{table}[t]
	\centering
	\small
	\begin{tabularx}{\textwidth}{l X X}
		\toprule
		\textbf{Hypothesis} & \textbf{Tweet} & \textbf{Humanitarian Category} \\
		
		\midrule
		$\cH_0$ & Flash Flooding Threatens Towns Devastated by Hurricane Maria. \#Weather \#PuertoRico:  via @YouTube & Caution and advice \\
		
		\addlinespace 
		
		\midrule
		$\cH_0$ & Louisiana – Hurricane categories tell only part of the story — here’s the damage storms like Irma and Harvey can do & Other relevant information \\
		\midrule
		\midrule
		
		$\cH_1$ & Community Blood Center needs donations to prepare for Hurricane Dorian  via @ndn & Requests or urgent needs \\
		
		\addlinespace 
				
				\midrule
				$\cH_1$ & Drone video: St. Augustine condos destroyed by Hurricane Irma -  $|$ Droneblog - \#drones & Infrastructure and utility damage \\
		
		\bottomrule
	\end{tabularx}
	\caption{Examples from the HumAID dataset.}
	\label{tab:supp_humaid_dataset}
\end{table}

\subsection{Experimental design}\label{sec:supp_experimental_design} 

In this section, we give further details on the implementation aspects of Section \ref{sec:experimental_design}.

\begin{figure}[t]
	\centering
	\begin{subfigure}[t]{0.4\textwidth}
		\centering
		\includegraphics[width=\textwidth]{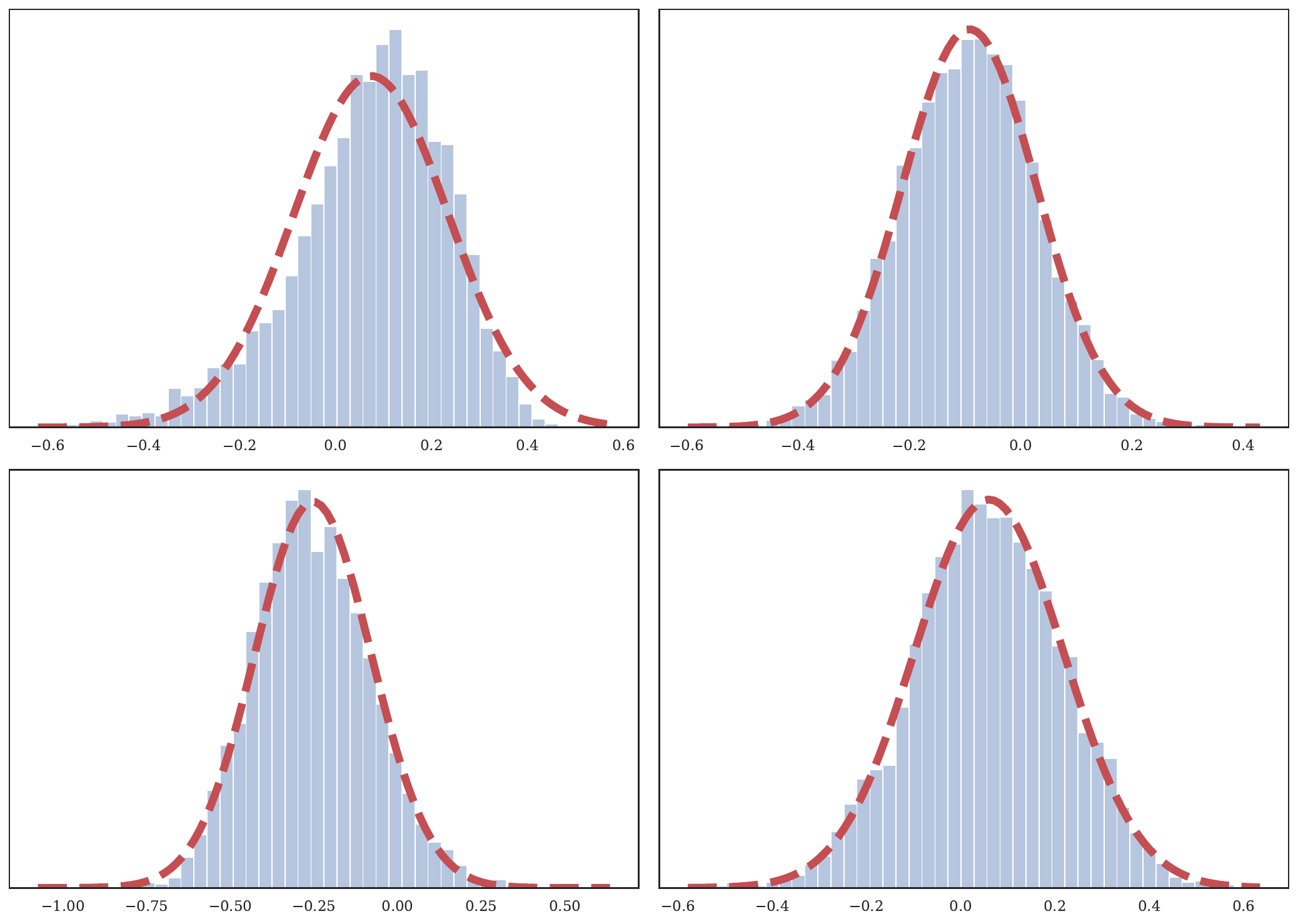}
		\caption{BEATs : SONYC UST V2.}
		\label{fig:supp_marginals_beats_dataset}
	\end{subfigure}
	\begin{subfigure}[t]{0.4\textwidth}
		\centering
		\includegraphics[width=\textwidth]{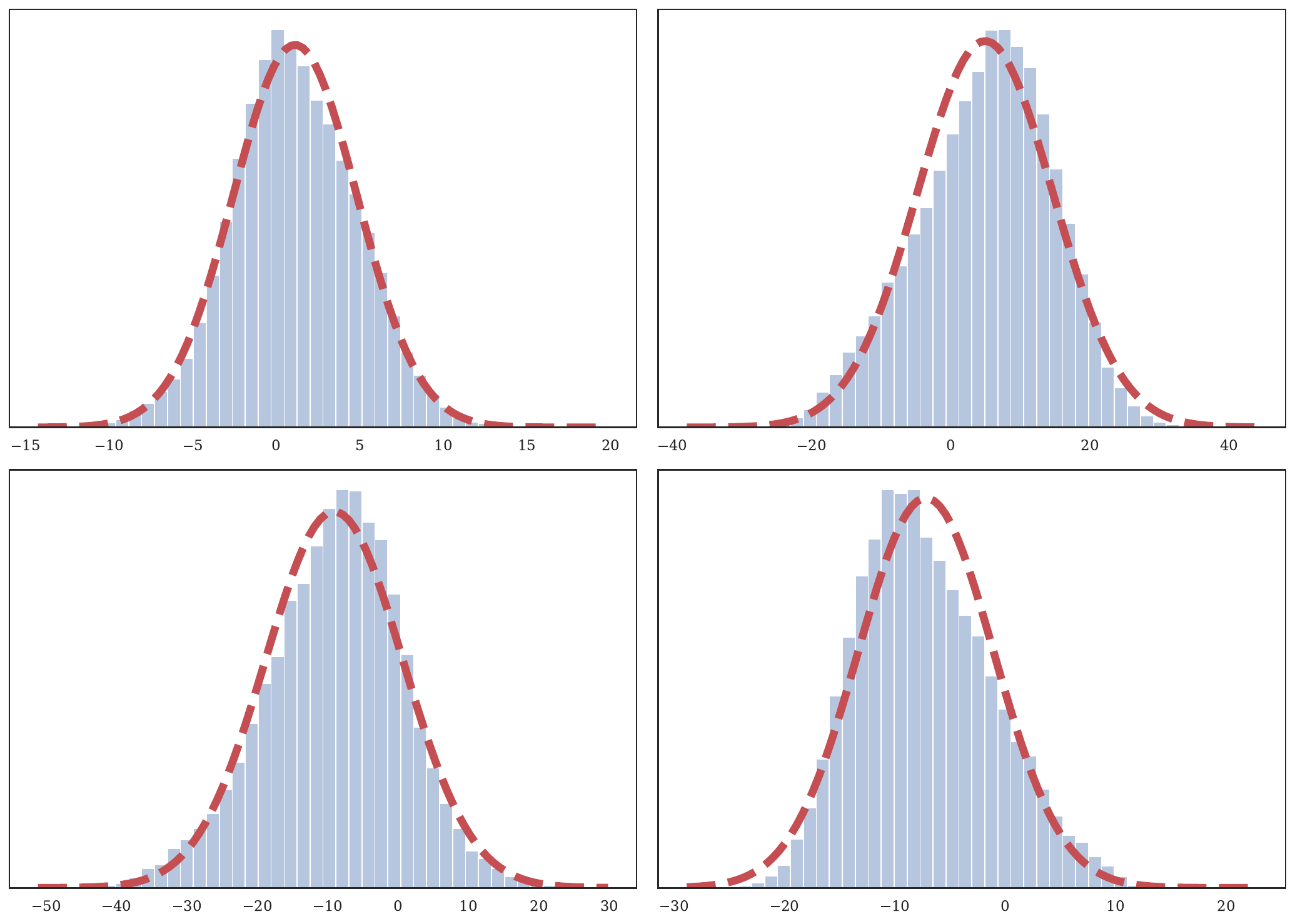}
		\caption{DINOv3 : Airbus.}
		\label{fig:supp_marginals_airbus_dataset}
	\end{subfigure}
	\begin{subfigure}[t]{0.4\textwidth}
		\centering
		\includegraphics[width=\textwidth]{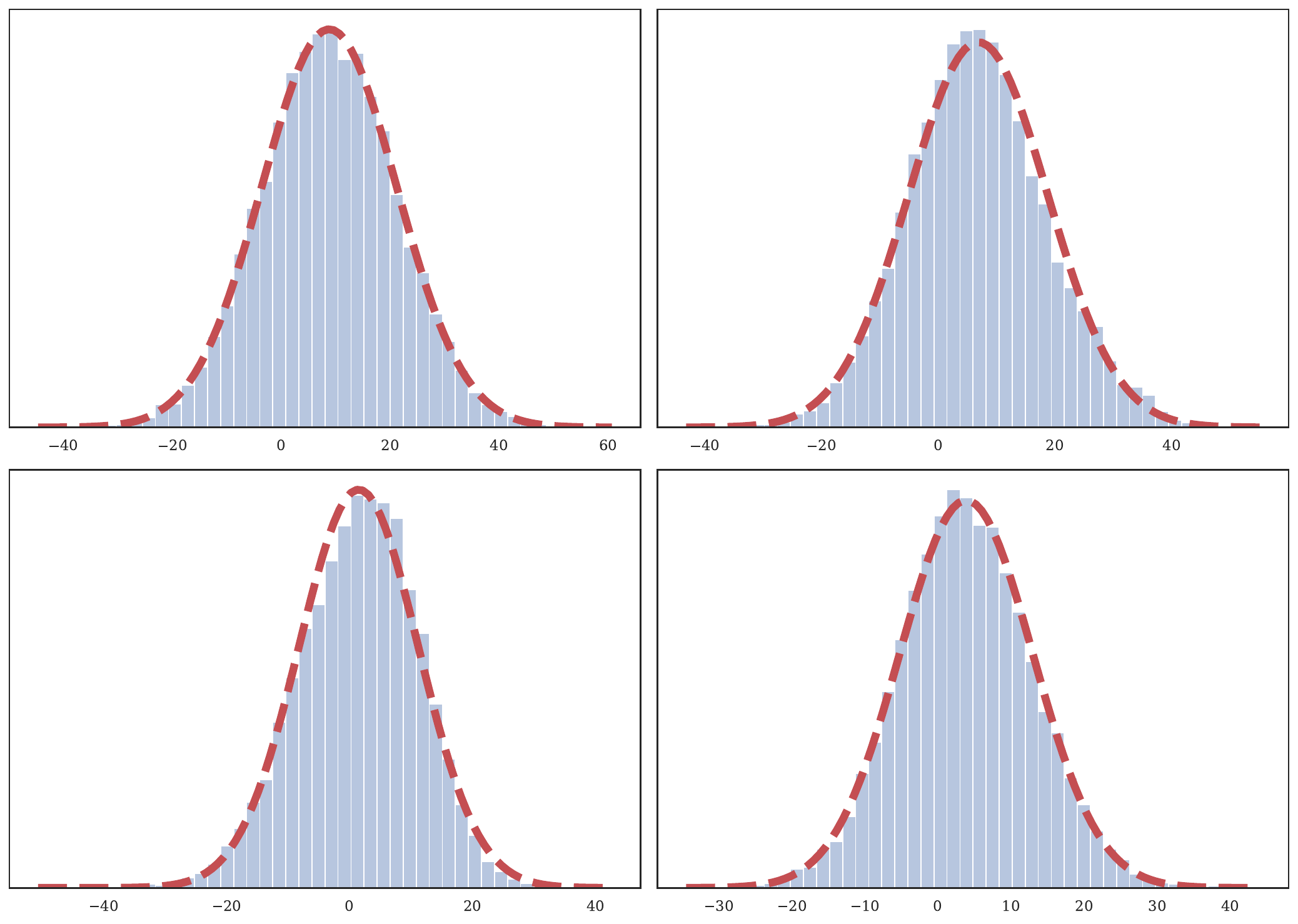}
		\caption{DINOv3 : FireRisk.}
		\label{fig:supp_marginals_fire_risk_dataset}
	\end{subfigure}
	\begin{subfigure}[t]{0.4\textwidth}
		\centering
		\includegraphics[width=\textwidth]{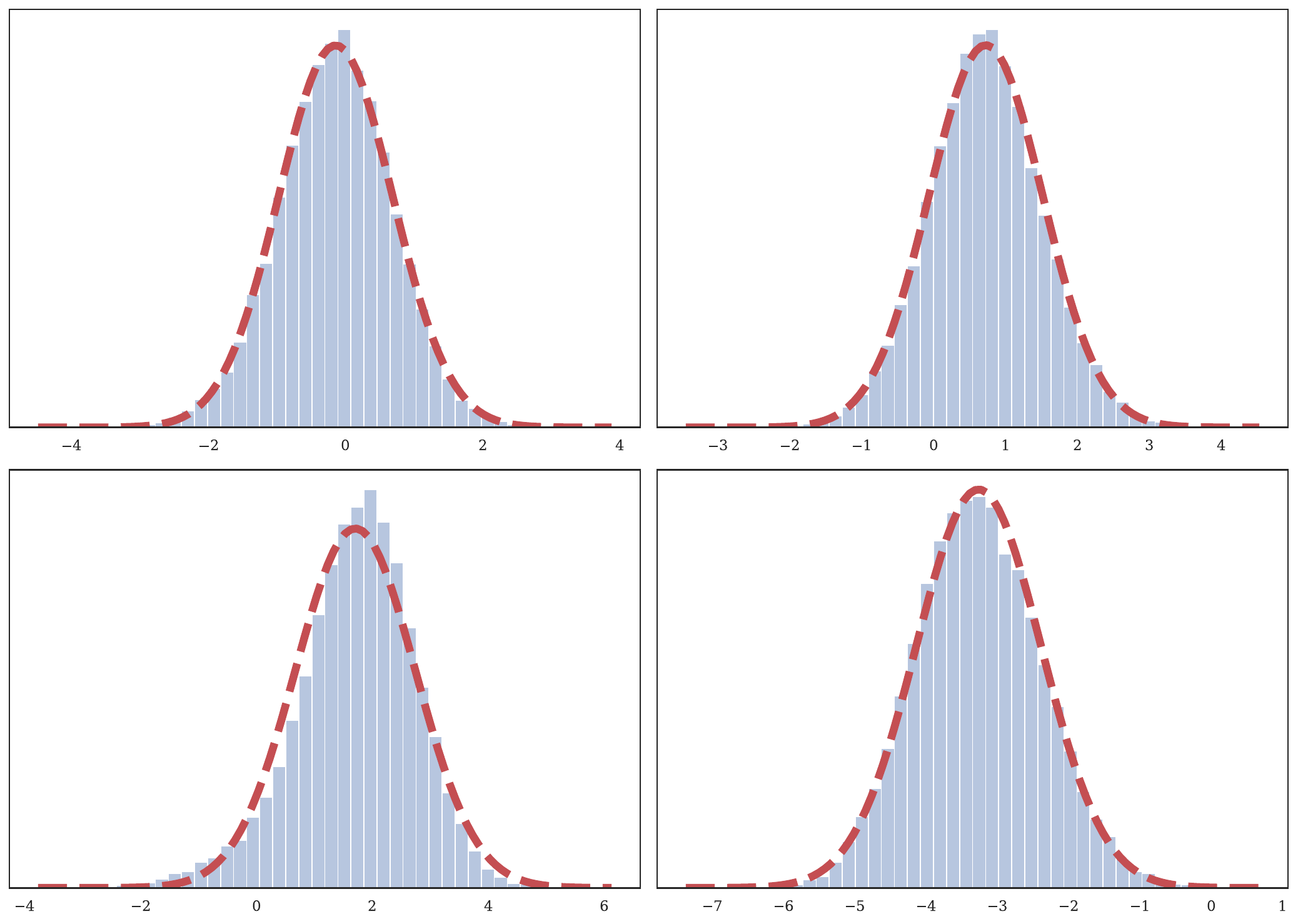}
		\caption{Llama NE1bv2 : HumAID.}
		\label{fig:supp_marginals_llama_dataset}
	\end{subfigure}
	\caption{Distribution of different training data in the latent space of different foundation models. Showing four randomly selected marginals and their Gaussian fit.}
	\label{fig:supp_nebula_marginals}
\end{figure}

As a preliminary check, we inspect randomly selected marginal distributions of each training data mapped into the latent space of the considered foundation models and fit them with Gaussian distributions. The results are displayed in Figure \ref{fig:supp_nebula_marginals}.

\rev{Second, Figure~\ref{fig:supp_multidescent_realdata} extends the theoretical rate curves of Figure~\ref{fig:proposed_model} to larger values of $\gamma$, for which empirical validation is not possible due to limited data availability. Interestingly, the curves exhibit non-monotonic behavior, reminiscent of the multiple-descent phenomenon observed in Figure~\ref{fig:simulated_data}.
\begin{figure}
    \centering
    \includegraphics[width=0.4\linewidth]{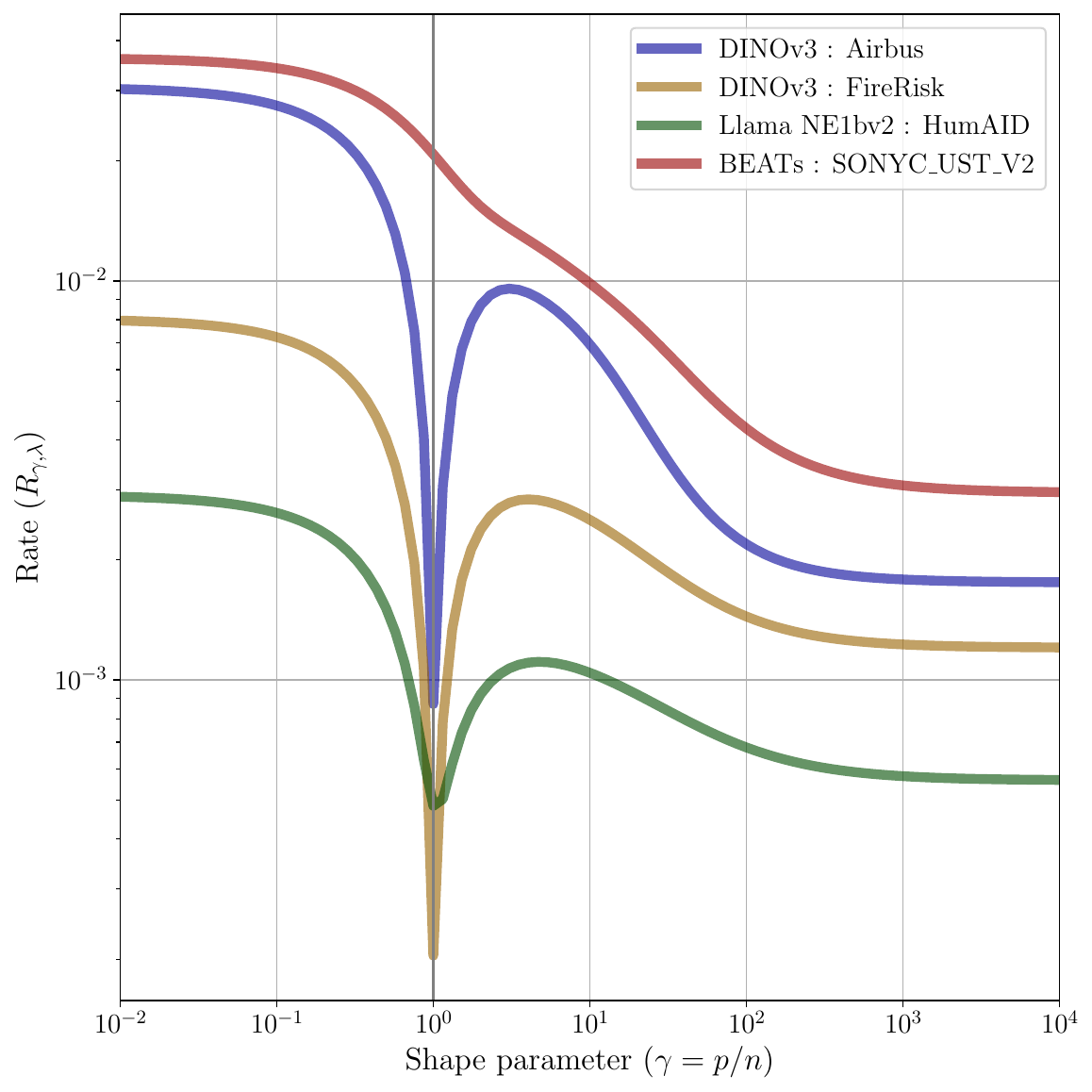}
    \caption{\rev{Theoretical rate as a function of the shape parameter $\gamma$ across different foundation models and datasets, extending Figure~\ref{fig:proposed_model} to larger values of $\gamma$ on a logarithmic scale.}}
    \label{fig:supp_multidescent_realdata}
\end{figure}
}

\subsubsection{Mismatched covariances}\label{sec:supp_covmismatch_1}
\begin{figure}[t]
	\centering
	\begin{subfigure}[t]{0.39\textwidth}
		\centering
		\includegraphics[width=\textwidth]{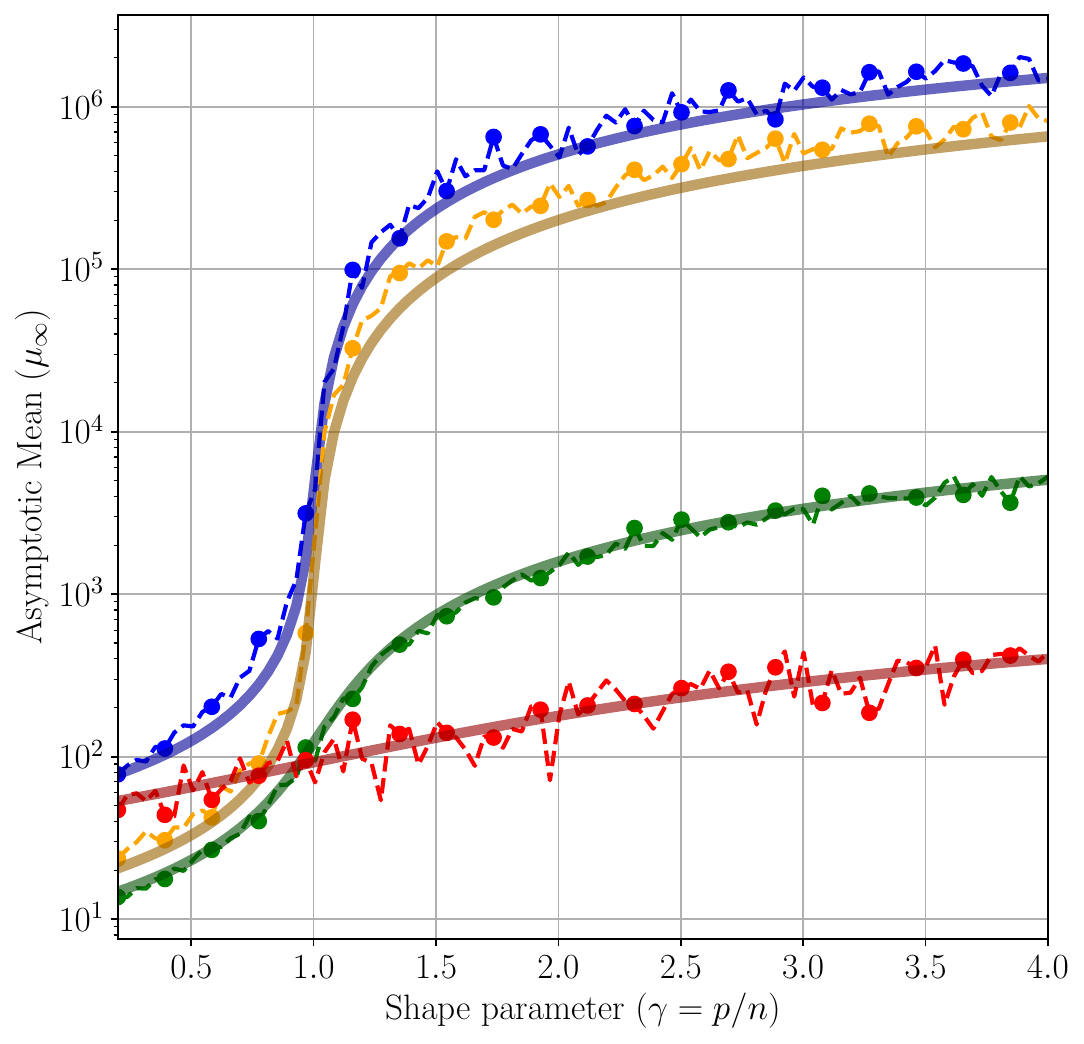}
	\end{subfigure}
	\begin{subfigure}[t]{0.39\textwidth}
		\centering
		\includegraphics[width=\textwidth]{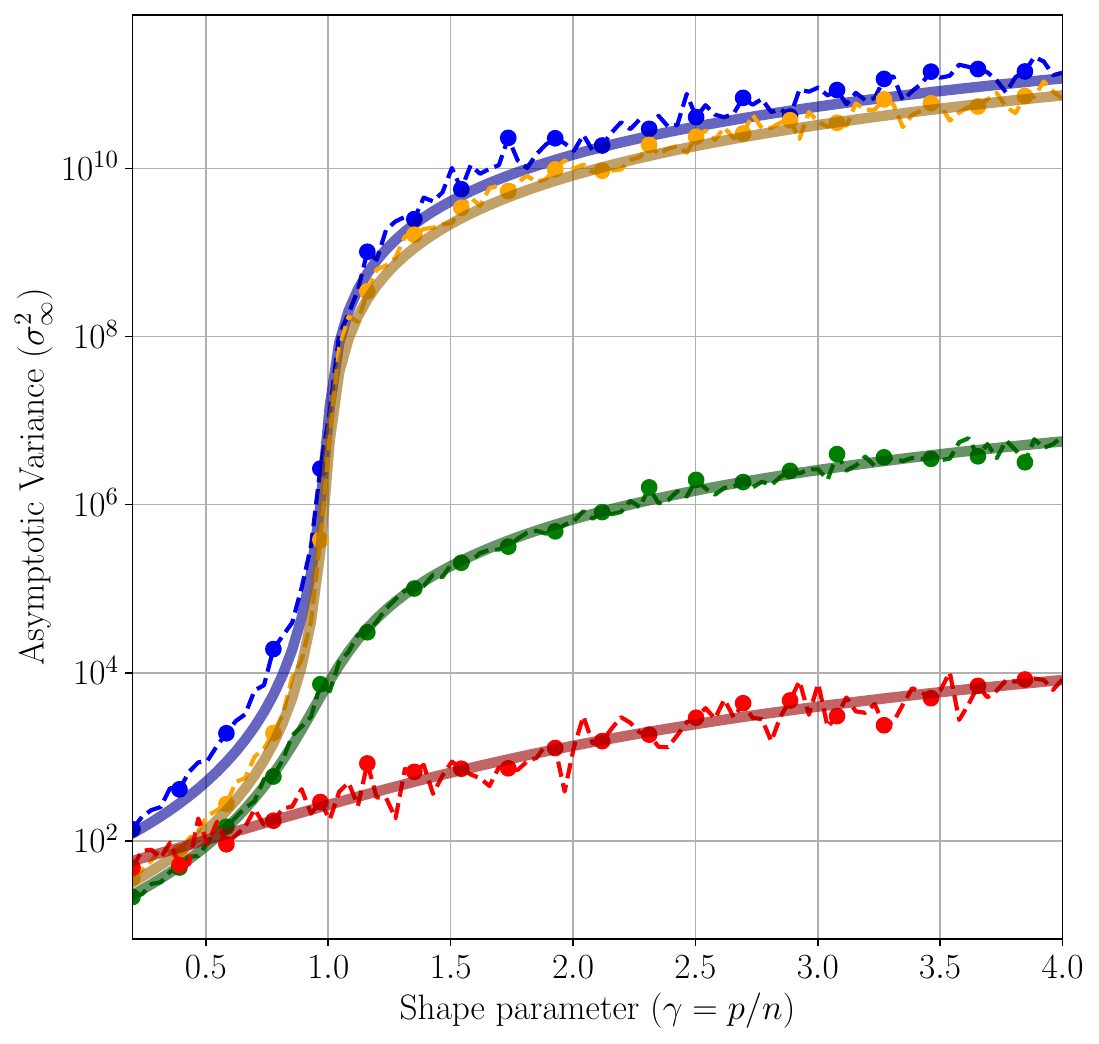}
	\end{subfigure}
	\caption{Mean and variance of the decision statistic $\hat T$ across the different models and datasets. Comparison between theoretical and empirical curves. Refer to Figure \ref{fig:proposed_model} for the legend.}
	\label{fig:supp_nebula_aggregate_moments}
\end{figure}

In Point \ref{item:nebula_compute_mean_rate} we adapt our theoretical formula to the case in which in the test set $\hat x \sim \cN(s, \Sigma_1)$, with $\Sigma_1\neq \Sigma_0:= \Sigma$, i.e., where the covariance matrix under $\cH_1$ does not coincide with the covariance matrix under $\cH_0$. In this case, we denote by $\hat \Sigma_{\lambda, 0}:= (\hat \Sigma_0 + \lambda I)$, where $\hat \Sigma_0$ is the centered sample covariance matrix from $\cH_0$ data and shall not be confused with $\hat \Sigma_\lambda$ from \eqref{eq:def_That_linear_unknown}, with $\lambda = 0$. First, the mean now reads as
\begin{equation*}
\begin{aligned}
\frac{\hat \mu}{p} & = \trace \bigg[\frac{s \hat s^T}{p} \big( \hat \Sigma_0 + \lambda I \big)^{-1}\bigg] \sim \frac{\kappa(\lambda)}{\lambda  }\trace\bigg[ \frac{s s^T}{p} \big(\Sigma_0 + \kappa(\lambda)I\big)^{-1}\bigg] \sim \frac{\kappa(\lambda)}{\lambda  } \n_1(\kappa(\lambda))\,,
\end{aligned}
\end{equation*}
where $\n_1$ is computed entirely from $\cH_0$. Next, note that the variances of $\hat T$ under $\cH_0$ and $\cH_1$ are no longer equal. We have: 
\begin{equation*}
\begin{aligned}
\frac{\hat \sigma^2_0}{p} & =\frac{1}{p} \E[\hat T(x)^2 \mid \hat X, \cH_0] = \frac{1}{p} \E[(\hat s^T \hat \Sigma_{\lambda, 0}^{-1} x)^2 \mid \hat X, \cH_0]\\
& = \frac{1}{p}  \hat s^T \hat \Sigma_{\lambda, 0}^{-1} \E\big[xx^T \mid \hat X, \cH_0\big] \hat \Sigma_{\lambda, 0}^{-1} \hat s = \frac{1}{p} \hat s^T \hat \Sigma_{\lambda, 0}^{-1} \Sigma_0 \hat \Sigma_{\lambda, 0}^{-1}\hat s = \frac{1}{p} \trace \big[ \hat s \hat s^T \hat \Sigma_{\lambda, 0}^{-1} \Sigma_0 \hat \Sigma_{\lambda, 0}^{-1}\big]\,.\\
& \sim \frac{\kappa(\lambda)^2}{\lambda^2} \frac{\n_2(\kappa(\lambda))}{\big(1 - \gamma \m_2(\kappa(\lambda))\big) }\,,
\end{aligned}
\end{equation*}
entirely computed on $\cH_0$, so once again the limit of $\hat \sigma^2_0$ is not affected by the mismatch. For $\hat \sigma^2_1$ things change, we have, again from Theorem \ref{thm:rmb} and neglecting infinitesimal terms,
\begin{equation*}
\begin{aligned}
\frac{\hat \sigma^2_1}{p} & = \frac{1}{p}\E[(\hat T(x) - \hat \mu)^2 \mid \hat X, \cH_1] = \frac{1}{p} \E[(\hat s^T \hat \Sigma_{\lambda, 0}^{-1} (x - s))^2 \mid \hat X, \cH_1]\\
& = \frac{1}{p} \hat s^T \hat \Sigma_{\lambda, 0}^{-1} \E\big[(x- s)(x - s)^T \mid \hat X, \cH_1\big] \hat \Sigma_{\lambda, 0}^{-1} \hat s = \frac{1}{p}  \hat s^T \hat \Sigma_{\lambda, 0}^{-1} \Sigma_1 \hat \Sigma_{\lambda, 0}^{-1}\hat s = \trace \bigg[ \frac{\hat s \hat s^T}{p} \hat \Sigma_{\lambda, 0}^{-1} \Sigma_1 \hat \Sigma_{\lambda, 0}^{-1}\bigg]\,.\\
& \sim \frac{\kappa(\lambda)^2}{\lambda^2} \trace\bigg[ \frac{s s^T}{p} \big(\Sigma_0 + \kappa(\lambda) I\big)^{-1}\Sigma_1 \big(\Sigma_0 + \kappa(\lambda) I\big)^{-1}\bigg] \\
& \qquad + \frac{\kappa(\lambda)^2}{\lambda^2} \trace\bigg[ \frac{s s^T}{p} \big(\Sigma_0 + \kappa(\lambda) I\big)^{-2}\Sigma_0\bigg]\trace\big[\Sigma_1 \big(\Sigma_0 + \kappa(\lambda) I\big)^{-2}\Sigma_0\big]\frac{1}{n - \diff f_2(\kappa(\lambda))}\,.
\end{aligned}
\end{equation*}
While we can easily derive a formula in the spirit of \eqref{eq:mean_var_linear}, we do not display it for brevity. We compare these theoretical formulas with their corresponding empirical ones in the four considered modalities in Figure \ref{fig:supp_nebula_aggregate_moments}. Please refer to Figure \ref{fig:proposed_model} for the corresponding legend.

\rev{
\subsubsection{Estimating $H$ and $\lSNR$ from truncated datasets}
In our experimental pipeline, we estimate $H$ and $\lSNR$ from the full training set. Although large, this dataset may still yield estimates affected by high-dimensional effects: The estimation aspect ratio is close to zero, but remains positive. A natural question is therefore whether replacing $H$ and $\lSNR$ by estimates obtained at a positive aspect ratio yields a good approximation of $R_{\gamma,\lambda}$. Although we can establish a quantitative convergence result, we limit ourselves here to a numerical check using simulated Gaussian data.

We consider a non-isotropic population spectral distribution $H=0.2\, \delta_1+0.5\, \delta_2+0.3 \, \delta_4$ and unit $\lSNR$. For each $\eta= p/n \in[0.01,0.5]$, we estimate $H$ and $\lSNR$ with a truncated dataset of independent Gaussian observations of size $\hat n := p / \eta$. We then used these estimates to compute the approximate rate curve over $\gamma \in [0,2]$. For small values of $\eta$, the estimated curves remain uniformly close to the rate curve obtained using the true population geometry, although they exhibit a systematic vertical shift. For values of $\eta$ larger than $0.5$, the mismatch is more significant, so we neglected these values. We show the results in Figure \ref{fig:supp_stability_gamma}.

The vertical shift---analogous to the mismatch observed in Figure~\ref{fig:proposed_model} for ``Llama NE1bv2: HumAID''---is not surprising. As $\eta$ increases, the empirical spectral estimate becomes increasingly biased toward zero, as the sample covariance matrix is indeed shifting closer to singularity. This effect is also visible in the third panel of Figure~\ref{fig:supp_stability_gamma}. This distortion of the estimated spectral geometry can, in turn, lead to an overestimation of the rate function.

\begin{figure}
    \centering
    \includegraphics[width=\linewidth]{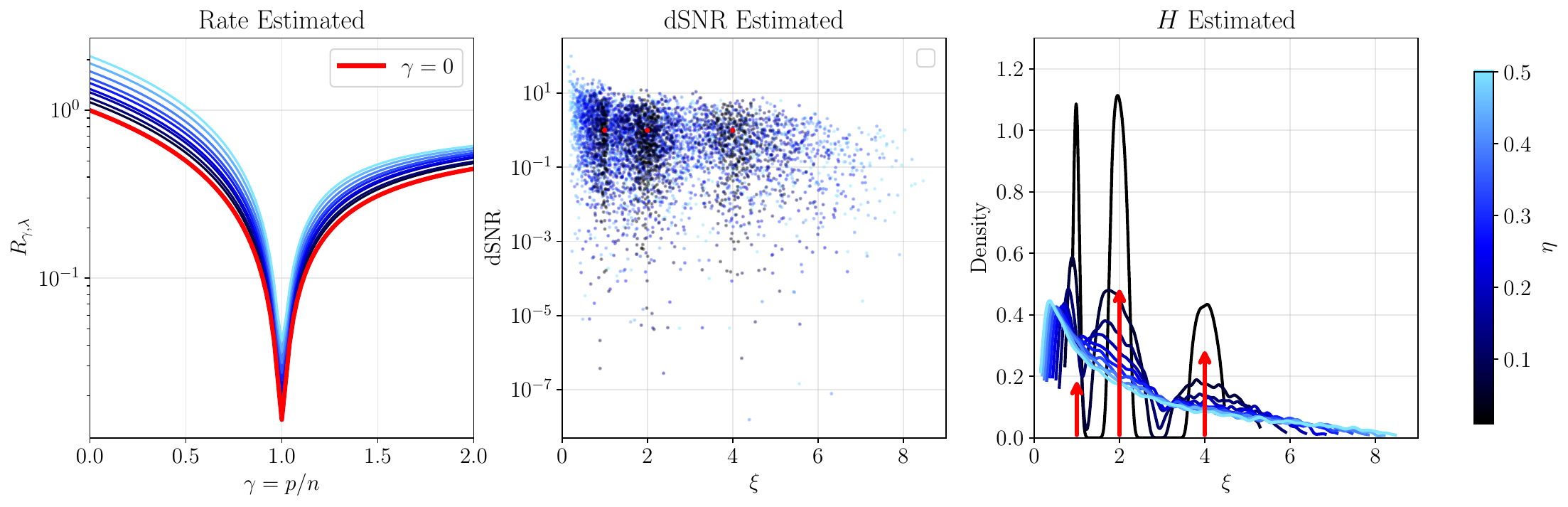}
    \caption{\rev{Stability of the estimated rate and population geometry under dataset truncation.}}
    \label{fig:supp_stability_gamma}
\end{figure}

}

\subsubsection{Computing the theoretical $P_{MD}$ with fixed $P_{FA}$} 
\begin{figure}[t]
	\centering
	\includegraphics[width=.8\linewidth]{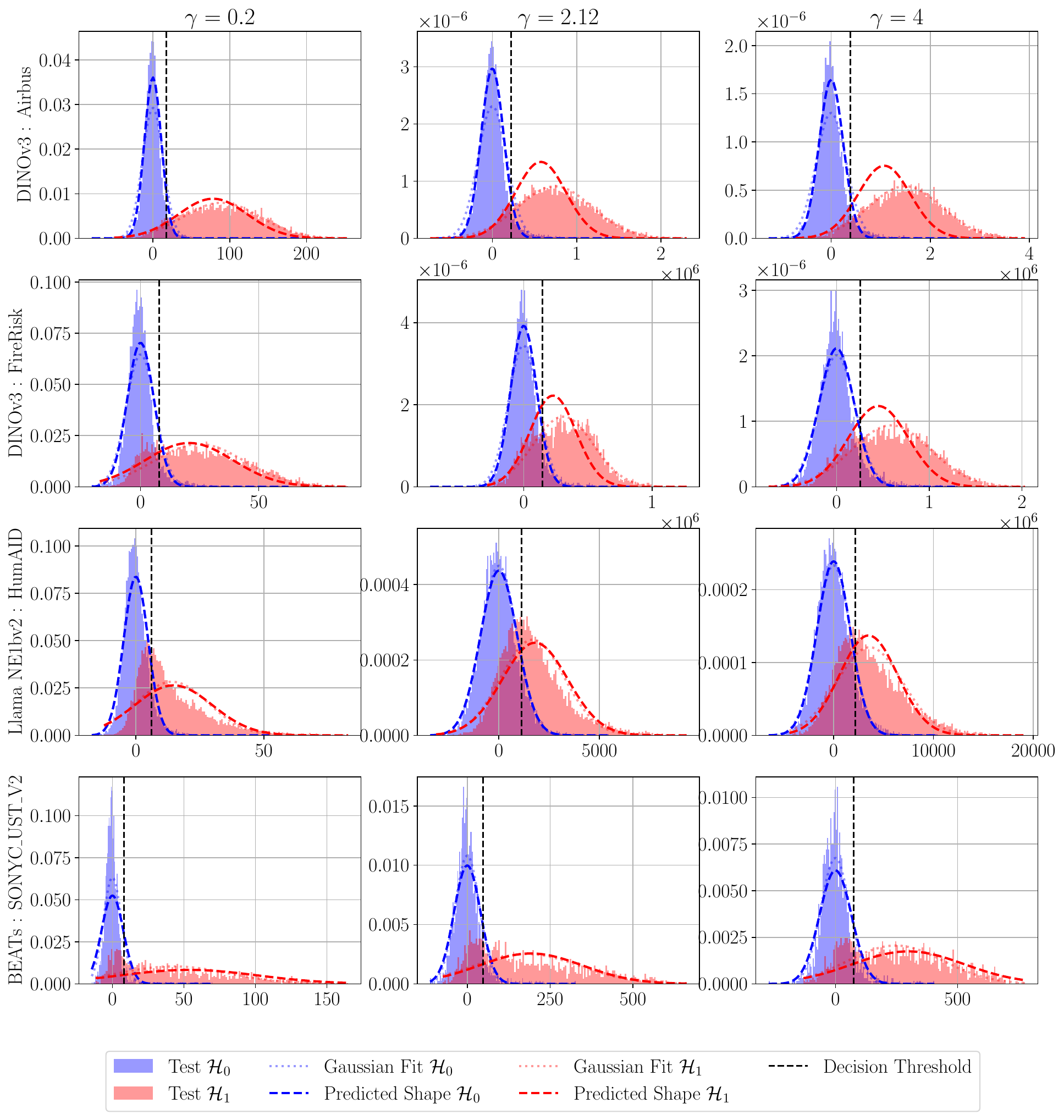}
	\caption{Sample distribution of $\hat T$ under $\cH_0$ and $\cH_1$.}
	\label{fig:supp_nebula_distributions}
\end{figure}

Once mean and variances for $\hat T$ are computed both under $\cH_0$ and under $\cH_1$, we can easily extend \eqref{eq:supp_pmd_and_pfa} to the mismatched case considered in this numerical example. The formula for $P_{MD}$ as a function of $P_{FA}$ simply becomes:
\begin{equation}
P_{MD} = 1 - Q\left( \frac{ \hat \sigma_0 Q^{-1}(P_{FA}) - \hat \mu}{ \hat \sigma_1} \right)\,.
\end{equation}
Plotting the theoretical prediction for $\hat \sigma_0$, $\hat \sigma_1$ and $\hat \mu$ in the high-dimensional regime, we obtain the results discussed in Figure \ref{fig:nebula_probability_of_detection}. Further, in Figure \ref{fig:supp_nebula_distributions} we show the agreement between the predicted distributions of $\hat T$ under $\cH_0$ and $\cH_1$ test data for different values of $\gamma$. In particular, we observe that while for $\cH_0$ data the predicted distribution is very close to the empirical one, for $\cH_1$ data the prediction is less accurate, yet still yielding a good match.

\subsubsection{LLR in different statistical settings}\label{sec:supp_covmismatch_2} In the case featuring different covariance matrices under $\cH_0$ and $\cH_1$, the data-driven log-likelihood ratio shall be replaced by
\begin{equation}\label{eq:supp_llr_quadratic}
\begin{aligned}
\text{DS}(H, K) \, : \quad	&\hat T (x) = -\frac{1}{2}(x - \hat s)^T H(x - \hat s)
+ \frac{1}{2}x^T K x\,,
\end{aligned}
\end{equation}
with $H, K$ being $\hat \Sigma_{\lambda, 0}^{-1}$ or $\hat \Sigma_{\lambda, 1}^{-1}$, yielding three possible decision statistics. Note that while using $H=K$ yield a linear decision statistic, using $H\neq K$ yields a quadratic form. For the linear decision statistics the analysis can be performed using the same arguments introduced Section \ref{sec:linear}. In contrast, for the quadratic setup, we can employ the approach presented in Section \ref{sec:quadratic}---leveraging RMT and the G\"artner--Ellis Theorem.

\begin{figure}[t]
	\centering
	\begin{subfigure}[t]{.65\linewidth}
		\includegraphics[width=\linewidth]{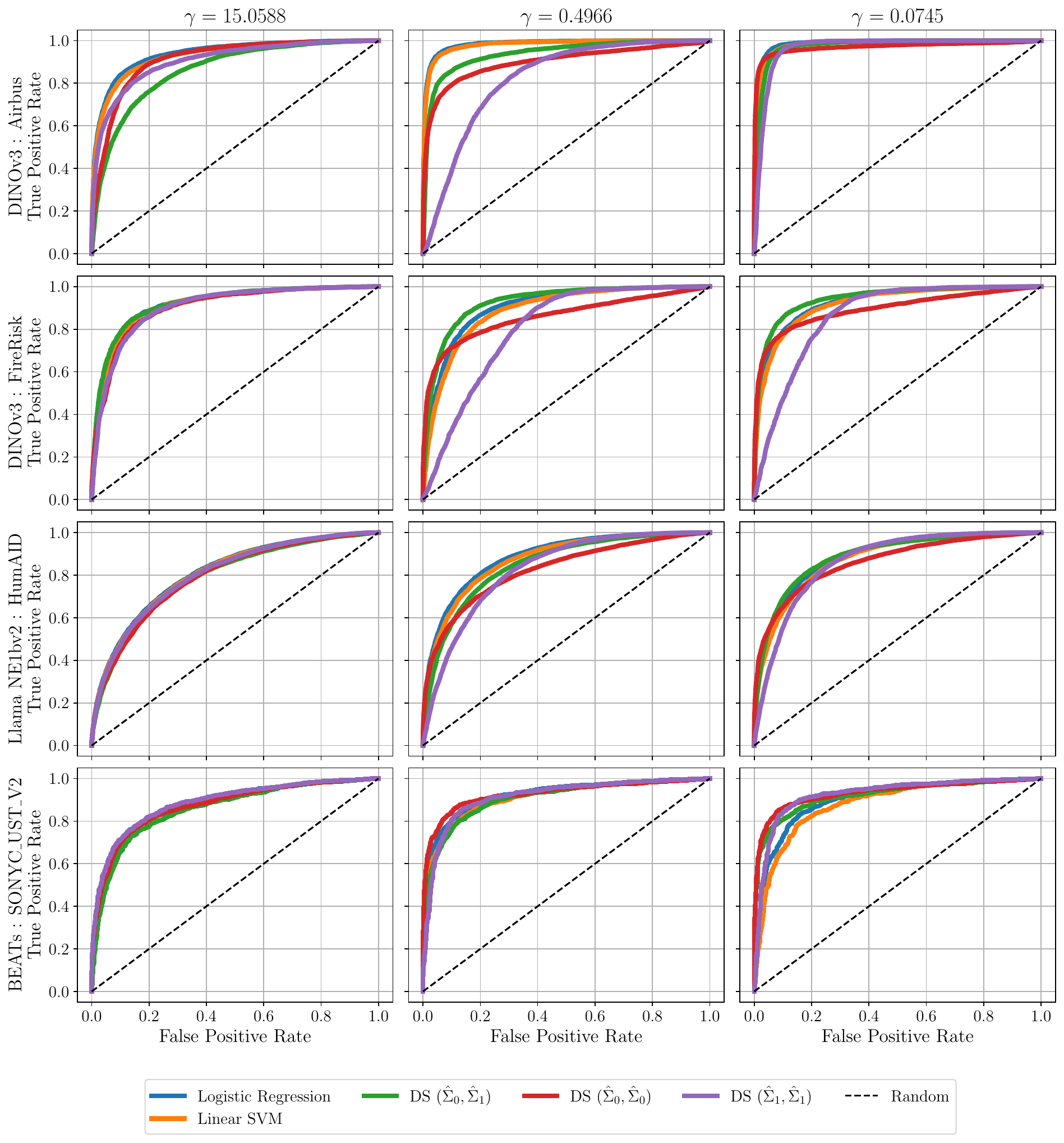}
		\caption{Linear scale.}
		\label{fig:nebula_roc_linear}
	\end{subfigure}
	\begin{subfigure}[t]{.65\linewidth}
		\includegraphics[width=\linewidth]{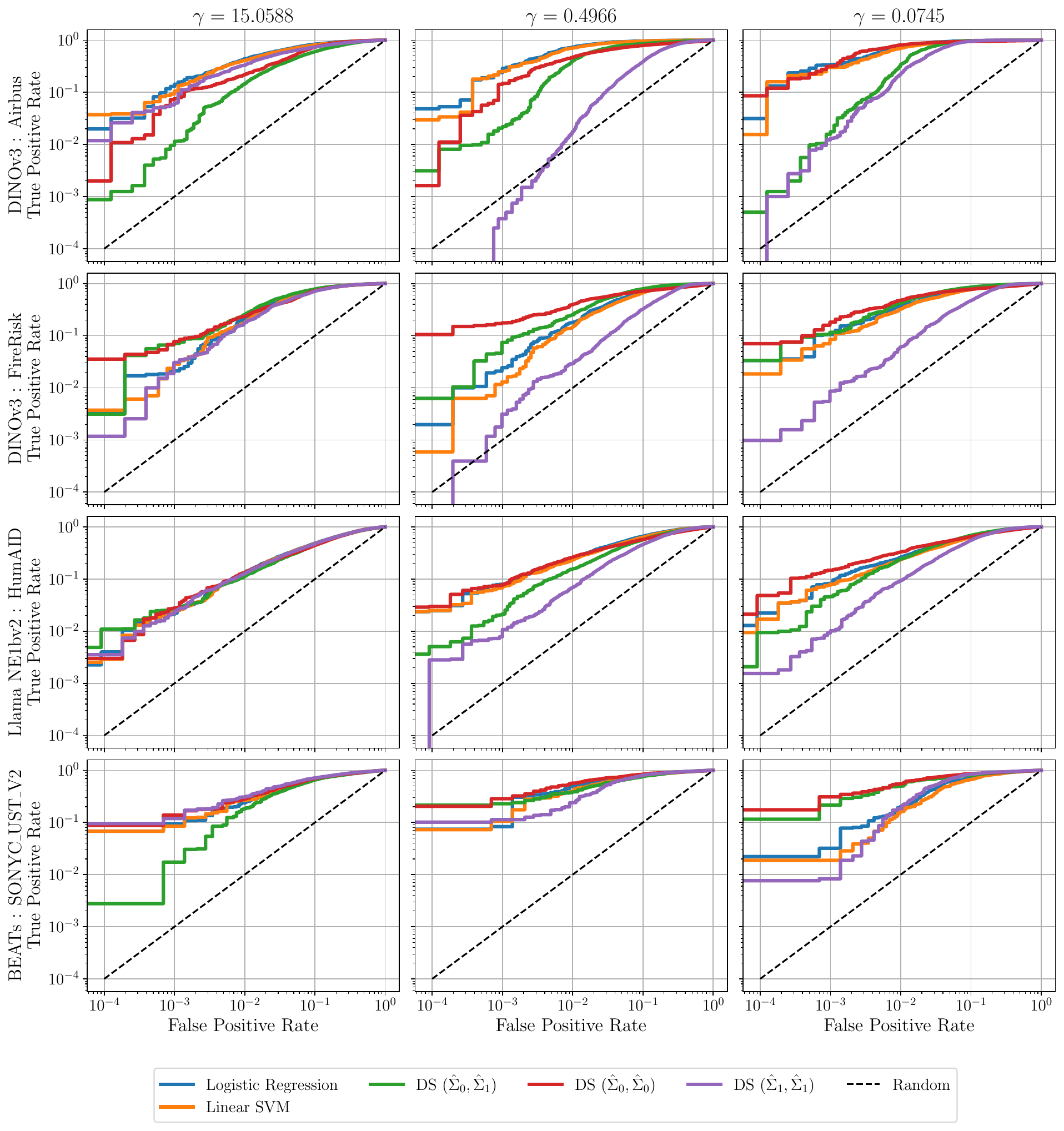}
		\caption{Logarithmic scale.}
	\end{subfigure}
	\caption{ROC curves: Comparison with benchmark classifiers in linear and log-scale.}
	\label{fig:supp_nebula_roc}
\end{figure}

For completeness, we also display the results of Figure \ref{fig:nebula_roc} displayed in the full range of false positive range both in linear and log space in Figure \ref{fig:supp_nebula_roc}.

\rev{
\subsection{Beyond foundation models}

To further assess the generality of our framework beyond modern foundation models, we extended our experimental evaluation to earlier and more compact neural architectures. In particular, for the text modality, we considered BERT-base (110M parameters) and ALBERT-base (11.8M parameters) on the HumAID dataset. Interestingly, despite their substantially different architectures and model sizes, both produce latent representations that closely resemble isotropic Gaussian distributions. Moreover, as illustrated in Figure~\ref{fig:supp_bert_and_albert}, the empirical error decay rates closely follow our theoretical predictions across different values of the shape parameter $\gamma$. These findings provide further evidence that the applicability of our high-dimensional framework is not restricted to recent large-scale foundation models, but extends to a broader family of representation-learning architectures.

\begin{figure}[h]
    \centering
    \includegraphics[width=\linewidth]{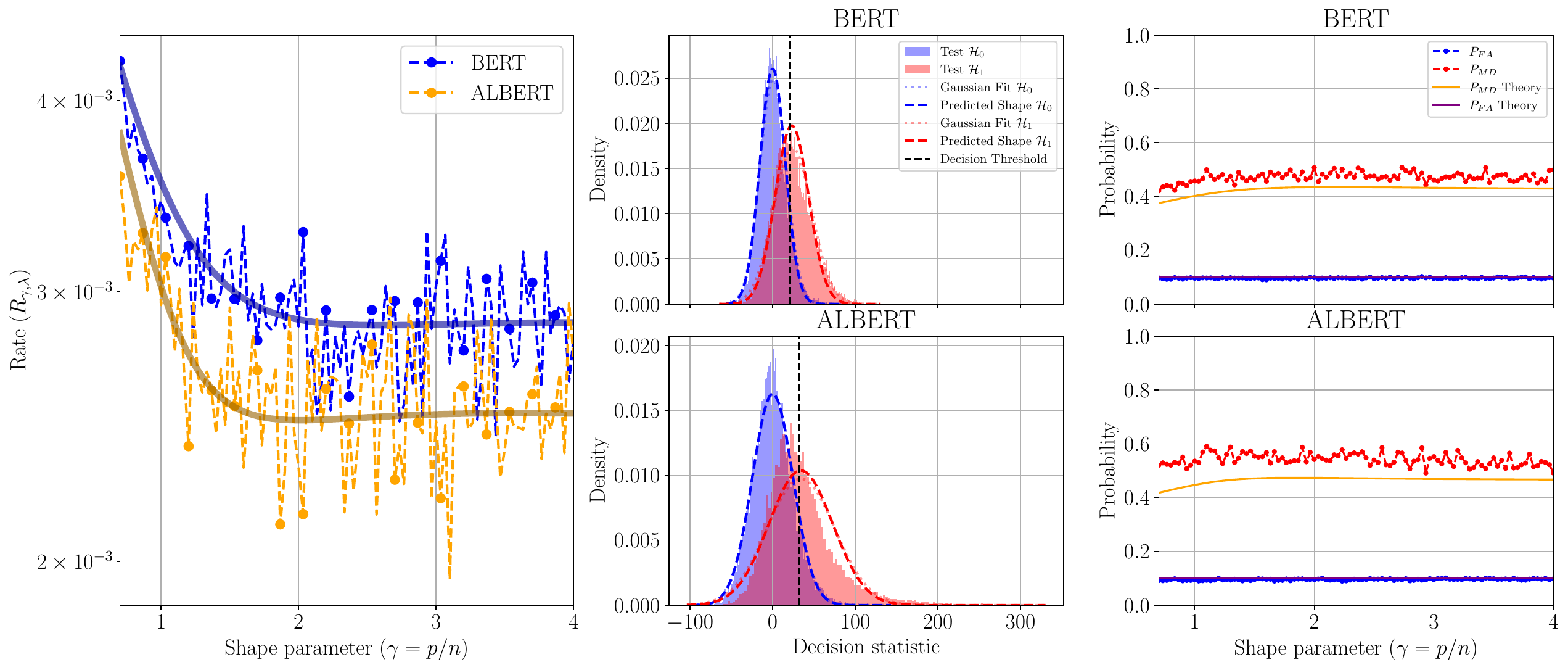}
    \caption{\rev{Beyond foundation models: Comparison between theoretical and empirical error decay rates as a function of the shape parameter $\gamma=p/n$ for BERT-base and ALBERT-base representations of the HumAID dataset.}}
    \label{fig:supp_bert_and_albert}
\end{figure}

}


\end{document}